\documentclass[10pt,letterpaper]{article}

\usepackage[margin=1in,headheight=14pt]{geometry}
\PassOptionsToPackage{table}{xcolor}
\usepackage[T1]{fontenc}
\usepackage{amsmath}
\usepackage{newpxtext}
\usepackage{newpxmath}
\usepackage{xcolor}
\definecolor{vizBlue}{HTML}{2A78D6}
\definecolor{vizOrange}{HTML}{EB6834}
\definecolor{vizAqua}{HTML}{1BAF7A}
\definecolor{vizYellow}{HTML}{EDA100}
\definecolor{vizMagenta}{HTML}{E87BA4}
\definecolor{vizGreen}{HTML}{008300}
\definecolor{vizViolet}{HTML}{6250D6}
\definecolor{vizRed}{HTML}{E34948}
\definecolor{vizInk}{HTML}{0B0B0B}
\definecolor{vizInkTwo}{HTML}{52514E}
\definecolor{vizMuted}{HTML}{898781}
\definecolor{vizGrid}{HTML}{E1E0D9}
\definecolor{vizAxis}{HTML}{C3C2B7}
\definecolor{vizPanel}{HTML}{F2F5FA}
\definecolor{vizLink}{HTML}{1F5FA8}
\usepackage{xspace}
\usepackage{tikz}
\usetikzlibrary{arrows.meta,positioning,fit,calc,backgrounds,decorations.pathreplacing,patterns}
\usepackage{pgfplots}
\pgfplotsset{compat=1.18}
\usepgfplotslibrary{fillbetween}
\colorlet{mSeq}{vizInkTwo}\colorlet{mKL}{vizBlue}\colorlet{mSATS}{vizAqua}\colorlet{mSEEKR}{vizYellow}\colorlet{mOPRD}{vizMagenta}\colorlet{mFSARs}{vizGreen}\colorlet{mFSAR}{vizViolet}\colorlet{mKLf}{vizRed}
\pgfplotsset{vizaxis/.style={axis x line*=bottom, axis y line*=left, axis line style={vizAxis}, tick style={vizAxis}, major tick length=2pt, ymajorgrids, grid style={vizGrid, line width=0.4pt}, xticklabel style={font=\scriptsize, text=vizInkTwo}, yticklabel style={font=\scriptsize, text=vizInkTwo}, label style={font=\footnotesize, text=vizInk}, legend style={font=\scriptsize, draw=none, fill=none, text=vizInk}, legend cell align=left, every axis plot/.append style={line width=1.3pt}}}

\newcommand{\method}{FSAR\xspace}
\newcommand{\Norm}{\textsc{Norm}\xspace}
\newcommand{\NormD}{\textsc{Norm}\textsubscript{dom}\xspace}
\newcommand{\BWT}{\mathrm{BWT}}
\newcommand{\ltok}{\ell_{\mathrm{tok}}}
\newcommand{\lsats}{\ell_{\mathrm{SATS}}}
\newcommand{\dlt}[1]{{\scriptsize\color{black!60}(#1)}}

\usepackage{microtype}
\usepackage{booktabs}
\usepackage{multirow}
\usepackage{graphicx}
\usepackage{capt-of}
\usepackage[numbers,square,sort&compress]{natbib}
\PassOptionsToPackage{hyphens}{url}
\usepackage[colorlinks=true,linkcolor=vizLink,citecolor=vizLink,urlcolor=vizLink]{hyperref}
\usepackage{cleveref}
\usepackage{float}
\usepackage[font=small,labelfont=bf]{caption}
\usepackage{parskip}
\makeatletter
\renewcommand{\@maketitle}{\newpage\null\vskip -1.4em\begin{center}{\LARGE \@title \par}\vskip 1.0em{\large \lineskip .5em \begin{tabular}[t]{c}\@author\end{tabular}\par}\end{center}\par\vskip 0.6em}
\makeatother
\usepackage[most]{tcolorbox}
\usepackage{fancyhdr}
\fancypagestyle{plain}{\fancyhf{}\fancyfoot[C]{\small\thepage}}
\renewenvironment{abstract}{\begin{tcolorbox}[enhanced, colback=vizPanel, colframe=vizPanel, boxrule=0pt, arc=3pt, left=12pt, right=12pt, top=9pt, bottom=9pt]\small\noindent\textbf{Abstract}\par\smallskip\noindent\ignorespaces}{\end{tcolorbox}}
\hypersetup{pdftitle={Replay for LLM Continual Learning with Multi-Teacher On-Policy Distillation}, pdfauthor={Yiqiao Qiu}}

\title{Replay for LLM Continual Learning with\\Multi-Teacher On-Policy Distillation}
\author{Yiqiao Qiu\thanks{Corresponding author: yiqiaoqiu@hotmail.com}\kern0.45em\addtocounter{footnote}{1}\thanks{Work not related to role at Fireworks AI.}\\[2pt] Independent Researcher}

\begin{document}
\begingroup\setlength{\parskip}{0pt}
\maketitle

\begin{abstract}
Multi-teacher on-policy distillation (MOPD) distills several domain experts into one student and is used in the post-training of recent large language models. It trains on all domains at once, so every expert must be available at the same time. In practice, capabilities often arrive one after another, and the experts of earlier domains may be gone when a new domain is learned. Distilling the experts in sequence then suffers from catastrophic forgetting of the earlier domains. Some recent pipelines store a snapshot after every stage and end with a multi-teacher recovery stage taught by all snapshots. The number of stored models then grows with the number of stages. Different from these pipelines, this paper formulates continual MOPD by what is kept from earlier stages and studies replay toward the previous student. Inside every stage, prompts of earlier domains are replayed with the student of the previous stage as their teacher, so neither earlier experts nor stage snapshots are kept. Two design choices of replay are explored. The replay budget of a stage is either shared among the earlier domains or fixed per earlier domain. Replay steps either distill the token distribution alone or add alignment of hidden states, region-pooled self-attention scores and value relations. A five-domain benchmark of logic, math, code, safety and instruction following is built from public experts of one 1.5B student. Experiments on its 14 benchmarks show that, without storing any snapshot, replay toward the previous student closes most of the gap between sequential distillation and joint MOPD. With a fixed replay share per earlier domain, it performs on par with recovery from all stored snapshots.
\end{abstract}

\begin{figure}[H]
\centering
\IfFileExists{figures/teaser.pdf}{\includegraphics{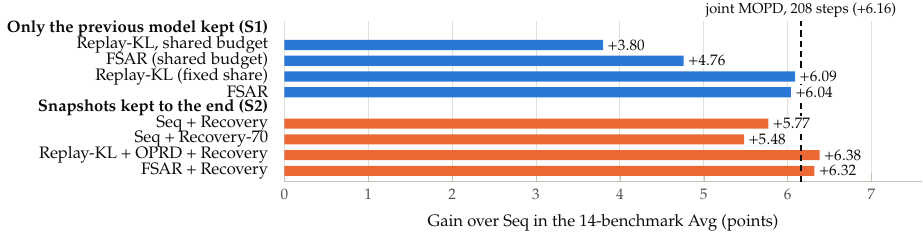}}{\begin{tikzpicture}
\begin{axis}[
  set layers, clip=false, scale only axis, xbar, bar shift=0pt, bar width=4.9pt, width=4.25in, height=1.07in,
  xmin=0, xmax=7.6, ymin=0.3, ymax=10.5, axis x line*=bottom, axis y line*=left, x axis line style={vizAxis}, y axis line style={draw=none}, tick style={vizAxis},
  xmajorgrids, grid style={vizGrid, line width=0.4pt}, xtick={0,1,2,3,4,5,6,7},
  xticklabel style={font=\scriptsize, text=vizInkTwo}, xlabel={Gain over Seq in the 14-benchmark Avg (points)}, xlabel style={font=\footnotesize, text=vizInk},
  ytick={1,2,3,4,5,6,7,8,9,10}, yticklabels={{\method + Recovery},{Replay-KL + OPRD + Recovery},{Seq + Recovery-70},{Seq + Recovery},{\textbf{Snapshots kept to the end (S2)}},{\method},{Replay-KL (fixed share)},{\method (shared budget)},{Replay-KL, shared budget},{\textbf{Only the previous model kept (S1)}}},
  yticklabel style={font=\footnotesize, text=vizInk, align=right}, ytick style={draw=none},
  nodes near coords={\pgfmathprintnumber[fixed, precision=2, fixed zerofill, showpos]{\pgfplotspointmeta}}, point meta=x,
  nodes near coords style={font=\scriptsize, text=vizInk, anchor=west, xshift=1pt, fill=white, inner sep=1pt},
]
\addplot[fill=vizBlue, draw=none] coordinates {(3.80,9) (4.76,8) (6.09,7) (6.04,6)};
\addplot[fill=vizOrange, draw=none] coordinates {(5.77,4) (5.48,3) (6.38,2) (6.32,1)};
\draw[on layer=axis grid, vizInk, densely dashed, line width=0.8pt] (axis cs:6.16,0.3) -- (axis cs:6.16,10.5);
\node[font=\scriptsize, text=vizInk, anchor=south, fill=white, inner sep=1.5pt] at (axis cs:6.16,10.5) {joint MOPD, 208 steps (+6.16)};
\end{axis}
\end{tikzpicture}}
\caption{Gain over Seq in the 14-benchmark Avg after five stages, by what is kept (\cref{tab:byscenario}). Seq is sequential distillation without replay or recovery. On replay steps, Replay-KL distills only the token distribution. Replay-KL + OPRD adds the hidden-state term, and \method adds all three alignment terms. Seq + Recovery-70 has a 70-step recovery stage. Values are means over three evaluation seeds, so gains and differences below 0.46 points are within the evaluation noise. Over two training runs, Replay-KL (fixed share), Seq + Recovery and Seq + Recovery-70 gain 5.77, 5.90 and 5.89 over the two-run mean of Seq (\cref{tab:trainseeds}). Blue: only the previous model is kept (S1); orange: every stage snapshot is kept (S2). Dashed: joint MOPD with every expert at once.}
\label{fig:teaser}
\end{figure}
\endgroup

\section{Introduction}
\label{sec:intro}

Knowledge distillation trains a student to match a teacher's output distribution \citep{hinton2015distilling}; on-policy distillation computes that match on the student's own samples, so the student is corrected on the states it actually visits \citep{agarwal2024gkd,gu2024minillm}. Multi-teacher on-policy distillation (MOPD) applies the same objective to several domain experts at once: each prompt is routed to the expert of its domain, and one student absorbs all of them. Recent model reports use this recipe in joint post-training stages \citep{xiaomi2026mimo,deepseek2026v4,nvidia2026nemotron3ultra}, and one MOPD study reports a normalized score of 0.937 for joint MOPD against 0.775 for sequential reinforcement learning on the same domains \citep{ma2026mopd}. In every case the joint stages need all experts at the same time.

Capabilities, however, often arrive one after another, and what can be kept differs between settings. In a snapshot-free setting the learner holds the new expert, the student produced by the previous stage and the prompts of earlier domains, but no earlier expert and no stage snapshot. Some recent pipelines keep more: GLM-5 trains its post-training stages in sequence and ends with a cross-stage distillation stage taught by the final checkpoint of every stage \citep[\S3.5]{glm2026glm5}, and ACLArena evaluates the same idea as a multi-teacher recovery stage \citep{wang2026aclarena}; such pipelines store every stage snapshot until the end and are therefore not snapshot-free. When all experts are available at once, joint MOPD is possible instead \citep{deepseek2026v4}. These reports do not compare the alternatives under the same training budget. A release cycle mixes these cases. When a released model learns a new segment of domains, the experts of the new segment are at hand. Those of its earlier domains may be gone, and the released model is then their only teacher.

This paper treats continual MOPD as a design space defined by what is kept (\cref{sec:scenarios}). Inside each stage, prompts of earlier domains can be replayed with the previous-stage student as their teacher (replay toward the previous student, or replay for short). The replay budget of a stage is either shared among the earlier domains or fixed per domain. Replay steps distill the token distribution alone or add alignment of hidden states \citep{yang2026oprd}, region-pooled self-attention scores \citep{qiu2023sats} and value relations \citep{wang2020minilm}. If the stage snapshots are kept, which relaxes the continual setting, a final multi-teacher recovery stage can follow. Replay with all three alignment terms is labelled \method (full-signal aligned replay). Throughout the paper, \method without a budget label uses the fixed share of seven replay steps per earlier domain and stage.

Experiments extend the same-family 1.5B setting of Latent-MOPD \citep{fang2026latentmopd} from three to five domains (logic, math, code, safety, instruction following), each with a public expert trained from the same student, and evaluate 14 benchmarks (\cref{sec:experiments}). After five stages, sequential distillation without replay or recovery ends 6.2 points below the 208-step joint MOPD on the 14-benchmark average. Without snapshots, replay closes 3.8 of these points when the earlier domains share the replay budget of a stage. With a fixed share of seven replay steps per earlier domain and stage, it closes 5.8 over two training runs. This is level with the 5.9 of a final recovery stage from all stored snapshots, both at its default length and when given as many training steps as the fixed share (\cref{sec:q-replay,sec:q-budget}). With the snapshots stored, replay followed by recovery ends level with the 208-step joint MOPD (\cref{sec:q-recovery}). Under the shared budget, each earlier domain is replayed little. There the hidden-state term on replay steps adds about one point in the 14-benchmark average, and so do the three terms together (\cref{sec:q-signal}). Against the shared budget, a fixed share lowers the score on the newest domain. Recovery lowers this score as well, after five of the six configurations it follows. The replay steps of the fixed share grow quadratically with the number of stages (\cref{sec:q-cost}).

Of four merges without training, the uniform average of the experts ends below every replay configuration. The two TIES merges of the experts end above every trained configuration, by more than 3 points when every expert is kept to the end. Applied stage by stage, TIES is judged safe on 19.6\% of the StrongREJECT prompts. All five experts are trained from the student itself, the case most favourable to merging in weight space: each task vector has 0.04\% to 0.50\% of the norm of the student's weights (\cref{app:baselines}). \Cref{sec:byscenario} summarizes the results by setting.

The contributions are:
\begin{enumerate}
\item A formulation of continual MOPD by what is kept, from the previous-stage student alone to every stage snapshot, which brings the teacher-stream setting of continual distillation \citep{michel2026cd} to on-policy distillation of LLMs, and a five-domain benchmark of 14 tests built from public experts of one 1.5B student (\cref{sec:setting,sec:setup}).
\item Measurements after five stages of replay in snapshot-free continual learning, with the replay budget shared among the earlier domains or fixed per earlier domain. The same measurements cover the pipelines that store every snapshot for a final recovery stage, with and without replay. Both are compared with sequential distillation and with joint MOPD trained nearly as long as the longest of these configurations. The shared budget and recovery without replay are also measured after three stages (\cref{sec:q-replay,sec:q-budget,sec:q-recovery}).
\item Measurements of when alignment on replay steps adds to token-level replay, across replay budgets, with and without recovery, and under three ways of counting safety. They include the direction of the gradient of each alignment term relative to the token-level gradient (\cref{sec:q-signal}).
\item The costs of each choice: the score on the newest domain, training steps and the models that must be stored (\cref{sec:q-cost}).
\end{enumerate}

\begin{figure*}[t]
\centering
\IfFileExists{figures/regimes.pdf}{\includegraphics[width=\textwidth]{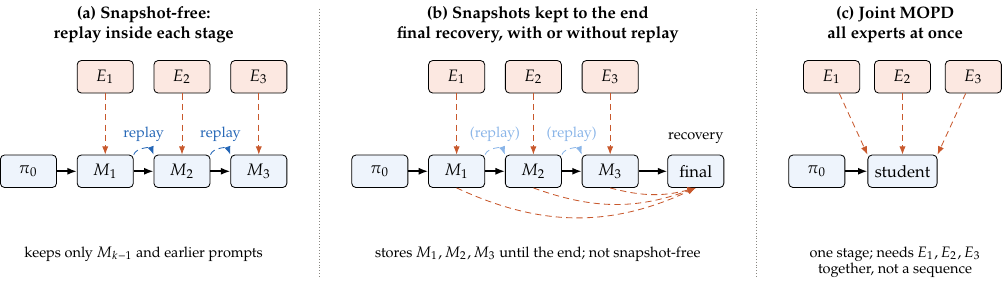}}{\begin{tikzpicture}[
  font=\footnotesize,
  stu/.style={draw, rounded corners=2pt, minimum width=0.95cm, minimum height=0.55cm, fill=vizBlue!8},
  exp/.style={draw, rounded corners=2pt, minimum width=0.95cm, minimum height=0.55cm, fill=vizOrange!15},
  arr/.style={-{Latex[length=1.6mm]}, thick},
  teach/.style={-{Latex[length=1.4mm]}, densely dashed, vizOrange!85!black},
  rep/.style={-{Latex[length=1.4mm]}, densely dashed, vizBlue!85!black},
  repopt/.style={-{Latex[length=1.4mm]}, densely dashed, vizBlue!55},
  lab/.style={font=\footnotesize\bfseries},
  note/.style={align=center, font=\scriptsize, anchor=north},
  sep/.style={vizMuted, densely dotted}
]
\node[lab, anchor=south, align=center] at (1.95,1.45) {(a) Snapshot-free:\\replay inside each stage};
\node[exp] (a1) at (1.3,1.0) {$E_1$};
\node[exp] (a2) at (2.6,1.0) {$E_2$};
\node[exp] (a3) at (3.9,1.0) {$E_3$};
\node[stu] (ap) at (0,-0.6) {$\pi_0$};
\node[stu] (am1) at (1.3,-0.6) {$M_1$};
\node[stu] (am2) at (2.6,-0.6) {$M_2$};
\node[stu] (am3) at (3.9,-0.6) {$M_3$};
\draw[arr] (ap) -- (am1);
\draw[arr] (am1) -- (am2);
\draw[arr] (am2) -- (am3);
\draw[teach] (a1) -- (am1);
\draw[teach] (a2) -- (am2);
\draw[teach] (a3) -- (am3);
\draw[rep] (am1.north east) to[out=70, in=110] node[above, font=\scriptsize] {replay} (am2.north west);
\draw[rep] (am2.north east) to[out=70, in=110] node[above, font=\scriptsize] {replay} (am3.north west);
\node[note] at (1.95,-1.75) {keeps only $M_{k-1}$ and earlier prompts};

\draw[sep] (4.925,2.15) -- (4.925,-2.4);

\begin{scope}[xshift=5.95cm]
\node[lab, anchor=south, align=center] at (2.675,1.45) {(b) Snapshots kept to the end\\final recovery, with or without replay};
\node[exp] (b1) at (1.3,1.0) {$E_1$};
\node[exp] (b2) at (2.6,1.0) {$E_2$};
\node[exp] (b3) at (3.9,1.0) {$E_3$};
\node[stu] (bp) at (0,-0.6) {$\pi_0$};
\node[stu] (bm1) at (1.3,-0.6) {$M_1$};
\node[stu] (bm2) at (2.6,-0.6) {$M_2$};
\node[stu] (bm3) at (3.9,-0.6) {$M_3$};
\node[stu] (bf) at (5.35,-0.6) {final};
\node[font=\scriptsize, above=0.6mm of bf] {recovery};
\draw[arr] (bp) -- (bm1);
\draw[arr] (bm1) -- (bm2);
\draw[arr] (bm2) -- (bm3);
\draw[arr] (bm3) -- (bf);
\draw[teach] (b1) -- (bm1);
\draw[teach] (b2) -- (bm2);
\draw[teach] (b3) -- (bm3);
\draw[repopt] (bm1.north east) to[out=70, in=110] node[above, font=\scriptsize] {(replay)} (bm2.north west);
\draw[repopt] (bm2.north east) to[out=70, in=110] node[above, font=\scriptsize] {(replay)} (bm3.north west);
\draw[teach] (bm1.south) to[bend right=25] (bf.south);
\draw[teach] (bm2.south) to[bend right=20] (bf.south);
\draw[teach] (bm3.south) to[bend right=15] (bf.south);
\node[note] at (2.675,-1.75) {stores $M_1, M_2, M_3$ until the end; not snapshot-free};
\end{scope}

\draw[sep] (12.325,2.15) -- (12.325,-2.4);

\begin{scope}[xshift=13.35cm]
\node[lab, anchor=south, align=center] at (1.325,1.45) {(c) Joint MOPD\\all experts at once};
\node[exp] (ja) at (0.25,1.0) {$E_1$};
\node[exp] (jb) at (1.45,1.0) {$E_2$};
\node[exp] (jc) at (2.65,1.0) {$E_3$};
\node[stu] (jp) at (0,-0.6) {$\pi_0$};
\node[stu] (js) at (1.45,-0.6) {student};
\draw[arr] (jp) -- (js);
\draw[teach] (ja) -- (js.north west);
\draw[teach] (jb) -- (js.north);
\draw[teach] (jc) -- (js.north east);
\node[note] at (1.325,-1.75) {one stage; needs $E_1, E_2, E_3$\\together, not a sequence};
\end{scope}
\end{tikzpicture}}
\caption[Three ways to train one student from experts]{Three ways to train one student from experts $E_1, E_2, E_3$. Orange dashed arrows: on-policy distillation from an expert or a stage snapshot. Blue dashed arrows: replay steps, on which the previous-stage student $M_{k-1}$ teaches the prompts of earlier domains. Solid arrows: the sequence of stages, each starting from the model before it.\newline
\textbf{(a)} Snapshot-free continual learning (S1 in \cref{sec:scenarios}): earlier prompts are replayed inside every stage with the previous model as their teacher, and only that model is kept. Replay-KL (token-level loss only) and \method are of this kind (\cref{fig:framework}).\newline
\textbf{(b)} Every stage snapshot kept to the end (S2): a final multi-teacher recovery stage taught by the snapshots follows the sequence. Without replay this is Seq + Recovery; with the replay of (a), drawn as light arrows, it is replay followed by recovery. Storing the snapshots relaxes the continual setting.\newline
\textbf{(c)} Joint MOPD: one stage that trains on all domains at once and needs all experts at the same time.}
\label{fig:regimes}
\end{figure*}

\section{Related Work}
\label{sec:related}

\paragraph{Multi-teacher on-policy distillation.}
On-policy distillation trains the student on its own samples against the teacher's token distribution \citep{agarwal2024gkd,gu2024minillm}, building on the soft-target view of distillation \citep{hinton2015distilling}; Qwen3 uses it with a single teacher \citep{yang2025qwen3}. With several experts, prompts are routed to the expert of their domain and the student is trained on all domains in joint stages \citep{xiaomi2026mimo,ma2026mopd,deepseek2026v4}. Nemotron 3 Ultra uses two such stages \citep{nvidia2026nemotron3ultra}. Variants that keep every teacher available counter interference between domains in several ways: token-budget balancing \citep{gao2026openmopd}, adaptive domain schedules \citep{sun2026d3mopd,gan2026nashmopd}, projected updates \citep{zhu2026upmopd,liu2026pmopd}, a moving-average reference of the student \citep{yin2026sfmopd}, non-routed teachers acting as critics \citep{tian2026roadmopd}, and token-level teacher weights that need no domain labels \citep{sun2026trustmopd} and follow each teacher's estimated reliability \citep{lakshminarayanan2017ensembles,zheng2024dmr}. Other methods fuse the specialists without training \citep{yang2026tfopd} or transfer the change that post-training made to a teacher \citep{sang2026deltamopd,hou2026waypointopd}.

Other studies compare MOPD with supervised fine-tuning, soft-label distillation and weight merging for experts trained from a shared base \citep{xie2026rethinking}, show that for experts in a shared basin the MOPD objective matches Fisher merging to second order \citep{liu2026fishermopd}, and compare weight merging, mixed reinforcement learning and MOPD with shared experts and data \citep{wu2026fusion}. \Cref{tab:five} lists four merges without training as references: the uniform average of the five experts and three TIES merges \citep{yadav2023ties}. These merge the five experts at the end, the experts stage by stage as in continual model merging \citep{tang2025opcm}, and the Seq snapshots, as MagMax does with the models of a sequential run \citep{marczak2024magmax}. Neither the model reports that use a joint stage nor GLM-5 compares the two pipelines under the same training budget. The controlled comparisons that could be found use neither replay toward the previous student nor a recovery stage. They compare joint MOPD with sequential reinforcement learning \citep{ma2026mopd,lian2026uimopd}, with sequential distillation from one soft-prompt teacher per task \citep{ma2026promptsd}, and with ordered task blocks in which each teacher teaches its own block and earlier blocks are protected by projecting gradients and optimizer updates out of their subspaces \citep{liu2026pmopd}. In the last study a single cycle uses each teacher once and keeps no earlier teacher, although its projection memory stores the update subspaces of earlier blocks. Its best one-cycle task order scored above joint MOPD with the domains mixed in every batch (65.81 against 64.43 on Qwen2.5-7B), and its default setting revisits every teacher in four cycles.

\paragraph{Sequential and continual distillation.}
Sequential on-policy distillation overwrites earlier skills when the experts conflict: in a controlled planning environment, a student taught by three teachers in turn forgets severely unless the teachers share planning patterns \citep{men2026physics}. When post-training is sequential, on-policy distillation from earlier checkpoints is used to recover afterwards. GLM-5 ends with a cross-stage distillation stage taught by the final checkpoints of the preceding stages \citep[\S3.5]{glm2026glm5}. Nemotron-Cascade 2 inserts one multi-domain distillation stage into its cascade of RL stages \citep{yang2026nemotroncascade2}. ACLArena evaluates distillation from earlier stage checkpoints, with one teacher and with several, in a four-stage sequence. It compares this with self-distilled fine-tuning and model merging as ways to recover earlier capabilities and proposes a recipe that combines offline replay of high-quality trajectories with routed LoRA experts \citep{wang2026aclarena}. Distillation from the checkpoint before fine-tuning restores capabilities lost to it \citep{liu2026sdrecovery}. After a post-trained Qwen3-8B is fine-tuned on internal documents, its IF-eval score drops from 85 to 79, and on-policy distillation toward the earlier model on unrelated prompts brings it back to 83; the authors suggest alternating such distillation phases with fine-tuning on new data \citep{lu2025opd}. A cross-domain expert distillation stage restores the general capabilities of a search-specialized student \citep{chen2026hybridopd}, and KuaiRP recovers general capabilities with a two-stage on-policy distillation in which the domain-adapted model teaches the base model \citep{wang2026kuairp}. The closest designs route each kind of prompt to its own teacher on the student's rollouts within one final stage. TS-OPD reconciles question answering and speech recognition in speech language models: the model before speech specialization teaches question answering, the specialized model teaches recognition, and the student starts from the former \citep{guo2026tsopd}. CaMOPD and UCMOPD recover the general capabilities of a specialized model with the original model as the general teacher and the specialized model, frozen, as the domain teacher \citep{chen2026camopd,liu2026ucmopd}; CaMOPD alternates three general steps with one domain step.

Seq + Recovery (\cref{sec:baselines}) is the multi-domain form of the GLM-5 stage, CaMOPD and UCMOPD: it starts from the newest model, and the prompts of each domain go to the checkpoint that ended that domain's stage. TS-OPD starts from the earlier model instead, so with two domains it matches one stage of the in-stage replay studied here.

Sequential versions of SDFT and SDPO keep the previous-stage checkpoint as the teacher (SDFT) or as the generator of preference pairs (SDPO) in every stage, on the new domain's data and without replay \citep{harrington2026cl}. None of these compares a final recovery stage with on-policy replay toward the previous student inside the stages, which \cref{sec:q-budget,sec:q-recovery} do.

\paragraph{The previous model as a second teacher.}
Keeping the previous model as a second teacher next to the new one is a common continual learning pattern in vision \citep{hou2018pdr,zhang2020dmc,zhao2026graft}, machine translation \citep{zhang2023ckd} and reinforcement learning, where PLAiD distills a new task expert and the previous policy on the policy's own rollouts \citep{berseth2018plaid}. Continual distillation \citep{michel2026cd} poses the constraint of the snapshot-free setting for image classifiers: a student learns from a stream of teachers without access to earlier ones, and in its method the previous student checkpoint teaches external unlabeled data next to the current teacher, with offline distillation of logits.

In LLMs, replay toward the previous model has mostly been off-policy and combined with supervised learning of each new task: on the previous model's completions of old or synthesized prompts \citep{li2026cagd,huang2024ssr}, on stored samples \citep{wen2026spark}, toward the optimal policy of earlier tasks on replayed prompts and ranked responses \citep{zhang2023copr}, or on replayed batches distilled toward the snapshot taken at the end of their task \citep{meng2026tsr}. Pailitao-MMSearch covers one transition on-policy: while domain training continues, a reverse KL toward the base model acts on the student's rollouts of general prompts \citep{ye2026pailitao}. OPR instead rolls out the most recent checkpoint on old prompts, keeps the completions that a task reward accepts and trains on them as supervised examples, without a teacher \citep{chen2026opr}. Generated pseudo-samples of old tasks have also been replayed with the language-modeling loss \citep{sun2020lamol,chuang2020l2kd}. HERD keeps the student within drift budgets around the previous checkpoint and the base model while it distills outputs produced with an evolving harness \citep{sun2026herd}. Sleep alternates use with consolidation phases in which on-policy distillation moves knowledge from the model before the update into added low-rank experts, with rehearsal on self-generated data instead of stored examples \citep{behrouz2026sleep}; here the earlier domains are taught by the previous-stage student on stored prompts, and the student keeps a fixed architecture. In continual RL of language models, the KL-to-old-policy baseline of CPR penalizes divergence from the pre-task policy on earlier prompts mixed into training, with a coefficient of 0.001 \citep{luo2026cpr}, a small-weight regularizer rather than a distillation stage, and CPPO, for continual learning from human feedback, adds a penalty toward the log-probabilities of the previous policy and weights it sample by sample over the rollouts \citep{zhang2024cppo}. In continual reinforcement learning of robots, CIDER freezes the accumulated policy as a teacher before each new task and interleaves task learning with distillation toward it on stored states \citep{li2026cider}; this is the closest design to the replay component here. Against one offline consolidation of independently trained experts at the end, which falls to 51.3\% mean success after six stages, CIDER keeps 100\%, and task-balanced sampling of earlier tasks retains more than sampling in proportion to their data.

Inside a single joint MOPD stage, the student from before the stage already teaches the domains that no specialized teacher covers. MiMo-V2-Flash counts the student itself among its teachers \citep{xiaomi2026mimo}. Nemotron 3 Ultra uses its student from reinforcement learning with verifiable rewards as a self-teacher for areas without a specialized teacher \citep{nvidia2026nemotron3ultra}. For a single sequential transition, TS-OPD starts from the model before speech-recognition specialization and lets it teach question answering while the specialized model teaches recognition \citep{guo2026tsopd}. No earlier LLM work that could be found repeats this in every stage of a sequence, where the previous-stage student is the only teacher of all earlier domains, or varies how many replay steps each earlier domain receives.

Replay of stored samples in LLMs often keeps a fixed fraction of the data of every earlier task \citep{wang2023trace,he2024seekr}, so that the share of each earlier task does not shrink. No LLM work that could be found compares such a fixed share with a budget shared among the earlier domains under on-policy replay toward the previous student, or compares on-policy replay with a final recovery stage. With off-policy replay and gold labels, TSR finds that distillation interleaved with training scores above a post-hoc distillation phase \citep{meng2026tsr}. In its three-task stream the most recent snapshot is the best teacher or level with the best. In its eight-task stream this snapshot is slightly below snapshots matched to each task.

\paragraph{Continual learning of LLMs.}
Following the taxonomy of a recent survey \citep{shi2025survey}, this work belongs to continual fine-tuning; another survey covers the same area from the training-stage view \citep{wu2024survey}, and TRACE is a standard benchmark for continual instruction tuning \citep{wang2023trace}. Several studies find that learning from on-policy data forgets less than supervised fine-tuning \citep{shenfeld2025razor,chen2025retaining}, which motivates self-distillation for continual post-training \citep{shenfeld2026sdft,khamis2026isdft}; the limits of on-policy self-distillation in this regime have been discussed \citep{wang2026denser}, and OPDArena finds that students of RL-trained teachers retain more prior capability than students of SFT-trained teachers \citep{xu2026opdarena}. Other lines constrain internal mechanisms, such as function vectors \citep{jiang2025fvg}, or adapt preference optimization to continual learning of multimodal models \citep{truong2026phidpo}. The present setting differs in that every new domain comes with an external expert, and the question is how to keep earlier experts' skills when those experts are gone, not how to transfer a model across domains, as in domain adaptation \citep{ganin2016dann,zhu2023tdan}.

\paragraph{Distillation and replay in continual learning.}
Classic methods regularize toward the previous model's outputs \citep{li2017lwf}, penalize changes to important weights \citep{kirkpatrick2017ewc}, replay stored samples \citep{rolnick2019replay}, combine replay of stored samples with distillation of earlier outputs \citep{rebuffi2017icarl,buzzega2020der}, or keep task-specific parameters \citep{mallya2018packnet,zhou2026tsbn}, which need the domain of each test input, for example from out-of-distribution detection \citep{hendrycks2017baseline,sun2024chm,ye2025background}; backward transfer is measured as in GEM \citep{lopezpaz2017gem}. In continual reinforcement learning of language and vision-language models, CPR replays the prompts of earlier tasks with responses regenerated by the current policy, which works better than a KL penalty toward the pre-task policy on those prompts \citep{luo2026cpr}. CoKL restricts the KL toward a frozen reference to its verified-correct responses \citep{wang2026cokl}, and CPO replaces the KL to the previous checkpoint on earlier tasks with a Fisher-based parameter anchor that needs no replay \citep{luo2026cpo}. How much to replay is set by a fixed ratio \citep{luo2026cpr,chen2026opr}, a learned mixture \citep{shihab2026proxymixing} or a forgetting-driven schedule \citep{lu2026mssr,feng2026forever}; with many tasks, replay has been found to lag multi-task training \citep{hazard2025sure}, and with abundant memory the limit moves from forgetting to learning the new task \citep{cho2025forget}.

Closest to the attention term of this work, SEEKR replays a small buffer and distills the old model's attention on heads selected by task sensitivity and forgettability \citep{he2024seekr}. ASR preserves attention without replay: it keeps per-skill spectral statistics of cross-modal attention in continual multimodal LLMs and constrains their drift \citep{zhao2026asr}. Other LLM continual methods act on attention heads or internal features without distilling attention maps: they partition attention heads across tasks \citep{shi2026heads} or distill features in the sparse basis of a pretrained sparse autoencoder \citep{zhang2026saefd}. GeRe replays a fixed set of general texts and imitates the base model's last-layer hidden states on them; in its full-parameter runs this retained more than imitating the logits or replay alone \citep{zhang2025gere}.

Pooled distillation targets go back to PODNet in incremental classification \citep{douillard2020podnet}; CPFD pools features for continual named entity recognition and outperforms both raw and over-pooled maps \citep{zhang2023cpfd}, and SATS pools self-attention vectors over class regions for continual semantic segmentation, where class-region pooling outperforms both global pooling and no pooling \citep{qiu2023sats}. Replay-KL + SATS carries the SATS idea to on-policy LLM distillation with reasoning steps as regions, and \method keeps this term next to hidden-state and value-relation alignment; no earlier LLM continual learning method that could be found pools attention over reasoning steps.

\paragraph{Representation and relation distillation.}
Attention transfer matches attention maps between teacher and student \citep{zagoruyko2017attention}, MiniLM distills self-attention distributions together with value relations \citep{wang2020minilm,wang2021minilmv2}, and MoLSAKI transfers the teacher's stepwise attention on key tokens in offline chain-of-thought distillation \citep{chen2025molsaki}. On the student's own samples, attention is distilled for multimodal models \citep{li2026ral}, from a privileged teacher that sees a verified solution \citep{arib2026opasd}, and from the final layer to intermediate layers during RL \citep{liu2026oisd}. In on-policy distillation of reasoning models, OPRD and LastOPD add a hidden-state alignment to the token-level objective \citep{yang2026oprd,yang2026lastopd}, and Latent-MOPD extends it to several teachers \citep{fang2026latentmopd}. OPMD transfers a student-compatible form of the teacher's intermediate computation \citep{shi2026opmd}, RIDE extrapolates the RL-induced residual of the teacher's hidden states \citep{li2026ride}, and SimOPD transfers response similarities instead of hidden states \citep{tang2026simopd}; none of them is continual.

Attention also exposes the structure of reasoning traces: windowed average attention distances (WAAD) segment reasoning traces into chunks \citep{li2026attention}, which gives the alternative pooling regions tested in \cref{app:pooling}. Gradient geometry in on-policy distillation has been analyzed for the language-prior and visual-grounding parts of the loss, whose gradients are nearly orthogonal \citep{yoon2026vgs}, for teachers whose gradients conflict \citep{tian2026roadmopd}, and for the gradients and Adam updates of different teachers in MOPD \citep{zhu2026gradmopd}. This paper compares attention and hidden-state alignment on the same replay steps, alone and stacked, with value-relation alignment added to the attention term. It also measures the direction of the gradient of each alignment term relative to the token-level gradient.

\section{Continual MOPD}
\label{sec:setting}

\subsection{Setting and notation}
\label{sec:notation}

Domains $d_1, \dots, d_K$ arrive in order, each with an expert $E_k$ and a prompt set $\mathcal{D}_k$. The student starts from a base model $\pi_0 = M_0$. Stage $k$ initializes the student $\pi_\theta$ from $M_{k-1}$, trains it, and outputs $M_k$:
\begin{equation*}
\pi_0 = M_0 \;\rightarrow\; M_1 \;\rightarrow\; M_2 \;\rightarrow\; \cdots \;\rightarrow\; M_K .
\end{equation*}
During stage $k$ the learner may use the new expert $E_k$, the previous-stage student $M_{k-1}$, and a replay buffer $\mathcal{B}_{<k}$ that holds prompts of $d_1, \dots, d_{k-1}$ without responses or labels. It may not use $E_j$ for $j < k$ or $M_j$ for $j < k-1$. All losses are computed on the student's own rollouts $y \sim \pi_\theta(\cdot \mid x)$, so the buffer needs no responses. By this rule, the trained configurations do not use the base student $\pi_0 = M_0$ from stage 2 on, and $M_0$ holds none of the skills learned in the stages. TIES applied stage by stage is the one snapshot-free merge that needs the base student. Besides the previous merge, it keeps this public model to compute task vectors (\cref{app:baselines}). Joint MOPD, which trains on all domains at once with every expert, is the non-continual reference.

\paragraph{Scenarios.}
\label{sec:scenarios}
Three scenarios differ in what is kept. In S1, the snapshot-free setting above, stage $k$ has only the new expert $E_k$, the previous-stage student $M_{k-1}$ and the prompts of earlier domains without responses; no earlier expert and no stage snapshot is kept, so the choices are sequential distillation and replay, with a replay budget and a replay signal. In S2 the stage snapshots $M_1, \dots, M_K$ or the original experts are still available at the end, so a final multi-teacher recovery stage can follow the sequence; when all experts are available at the same time, joint MOPD can replace the sequence. In S3 a long sequence is split into segments, for example 15 stages into three segments of five. S2 holds inside a segment and S1 between segments, so the teacher of an earlier segment is the model consolidated at the end of the previous segment (\cref{sec:discussion}). A release cycle is of this kind. A model released after one segment of domains later learns a new segment with that segment's experts. The experts of the earlier segment are then gone, and the released model is the only teacher of its domains.

The configurations compared fall into two families and a reference. Replay keeps earlier skills by on-policy distillation from one model, the previous-stage student, on replayed prompts inside every stage, and needs only that student; it is the snapshot-free family (S1). The recovery family keeps them by one multi-teacher on-policy distillation stage at the end, the recovery stage, and needs every stage snapshot or expert until then (S2); since it stores what the continual setting discards, it is a sequential pipeline with a final consolidation rather than snapshot-free continual learning. Joint MOPD avoids the sequence: it trains on all domains at once and needs every expert at the same time. Sequential distillation without replay or recovery (Seq) is the common baseline, and replay followed by recovery combines the two families.

\paragraph{Steps.} Each training step uses one block of prompts from a single domain. A new-domain step draws its block from $\mathcal{D}_k$ and a replay step from $\mathcal{B}_{<k}$; stage $k$ has $n_k$ new-domain steps. Seq has no replay steps and trains stage $k$ on $\mathcal{D}_k$ with $E_k$ only.

\paragraph{Replay budget.} Replay methods interleave replay steps with the new-domain steps of every stage $k \ge 2$, and they differ in how many replay steps each earlier domain receives. With a \emph{shared budget}, one replay step follows every $m$ new-domain steps, so stage $k$ has $n_k/m$ replay steps whatever the number of earlier domains, and the replay steps cycle over $d_1, \dots, d_{k-1}$. Each earlier domain then receives about $n_k / \big(m(k-1)\big)$ replay steps in stage $k$, a share that shrinks as domains accumulate. With a \emph{fixed share}, every earlier domain receives $b$ replay steps in every stage, so stage $k$ has $(k-1)\,b$ replay steps, spread evenly among its new-domain steps, and the replay budget of a stage grows linearly with $k$. Under both budgets the replay blocks of a domain are spread evenly over the blocks that the domain used in its own stage.

\paragraph{Recovery.} Some recent pipelines repair forgetting after the sequence instead \citep[\S3.5]{glm2026glm5}. A recovery stage follows stage $K$, starts from $M_K$, and routes the prompts of each domain $d_j$ to the student $M_j$ that ended that domain's stage, with the same number of steps for every domain, taken in turn. A replay step of stage $k$ instead teaches $d_j$ with $M_{k-1}$, which has already learned the domains after $d_j$, so its target is what $M_{k-1}$ still holds of $d_j$, not what $M_j$ held. Recovery relaxes the constraint above, because every stage snapshot $M_1, \dots, M_K$ must be kept until the end. Seq + Recovery applies it after Seq, and X + Recovery after a replay method X.

\subsection{Metrics}
\label{sec:metrics}

For a benchmark $\tau$, let $s_\tau(\cdot)$ be its score, and let $\mathcal{T}$ be the set of benchmarks of the domains learned up to stage $K$. Results are aggregated by the average
\begin{equation}
\mathrm{Avg} = \frac{1}{|\mathcal{T}|} \sum_{\tau \in \mathcal{T}} s_\tau(M_K) ,
\label{eq:avg}
\end{equation}
by two normalized scores, and by domain averages, each the mean over the benchmarks of one domain. \Norm is the normalized score of Latent-MOPD \citep[Eq.~5]{fang2026latentmopd}, the total gain over the base student divided by the total gain of the best expert on each benchmark:
\begin{equation}
\Norm = \frac{\sum_{\tau \in \mathcal{T}} \big( s_\tau(M_K) - s_\tau(\pi_0) \big)}{\sum_{\tau \in \mathcal{T}} \big( \max_{k} s_\tau(E_k) - s_\tau(\pi_0) \big)} ,
\label{eq:norm}
\end{equation}
so that 0 is the base student and 1 the best expert on every benchmark. On a fixed benchmark set \Norm is an increasing affine function of Avg, so the two rank methods in the same order; \Norm states the gain as a share of the experts' headroom.

\NormD is the per-domain normalized score of an earlier MOPD study \citep{ma2026mopd}. It rescales each domain average so that the base student scores 0 and that domain's own expert 1, then averages over the domains, so a domain with more benchmarks or more headroom does not dominate. The experts' gains over the student range from 12.10 points (instruction following) to 23.78 (safety), so in \NormD a point of instruction following weighs about twice a point of safety. A model that scores above a domain's own expert on that domain gets more than 1 for it, so \NormD can exceed 1. All aggregates are computed for each evaluation seed and then averaged over the seeds; differences in Avg and in the domain averages are given in points of the raw scores.

Backward transfer follows GEM \citep{lopezpaz2017gem}: for a benchmark $\tau$ of the domain learned in stage $j(\tau)$,
\begin{equation}
\BWT_\tau = s_\tau(M_K) - s_\tau\big(M_{j(\tau)}\big) ,
\label{eq:bwt}
\end{equation}
so a negative value is forgetting. For a configuration followed by recovery, $M_K$ in \cref{eq:avg,eq:norm,eq:bwt} is the model after the recovery stage.

\section{Replay toward the Previous Student}
\label{sec:method}

\begin{figure*}[t]
\centering
\IfFileExists{figures/framework.pdf}{\includegraphics[width=\textwidth]{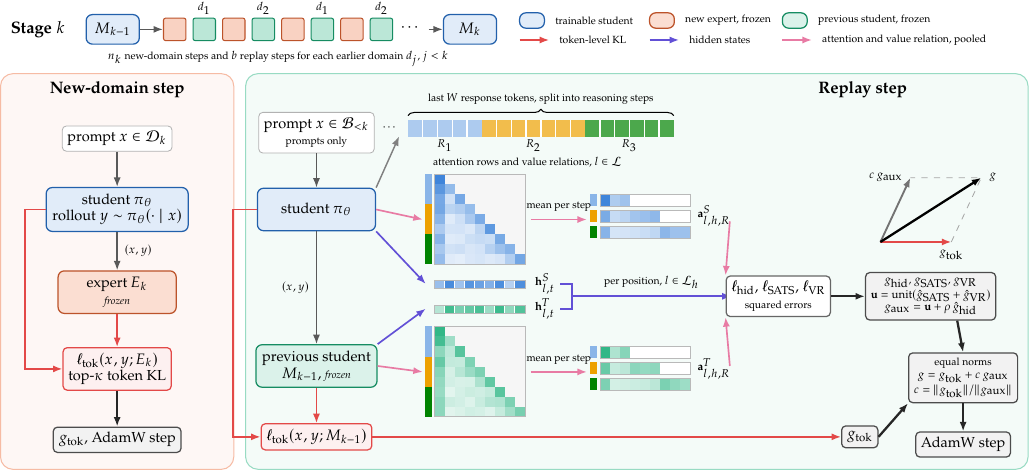}}{\input{figures/framework-body}}
\caption{Replay toward the previous student in stage $k$, drawn with the fixed share and the three alignment terms of \method. New-domain steps distill the new expert $E_k$ into the student on its own rollouts. Every earlier domain $d_j$, $j < k$, receives $b$ replay steps, spread among the new-domain steps. On a replay step the student answers a prompt of an earlier domain, and the frozen previous-stage student $M_{k-1}$ teaches on the same rollout through a token-level KL on the student's most probable tokens and three alignment terms on the last $W$ response tokens: hidden states compared per position, and attention rows and value relations averaged over each reasoning step before student and teacher are matched. The gradients of the two pooled terms are combined into one unit direction. The unit hidden-state gradient is added with weight $\rho$, and the sum is scaled to the norm of the token gradient before the optimizer step. The stage strip and the drawn maps are schematic.}
\label{fig:framework}
\end{figure*}

This section describes the four optional parts of the framework (\cref{fig:framework,fig:regimes}): replay steps, the replay budget, the replay signal and a final recovery stage. New-domain steps are those of sequential distillation. Replay steps are distilled toward the frozen previous-stage student with the same token-level loss as the new-domain steps (\cref{sec:replay}). The replay budget sets how many replay steps each earlier domain receives in each stage (\cref{sec:share}). The replay signal can add alignment terms on hidden states, attention and value relations toward the same teacher (\cref{sec:alignterms}), whose gradients are balanced against the token-level gradient by norm (\cref{sec:balance}). When stage snapshots are kept, a final recovery stage can follow (\cref{sec:notation,sec:cost}). Replay with all three alignment terms is labelled \method (full-signal aligned replay). Without a budget label \method uses the fixed share with $b = 7$ (\cref{sec:setup}); \method (shared budget) and \method, $b = 3$ use the same signal under the other budgets.

\subsection{Replay with the previous student}
\label{sec:replay}

Let $T$ denote the teacher of a step: $T = E_k$ on a new-domain step with $x \in \mathcal{D}_k$, and $T = M_{k-1}$ on a replay step with $x \in \mathcal{B}_{<k}$. The token-level channel is the same for both. At each response position $t$, let $V_t$ be the $\kappa$ most probable tokens of the student before the update (\cref{app:training}); the teacher scores the same tokens, and every candidate $v \in V_t$ receives its own reward
\begin{equation}
r_{t,v} = - w_{t,v} \big( \log \pi_\theta(v \mid x, y_{<t}) - \log \pi_T(v \mid x, y_{<t}) \big) ,
\label{eq:tokreward}
\end{equation}
where both log-probabilities are taken over the full vocabulary, so the teacher's probabilities are not renormalized over $V_t$, and $w_{t,\cdot}$ are the student's probabilities renormalized over $V_t$. Summed over $V_t$, $-r_{t,v}$ is the student-weighted log-ratio of student and teacher on the student's top-$\kappa$ support. Each reward is divided by the standard deviation of all rewards of its domain in the batch, with no mean subtracted, which gives the advantage $A_{t,v}$. The loss holds the advantages fixed and weights the log-probability of every candidate:
\begin{equation}
\ltok(x, y; T) = - \frac{1}{N} \sum_{t} \sum_{v \in V_t} A_{t,v} \log \pi_\theta(v \mid x, y_{<t}) ,
\label{eq:tokloss}
\end{equation}
where $t$ runs over the response positions of a micro-batch and $N$ is their number (\cref{app:training} gives how micro-batches are weighted). As in the top-$\kappa$ code of LastOPD \citep{yang2026lastopd}, the advantage thus has one entry per candidate token rather than one per sampled token. The LastOPD paper writes the loss as a KL divergence between both distributions renormalized over the student's top-$\kappa$ tokens. With the advantages held fixed, the gradient at position $t$ is that of the generalized KL divergence $\sum_{v \in V_t} ( p_v \log (p_v / q_v) - p_v + q_v )$ between the student and teacher probabilities $p$ and $q$ on $V_t$, divided by the standard deviation and by the student's probability mass on $V_t$. This divergence is minimal when the student matches the teacher on $V_t$. Subtracting the batch mean from the advantages would keep the exact match a stationary point. It would also add a term to the gradient wherever that mean is not zero, and this term would move the stationary point when the match cannot be reached. Each step makes one policy update on its own rollouts, so the importance ratio of the clipped policy-gradient objective \citep{schulman2017ppo} equals one and clipping never acts; there is no KL regularizer and no entropy term. Replay-KL, this component alone under the shared budget of \cref{sec:notation}, uses $\ltok(x, y; M_{k-1})$ on replay steps. Replay-KL (fixed share) uses the fixed share with the token-level loss only.

\subsection{Replay budget}
\label{sec:share}

Under the shared budget the number of replay steps in a stage does not depend on the number of earlier domains, so each earlier domain receives fewer of them as domains accumulate, and a domain learned late in the sequence is replayed in few stages with few steps in each. The fixed share gives each earlier domain $b$ replay steps in every stage instead (\cref{sec:notation}). The $(k-1)\,b$ replay steps of stage $k$ are placed evenly among its $n_k$ new-domain steps, each after a new-domain step, and the blocks of a domain are chosen evenly over its own blocks across all the stages that replay it, so that successive stages revisit different prompts of that domain. The teacher of every replay step remains $M_{k-1}$: the budget sets how often each earlier domain is distilled toward the previous student, not which models are kept.

\subsection{Replay signal: alignment terms}
\label{sec:alignterms}

On a replay step the student and the teacher $M_{k-1}$ are run on the same sequence $(x, y)$, so their positions correspond one to one, and three alignment terms compare their internal states on the last $W$ response tokens.

\paragraph{Hidden states.} The hidden-state term of OPRD \citep{yang2026oprd} is the mean squared error between student and teacher hidden states at the layers of a set $\mathcal{L}_{h}$; the cross-architecture branch of the OPRD code, OPRD-Bridge, normalizes the projected states to unit length, and the LastOPD implementation used here, which builds on that code, normalizes the states themselves, without a projector:
\begin{equation}
\ell_{\mathrm{hid}} = \frac{1}{|\mathcal{L}_{h}|\, N_W} \sum_{l \in \mathcal{L}_{h}} \sum_{t} \frac{1}{D} \bigg\lVert \frac{\mathbf{h}^{S}_{l,t}}{\lVert \mathbf{h}^{S}_{l,t} \rVert_2} - \frac{\mathbf{h}^{T}_{l,t}}{\lVert \mathbf{h}^{T}_{l,t} \rVert_2} \bigg\rVert_2^2 ,
\label{eq:hid}
\end{equation}
where $t$ runs over the last $W$ response positions of each response in a micro-batch, $N_W$ is their number and $D$ is the hidden size; the same-architecture objective of OPRD compares the unnormalized states.

\paragraph{Reasoning-step attention score pooling.}
\label{sec:sats}
The attention term is SATS \citep{qiu2023sats} with reasoning steps as regions. The last $W$ response tokens are the queries, split into reasoning steps at the step separators of the response. For a step $R$, a layer $l$ of a set $\mathcal{L}$ and a head $h$, the region-pooled attention vector is the mean of the attention rows of the step's queries. Each row covers all causal keys at full resolution:
\begin{equation}
\mathbf{a}_{l,h,R} = \frac{1}{|R|} \sum_{t \in R} \operatorname{softmax}\!\left( \frac{\mathbf{q}^{l,h}_t \, (\mathbf{K}^{l,h})^{\top}}{\sqrt{d_{\mathrm{head}}}} \right) ,
\label{eq:pool}
\end{equation}
where the softmax runs over the keys $\mathbf{K}^{l,h}$ of the positions up to $t$ and is zero beyond $t$, $\mathbf{q}$ and $\mathbf{K}$ are taken after the rotary position embedding, $d_{\mathrm{head}}$ is the head dimension, and grouped-query keys are repeated to the query heads. As in Eq.~2 of SATS \citep{qiu2023sats}, the loss is the squared distance between student and teacher vectors concatenated over the layers, averaged over the heads, over the steps of each response and over the responses:
\begin{equation}
\lsats = \frac{1}{|\mathcal{Y}|\,H} \sum_{y \in \mathcal{Y}} \frac{1}{|\mathcal{R}_y|} \sum_{R \in \mathcal{R}_y} \sum_{h=1}^{H} \sum_{l \in \mathcal{L}} \big\lVert \mathbf{a}^{S}_{l,h,R} - \mathbf{a}^{T}_{l,h,R} \big\rVert_2^2 ,
\label{eq:sats}
\end{equation}
where the $H$ query heads of each layer are matched by index, $\mathcal{R}_y$ is the set of steps of response $y$ in the window and $\mathcal{Y}$ holds the responses of a micro-batch with at least one step. Pooling over a step constrains the step's dependence on its context rather than every token's attention row, which in the original SATS keeps the constraint loose enough for new learning \citep{qiu2023sats}.

\paragraph{Value relation.} Following MiniLM \citep{wang2020minilm,wang2021minilmv2}, the value-relation term (VR) replaces the attention rows of \cref{eq:pool} with the rows of $\operatorname{softmax}\big(\mathbf{V}^{l,g} (\mathbf{V}^{l,g})^{\top} / \sqrt{d_{\mathrm{head}}}\big)$ for each key-value head $g$ over the same causal positions. The rows are pooled over the same reasoning steps and matched over the key-value heads with the squared error of \cref{eq:sats}, where MiniLM uses a KL divergence. The loss is denoted $\ell_{\mathrm{VR}}$. The term is used only together with the attention term: next to it on the three-stage prefix and in \method.

\subsection{Gradient-norm balancing}
\label{sec:balance}

The gradients of the token loss and of the alignment terms can differ in norm by orders of magnitude, and their ratio changes from step to step. On each replay step the gradients are computed in separate backward passes over the same micro-batches, $g_{\mathrm{tok}}$ from $\ltok$ and one gradient from each alignment term, and their norms are taken on the accumulated gradients before clipping (\cref{fig:replaystep}). For an auxiliary gradient $g_{\mathrm{aux}}$ and a target ratio $r$, the norm of the token part of the update divided by the norm of its auxiliary part, the update direction is
\begin{equation}
g =
\begin{cases}
g_{\mathrm{tok}} + \dfrac{\lVert g_{\mathrm{tok}} \rVert}{r \, \lVert g_{\mathrm{aux}} \rVert}\, g_{\mathrm{aux}}, & r \ge 1, \\[2.2ex]
r\, g_{\mathrm{tok}} + \dfrac{\lVert g_{\mathrm{tok}} \rVert}{\lVert g_{\mathrm{aux}} \rVert}\, g_{\mathrm{aux}}, & r < 1 .
\end{cases}
\label{eq:balance}
\end{equation}
For $r < 1$ the token gradient is scaled down instead of scaling the auxiliary gradient above the token norm. The auxiliary part is then the larger of the two and has the norm of the original token gradient. The update therefore grows less than it would with an auxiliary part of norm $\lVert g_{\mathrm{tok}} \rVert / r$. The sum still exceeds the token norm unless the two gradients point apart, with $\cos\varphi < -r/2$ for the angle $\varphi$ between them; at $r = 1$ it has $\sqrt{2 + 2\cos\varphi}$ times the token norm (\cref{sec:direction}). Fixing the ratio of gradient norms follows the aim of GradNorm, which tunes loss weights so that the gradient norms of several tasks stay balanced \citep{chen2018gradnorm}; the rule only rescales the gradients and neither removes conflicting components, as PCGrad does \citep{yu2020pcgrad}, nor solves for a common descent direction, as MGDA does \citep{sener2018mgda}. With one alignment term, $g_{\mathrm{aux}}$ is its gradient. With the three terms of \method, the gradients of the two pooled terms are first combined into one unit direction, and the unit hidden-state gradient is added with a weight $\rho$:
\begin{equation}
\mathbf{u} = \frac{\hat{g}_{\mathrm{SATS}} + \hat{g}_{\mathrm{VR}}}{\lVert \hat{g}_{\mathrm{SATS}} + \hat{g}_{\mathrm{VR}} \rVert} , \qquad
g_{\mathrm{aux}} = \mathbf{u} + \rho\, \hat{g}_{\mathrm{hid}} ,
\label{eq:joint}
\end{equation}
where $\hat{g} = g / \lVert g \rVert$. The two pooled terms then count equally whatever their raw norms, the hidden-state and pooled parts have norms in the ratio $\rho : 1$, and \cref{eq:balance} sets the norm of their sum relative to the token gradient.

\subsection{What is kept and what it costs}
\label{sec:cost}

At stage $k$, every replay configuration, with or without alignment terms, keeps the new expert $E_k$, the previous student $M_{k-1}$ and the replay prompts; none needs $M_{k-2}$ or an earlier expert. A recovery stage instead keeps every stage snapshot until it runs, $K$ models at once. Under the shared budget a stage of $n_k$ new-domain steps grows by $n_k/m$ replay steps. Under the fixed share it grows by $(k-1)\,b$, so the replay steps of $K$ stages add up to $b\,K(K-1)/2$, quadratic in $K$, while a recovery stage with a fixed number of steps per domain grows linearly in $K$. A replay step costs one forward and backward pass of the student for each loss term and the teacher's capture of the aligned states; new-domain steps cost the same as in sequential distillation.

\section{Experiments}
\label{sec:experiments}

\subsection{Setup}
\label{sec:setup}

\paragraph{Models.}
The student is DeepSeek-R1-Distill-Qwen-1.5B \citep{deepseek2025r1}. The five experts are public models trained from this same student. JustRL-DeepSeek-1.5B for math \citep{he2025justrl}, Archer2.0-Code-1.5B-Preview for code \citep{wang2025aspo,fatezero2025archer2card} and Nemotron-Research-Reasoning-Qwen-1.5B (v2) for logic \citep{liu2025prorl,nvidia2025nemotronv2card} were trained with reinforcement learning. STAR1-R1-Distill-1.5B for safety was trained with supervised fine-tuning on the 1K STAR-1 data \citep{wang2025star1}, and UltraData-IF-1.5B for instruction following with GRPO on the instruction-following subset of UltraData-SFT-2605 \citep{thinkingspace2026ultradataif}.

Nemotron serves as the logic expert because the logic prompts come from Reasoning Gym. Its model card lists prolonged reinforcement learning (3000 steps for v2) on math, code, STEM, logic puzzles and instruction following, with Reasoning Gym and IFEval-style data among its training sets.\footnote{\raggedright Hugging Face names: \nolinkurl{deepseek-ai/DeepSeek-R1-Distill-Qwen-1.5B}, \nolinkurl{hbx/JustRL-DeepSeek-1.5B}, \nolinkurl{Fate-Zero/Archer2.0-Code-1.5B-Preview}, \nolinkurl{nvidia/Nemotron-Research-Reasoning-Qwen-1.5B} (branch v2), \nolinkurl{UCSC-VLAA/STAR1-R1-Distill-1.5B}, \nolinkurl{Thinking-Space/UltraData-IF-1.5B} (revision be67d8b).} The first three domains form the same-family setting of Latent-MOPD \citep{fang2026latentmopd}, and the three-stage prefix of the sequence is reported separately as a short sequence.

The training code builds on the LastOPD implementation \citep{yang2026lastopd}, and the joint references in \cref{tab:three} are local reproductions (\cref{app:eval}).

\paragraph{Data and schedule.}
The prompts of the first three stages are those of the first 62 steps of the three-domain joint MOPD run of \cref{tab:three}. Math comes from DAPO-Math-17k \citep{yu2025dapo}, code from OpenCodeReasoning \citep{ahmad2025opencodereasoning} and logic from Reasoning Gym \citep{stojanovski2025reasoninggym}. The prompts are split by domain into 20 logic, 21 math and 21 code blocks of 32 prompts, and every block holds one domain. Stage 4 uses 672 safety prompts, the 336 harmful prompts of STAR-1 and 336 benign prompts of STAR-benign-915, both released with STAR-1 \citep{wang2025star1}; stage 5 uses 672 prompts drawn with seed 0 from the instruction-following subset of UltraData-SFT-2605, the training set of the instruction-following expert, after removing 4 that duplicate IFEval or IFBench prompts.

The domain order is logic, math, code, safety, instruction following, and every stage after the first has 21 new-domain blocks. The logic stage (20 steps from the student) is shared by all methods. The shared budget uses $m = 3$: every later stage has 7 replay steps, which cycle over the earlier domains. Stage 2 has 7 logic replay steps; stage 3 has 4 logic and 3 math; stage 4 has 3 logic, 2 math and 2 code; stage 5 has 2 logic, 2 math, 2 code and 1 safety. In total, logic receives 16 replay steps, math 7, code 4 and safety 1. The fixed share uses $b = 7$, so stages 3, 4 and 5 have 35, 42 and 49 steps (14, 21 and 28 replay steps); its stage 2 is identical to that of the shared budget (7 logic replay steps), so for each replay signal the budgets share their stage-2 model. Replay-KL (fixed share) uses the same fixed-share schedule with the token-level loss only, and its stage 2 is that of Replay-KL. With $b = 3$, stages 3, 4 and 5 have 27, 30 and 33 steps (6, 9 and 12 replay steps), after the same stage 2.

The recovery stage has 7 blocks per domain in turn, 35 steps after five stages, taught by the method's own stage-final students (the shared stage-1 model for logic) with the token-level loss only, and it starts from the last stage's model; recovery from the original experts uses the same blocks and routes each domain to its expert instead. Joint MOPD over the five domains starts from the student and trains on the 104 new-domain blocks of Seq in one pass, with the domains in turn, as in Latent-MOPD, and each block routed to its expert; joint2 trains on the same blocks twice, 208 steps. The replay buffer holds prompts only.

\paragraph{Training.}
Each step samples 4 responses for each of 32 prompts at temperature 1.0, with at most 16384 response tokens; the learning rate is $10^{-5}$, constant. Training uses verl \citep{sheng2025hybridflow} with vLLM rollouts \citep{kwon2023vllm} on four A100 80GB GPUs. Precision follows the LastOPD launcher (\cref{app:training}).

\paragraph{Implementation details.}
The token-level channel scores the student's $\kappa = 16$ most probable tokens, the value of the LastOPD and OPRD scripts. Replay-KL + SATS follows the original choice of one layer at the end of each encoder block \citep{qiu2023sats}: $\mathcal{L}$ holds the last layer of each quarter of the 28-layer student (layers 6, 13, 20 and 27, counted from 0), and all $H = 12$ query heads are matched. The queries are the last $W = 2000$ response tokens, the window of OPRD \citep{yang2026oprd}, and a reasoning step ends at a token whose text contains a blank line or at the end of the thinking block. The pooled rows are computed in query chunks with gradient checkpointing, so the full attention matrix is never stored (\cref{app:sats}). On the replay steps of Replay-KL + SATS, the token-level gradient is two to three orders of magnitude larger than the attention gradient before balancing. This configuration uses $r = 1$ in \cref{eq:balance}, the ratio with the highest nine-benchmark average in the sweep of \cref{app:ratio} on the three-stage prefix.

Replay-KL + OPRD uses the hidden-state term on all 28 layers over the same window, with $r = 1$. \method uses all 28 layers for all three terms ($\mathcal{L}_h$ and $\mathcal{L}$), $W = 2000$, $\rho = 2$, $r = 1$ and $b = 7$. The teacher's targets of all three terms are rounded to bf16, as are the hidden-state targets of Replay-KL + OPRD. The hidden states are the outputs of the 28 decoder layers as the model returns them, the last one after the final normalization. The values $\rho = 2$ and $m = 3$ were fixed before the runs, and $b = 7$ makes stage 2 the same under both budgets. Before balancing, the token-level gradient exceeds the raw gradients of the attention and value-relation terms by about one to three orders of magnitude and that of the hidden-state term by five to six and a half (\cref{app:sats}).

\paragraph{Evaluation.}
Fourteen benchmarks cover the five domains: BBH \citep{suzgun2023bbh}, Reasoning Gym (GYM) \citep{stojanovski2025reasoninggym} and MuSR \citep{sprague2024musr} for logic; AIME 2024 and 2025\footnote{The 30 problems of each year as packaged with the OPRD evaluation data.} and Minerva \citep{lewkowycz2022minerva} for math; LiveCodeBench easy problems (LCB-e) \citep{jain2025livecodebench}, MBPP \citep{austin2021mbpp} and MBPP+ \citep{liu2023evalplus} for code; StrongREJECT (313 prompts) \citep{souly2024strongreject}, the adversarial harmful prompts of WildJailbreak (250, sampled with seed 42 from the walledai mirror) \citep{jiang2024wildteaming} and XSTest (250 safe and 200 unsafe prompts) \citep{rottger2024xstest} for safety; IFEval (541) \citep{zhou2023ifeval} and IFBench (300) \citep{pyatkin2025ifbench} for instruction following. MBPP and MBPP+ are the same 378 problems with the original and the extended tests, so these problems count twice in the average. Counting them once changes the label (gain, loss or level, defined below) of 3 of the 171 pairs among the 19 rows of \cref{tab:five} used to estimate the evaluation noise (\cref{app:eval}). No GYM, safety or IFEval evaluation prompt is identical to or contained in a training prompt (\cref{app:eval}). The three-stage prefix is evaluated on the first nine benchmarks (logic, math and code).

The safety and instruction-following benchmarks sample 4 responses per prompt at temperature 0.6 and top-p 0.95, with at most 8192 tokens for safety and 16384 for instruction following. The other settings are listed in \cref{app:eval}. Safety is judged by Qwen3Guard-Gen-8B in response moderation mode \citep{zhao2025qwen3guard}, which reads the answer after the thinking block, or the whole response when the reasoning does not close. On harmful prompts the score is the safe rate, with Controversial counted as unsafe; StrongREJECT therefore denotes the Qwen3Guard safe rate on the StrongREJECT prompts, not the rubric score of its own evaluator. On the safe prompts of XSTest the score is the compliance rate, the share of responses that close their reasoning and are judged not to refuse, so over-refusal lowers the safety average. The safe rate on the unsafe prompts of XSTest is reported separately. Instruction following is scored by the verifiers of the IFBench repository (the 25 instruction types of IFEval and the 58 of IFBench) on the answer after the thinking block, as prompt-level strict accuracy. In \cref{eq:norm} the upper anchor of each benchmark is the best of the experts in the sequence.

Each configuration is trained with the preset rollout seed 0 and evaluated at its final checkpoint with the evaluation seeds 0, 1 and 2 unless a table or its caption states otherwise. Every score is the mean over these seeds, and differences between configurations are computed before rounding. Four five-stage configurations were also trained with rollout seed 1 (\cref{tab:trainseeds}), and two configurations of the three-stage prefix (\cref{sec:q-signal}).

Evaluating the same checkpoint with another evaluation seed changes its score. This evaluation noise is measured as the standard deviation across evaluation seeds, pooled over checkpoints with three evaluation seeds. For the five-stage Avg it is 0.28 points with 38 degrees of freedom. Its pool holds the trained configurations with rollout seed 0, the uniform average of the five experts and the TIES merge of the five experts. For the three-stage Avg it is 0.32 points with 26 degrees of freedom, pooled over the continual configurations of the three-stage prefix. A difference is described as a gain or a loss when its 95\% interval under this standard deviation, with the $t$ quantile, excludes zero, and as level or within the evaluation noise otherwise; between two configurations with three seeds each, the half-width of this interval is 0.46 points on the five-stage Avg and 0.54 on the three-stage Avg, and 0.66 on the five-stage Avg when one of them has seed 0 only. The evaluation seeds show no common effect across checkpoints ($F(2, 36) = 0.46$ on the five-stage Avg and $F(2, 24) = 0.00$ on the three-stage prefix), so differences are not paired by seed. These intervals and labels cover the variation between evaluation seeds. The two training runs of the four configurations of \cref{tab:trainseeds} differ by 0.06 to 0.75 points in Avg. A comparison of means over the two runs uses the variance between training runs of a configuration's three-seed mean, estimated from these four pairs (4 degrees of freedom). Between two means over the two runs, the half-width is then 1.03 points in Avg and 0.82 in Avg$_{\mathrm{bal}}$, the average with balanced safety (\cref{tab:five}). Between such a mean and a configuration trained once, it is 1.26 points in Avg. The half-widths of the domain averages are in \cref{app:eval}, and the Avg of every evaluation seed is listed in \cref{app:full,tab:trainseeds}.

\paragraph{Baselines and references.}
\label{sec:baselines}
Seq trains the stages without replay. Seq + Recovery follows Seq with one multi-teacher on-policy distillation stage, in which the prompts of each domain are routed to that domain's stage-final student ($M_1, \dots, M_K$), 7 blocks per domain in turn: 21 steps after three stages and 35 after five. It is the multi-domain form of the cross-stage distillation of GLM-5 \citep[\S3.5]{glm2026glm5} and of the recovery stage in the ACLArena text \citep{wang2026aclarena}. TS-OPD \citep{guo2026tsopd} uses the same two kinds of teacher but starts from the model before specialization. With two domains, this design matches a stage of replay toward the previous student rather than recovery. The recovery stage starts from $M_K$.

Replay-KL uses the shared budget with the token-level loss toward $M_{k-1}$ only. Replay-KL + SATS adds the attention term and Replay-KL + OPRD the hidden-state term of OPRD, each balanced to the token-level gradient. \method (shared budget) uses the three alignment terms and their balancing under the shared budget of Replay-KL, and Replay-KL (fixed share) uses the fixed share with the token-level loss only. X + Recovery applies the recovery stage after a replay method X, with X's own stage-final students as teachers, and X + expert recovery routes the same blocks to the five original experts. Replay-KL + SEEKR, run on the three-stage prefix, adds the per-head attention KL of SEEKR \citep{he2024seekr} on 42 of the 336 heads, selected after each stage by task sensitivity times forgettability, on 100 random query rows per response, balanced 1:1 like Replay-KL + SATS; \cref{app:baselines} lists the changes needed without gold answers.

Joint MOPD and Joint Latent-MOPD train on the three domains of the prefix at once with all three experts, and joint MOPD over the five domains and its two-pass version joint2 train on all five at once; they are references, not continual methods.

\subsection{Q1: Replay toward the previous student}
\label{sec:q-replay}

Replay toward the previous student raises the average over sequential distillation. After five stages Replay-KL ends 3.80 points above Seq in Avg and 0.161 in \Norm (\cref{tab:matrix,tab:five}), with gains on logic (+3.19), math (+2.66), code (+2.55) and safety (+9.36) and none on the newest domain, instruction following ($-0.05$). After the three-stage prefix the difference is 2.34 points in Avg and 0.146 in \Norm (\cref{tab:three}). Most of it is on logic: the logic average is 5.12 points higher, which contributes 1.71 of the 2.34 points.

\begin{figure}[t]
\centering
\IfFileExists{figures/musr.pdf}{\includegraphics{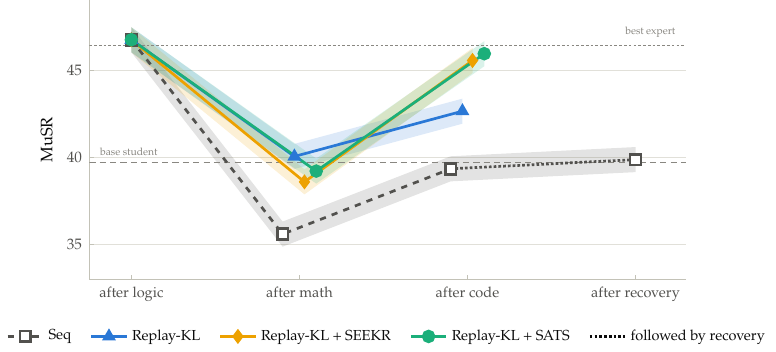}}{\begin{tikzpicture}
\begin{axis}[vizaxis, width=\columnwidth, height=6.3cm, xmin=0.75, xmax=4.3, ymin=33, ymax=49, xtick={1,2,3,4}, xticklabels={after logic, after math, after code, after recovery}, ylabel={MuSR}, legend style={at={(0.5,-0.15)}, anchor=north, legend columns=5, /tikz/every even column/.append style={column sep=7pt}}]
\addplot[vizMuted, densely dashed, line width=0.6pt, domain=0.75:4.3, samples=2, forget plot] {39.68};
\addplot[vizMuted, densely dotted, line width=0.6pt, domain=0.75:4.3, samples=2, forget plot] {46.43};
\node[font=\tiny, text=vizMuted, anchor=south west] at (axis cs:0.77,39.68) {base student};
\node[font=\tiny, text=vizMuted, anchor=south east] at (axis cs:4.28,46.43) {best expert};
\addplot[name path=u18, draw=none, forget plot] coordinates {(1,47.46) (1.9,36.30) (2.9,40.05)};
\addplot[name path=l18, draw=none, forget plot] coordinates {(1,46.02) (1.9,34.86) (2.9,38.61)};
\addplot[mSeq, fill opacity=0.16, draw=none, forget plot] fill between[of=u18 and l18];
\addplot[mSeq, dashed, mark=square*, mark options={solid, fill=white}, mark size=2.6pt] coordinates {(1,46.74) (1.9,35.58) (2.9,39.33)};
\addlegendentry{Seq}
\addplot[name path=u19, draw=none, forget plot] coordinates {(2.9,40.05) (4,40.58)};
\addplot[name path=l19, draw=none, forget plot] coordinates {(2.9,38.61) (4,39.14)};
\addplot[mSeq, fill opacity=0.16, draw=none, forget plot] fill between[of=u19 and l19];
\addplot[mSeq, densely dotted, mark=square*, mark options={solid, fill=white}, mark size=2.6pt, forget plot] coordinates {(2.9,39.33) (4,39.86)};
\addplot[name path=u20, draw=none, forget plot] coordinates {(1,47.46) (1.97,40.76) (2.97,43.36)};
\addplot[name path=l20, draw=none, forget plot] coordinates {(1,46.02) (1.97,39.32) (2.97,41.92)};
\addplot[mKL, fill opacity=0.16, draw=none, forget plot] fill between[of=u20 and l20];
\addplot[mKL, solid, mark=triangle*, mark options={solid}, mark size=2.6pt] coordinates {(1,46.74) (1.97,40.04) (2.97,42.64)};
\addlegendentry{Replay-KL}
\addplot[name path=u21, draw=none, forget plot] coordinates {(1,47.46) (2.03,39.30) (3.03,46.27)};
\addplot[name path=l21, draw=none, forget plot] coordinates {(1,46.02) (2.03,37.86) (3.03,44.83)};
\addplot[mSEEKR, fill opacity=0.16, draw=none, forget plot] fill between[of=u21 and l21];
\addplot[mSEEKR, solid, mark=diamond*, mark options={solid}, mark size=2.6pt] coordinates {(1,46.74) (2.03,38.58) (3.03,45.55)};
\addlegendentry{Replay-KL + SEEKR}
\addplot[name path=u22, draw=none, forget plot] coordinates {(1,47.46) (2.1,39.92) (3.1,46.66)};
\addplot[name path=l22, draw=none, forget plot] coordinates {(1,46.02) (2.1,38.48) (3.1,45.22)};
\addplot[mSATS, fill opacity=0.16, draw=none, forget plot] fill between[of=u22 and l22];
\addplot[mSATS, solid, mark=*, mark options={solid}, mark size=2.6pt] coordinates {(1,46.74) (2.1,39.2) (3.1,45.94)};
\addlegendentry{Replay-KL + SATS}
\addlegendimage{vizInk, densely dotted, line width=1.3pt}
\addlegendentry{followed by recovery}
\end{axis}
\end{tikzpicture}}
\caption{MuSR after each stage of the three-stage prefix; means over three evaluation seeds, with the pooled standard error of each point as shaded bands (\cref{app:eval}). The stage-1 model is shared by all methods; the dotted segment is the recovery stage after Seq (Seq + Recovery). Thin gray lines: base student and best expert. Scores use the benchmark's parser, which reads only the last line containing ``answer:'' (\cref{app:eval}); after the math stage the student ends most MuSR responses with a boxed answer instead of that line (\cref{sec:discussion}).}
\label{fig:musr}
\end{figure}

\paragraph{Where sequential distillation forgets.}
\label{sec:diagnosis}
After the logic stage the student scores 54.14 on BBH, 34.25 on Reasoning Gym (GYM) and 46.74 on MuSR. Learning math without replay raises BBH to 63.15 and GYM to 43.24, close to the logic expert (64.17 and 43.83). These two gains coincide with a switch of answer format in the math stage. The student moves from the \texttt{<answer>} tags that the logic prompts ask for to the boxed answers that BBH and GYM read. Read with these tags, the stage-1 model already scores 61.20 and 40.08 (\cref{app:eval}). Under the benchmark's parser MuSR moves the other way and falls to 35.58, a loss of 11.16 points (\cref{fig:musr}). The code stage then causes the largest single drop of the three-stage prefix: GYM falls from 43.24 to 30.55, while MuSR partly returns to 39.33. Forgetting is therefore concentrated: under boxed-only scoring BBH ends above its post-logic value, and the losses sit on GYM during the code stage and, under the benchmark's parser, on MuSR throughout. A final recovery stage repairs only part of it: Seq + Recovery brings GYM back to 41.30 and BBH to 62.07, but MuSR stays at 39.86 under the parser. Over the two later stages the losses grow. In the safety stage, the average of Seq over the first nine benchmarks falls from 49.10 to 39.52. Seq's BBH, scored on the boxed answer only, falls from 60.31 to 31.85. The shared-budget replay methods end the stage at 48.20 to 49.20 on the same nine benchmarks (\cref{tab:stages}). After the instruction-following stage, Seq returns to 45.51 and the replay methods stay within 0.6 points of their stage-4 values.

\paragraph{Backward transfer.}
\label{sec:bwt}
\Cref{tab:bwt} separates forgetting from transfer on the three-stage prefix. Under boxed-only scoring BBH rises in every method, and GYM rises in every method except Seq; both rise during the math stage, mainly through the switch to boxed answers (\cref{app:eval}). MuSR is the one benchmark on which every method ends lower under its parser. Replay cuts the loss from 7.41 points to 4.10, and Replay-KL + SATS cuts it to 0.79. This loss and the 1.19 of Replay-KL + SEEKR are within the noise of MuSR (half-width 2.09). Math changes little after its own stage: its backward transfer lies between $-0.92$ and +3.54. \Cref{fig:musr} shows when the MuSR gap between Replay-KL + SATS and Replay-KL opens. After the math stage the two are close (Replay-KL + SATS 39.20, Replay-KL 40.04). Under the benchmark's parser, replay already blocks about 40\% of the stage-2 drop ($-6.70$ for Replay-KL against $-11.16$ for Seq). The gap appears during the code stage, where Replay-KL + SATS gains 6.75 points on MuSR and Replay-KL 2.60.

\begin{table}[t]
\centering
\footnotesize
\setlength{\tabcolsep}{2.2pt}
\begin{tabular}{l ccc ccc}
\toprule
& \multicolumn{3}{c}{Logic} & \multicolumn{3}{c}{Math} \\
\cmidrule(lr){2-4} \cmidrule(lr){5-7}
Method & BBH & GYM & MuSR & AIME24 & AIME25 & Minerva \\
\midrule
Seq & +6.17 & $-$3.69 & $-$7.41 & +3.40 & $-$0.56 & $-$0.92 \\
Seq + Recovery & +7.93 & +7.05 & $-$6.88 & +1.74 & \textbf{+1.81} & +0.12 \\
Replay-KL & +7.87 & +6.66 & $-$4.10 & \textbf{+3.54} & +1.53 & $-$0.40 \\
Replay-KL + SEEKR & +9.20 & \textbf{+7.49} & $-$1.19 & $-$0.21 & +0.49 & \textbf{+0.40} \\
Replay-KL + SATS & \textbf{+9.32} & +6.91 & \textbf{$-$0.79} & +2.36 & +0.42 & +0.03 \\
\bottomrule
\end{tabular}
\caption{Backward transfer (\cref{eq:bwt}) after the third stage, and for Seq + Recovery after its recovery stage. Values are means over three evaluation seeds of the score after that stage minus the score after the stage that learned the domain. Logic is measured from the shared stage-1 model (BBH 54.14, GYM 34.25, MuSR 46.74), math from each method's stage-2 model (Seq's for Seq + Recovery; \cref{tab:stages}). Negative values are forgetting; on MuSR they mainly measure whether a response has a line with ``answer:'' that names an option, the only line the benchmark's parser reads (\cref{app:eval}). On BBH and GYM the stage-1 reference is lowered by answers given in \texttt{<answer>} tags, the format of the logic training prompts, which these two benchmarks do not read: read with the tags the stage-1 model scores 61.20 and 40.08, so the positive values in these columns mainly reflect the switch to boxed answers during the math stage (\cref{app:eval}). Bold marks the highest value in each column.}
\label{tab:bwt}
\end{table}

\subsection{Q2: How much to replay}
\label{sec:q-budget}

\begin{table*}[t]
\centering
\small
\setlength{\tabcolsep}{4pt}
\begin{tabular}{l l c c c c}
\toprule
Replay budget (blocks) & Recovery & Token KL & + SATS & + OPRD & All three terms \\
\midrule
None (Seq) & no & \cellcolor{vizBlue!6}42.87 & n/a & n/a & n/a \\
None & snapshots & \cellcolor{vizBlue!50}48.64 & n/a & n/a & n/a \\
None & snapshots, 70 steps & \cellcolor{vizBlue!48}48.35 & n/a & n/a & n/a \\
\midrule
Shared (28) & no & \cellcolor{vizBlue!35}46.66 & \cellcolor{vizBlue!36}46.80 \dlt{+0.14} & \cellcolor{vizBlue!43}47.71 \dlt{+1.04} & \cellcolor{vizBlue!43}47.62 \dlt{+0.96} \\
Shared (28) & snapshots & \cellcolor{vizBlue!54}\textbf{49.14} & \cellcolor{vizBlue!54}\textbf{49.15} \dlt{+0.02} & \cellcolor{vizBlue!55}\textbf{49.24} \dlt{+0.11} & \cellcolor{vizBlue!54}49.07 \dlt{$-$0.07} \\
\midrule
Fixed, $b = 3$ (34) & no & \cellcolor{vizBlue!46}48.12 & & & \cellcolor{vizBlue!48}48.36 \dlt{+0.25} \\
Fixed, $b = 7$ (70) & no & \cellcolor{vizBlue!53}48.95 & & & \cellcolor{vizBlue!52}48.91 \dlt{$-$0.05} \\
Fixed, $b = 7$ (70) & snapshots & & & & \cellcolor{vizBlue!55}\textbf{49.19} \\
\midrule
\multicolumn{6}{l}{\textit{No training, snapshot-free: running merges of the experts}} \\
\multicolumn{2}{l}{Uniform average} & 45.07 & & & \\
\multicolumn{2}{l}{TIES, stage by stage} & 50.42 & & & \\
\midrule
\multicolumn{6}{l}{\textit{No training, snapshots or experts kept to the end (S2)}} \\
\multicolumn{2}{l}{TIES of the Seq snapshots} & 47.74 & & & \\
\multicolumn{2}{l}{TIES merge of the five experts} & 52.29 & & & \\
\midrule
\multicolumn{6}{l}{\textit{Not continual: every expert at the same time}} \\
\multicolumn{2}{l}{Joint MOPD, 104 steps, one pass} & 48.68$^{\dagger}$ & & & \\
\multicolumn{2}{l}{Joint MOPD, 208 steps, two passes (joint2)} & 49.02 & & & \\
\bottomrule
\end{tabular}
\caption{The design space of continual MOPD after five stages: 14-benchmark Avg (\cref{eq:avg}), mean over three evaluation seeds. Rows are the replay budget, with its replay blocks summed over stages 2 to 5, and whether a final recovery stage taught by the stage snapshots follows; columns are the signal on replay steps, and the last column holds the three terms of \method, so its fixed-share rows are \method and \method + Recovery. In parentheses: the difference to the token-KL cell of the same row. Cells of continual configurations are shaded by their Avg, from Seq (42.87, lightest) to the highest (49.24, darkest); the Avg of a merge or of joint MOPD is printed in the first score column. With $b = 3$ the fixed share starts at stage 3, after the stage 2 of the shared budget. n/a: no replay steps, so no alignment term applies. The 70-step recovery uses 14 blocks per domain, so that Seq and recovery train 174 steps, as many as the fixed share with $b = 7$. Joint MOPD uses the token-level loss toward the five experts. The merges are references without training: the uniform average and TIES applied stage by stage, which also keeps the base model, fit the snapshot-free setting, TIES of the Seq snapshots needs every stage snapshot, and TIES of the five experts needs the experts at the end, like recovery from the original experts (\cref{tab:five}). $^{\dagger}$: evaluation seed 0 only. Domain averages, the average over the 11 benchmarks outside safety, the average with balanced safety and \Norm are in \cref{tab:five}. Bold marks the highest trained configuration in each column.}
\label{tab:matrix}
\end{table*}

\begin{table*}[tp]
\centering
\small
\setlength{\tabcolsep}{3.5pt}
\begin{tabular}{l ccccc c c c c c}
\toprule
& \multicolumn{5}{c}{Domain average, in learning order} & & & & & \\
\cmidrule(lr){2-6}
Model & Logic & Math & Code & Safety & IF & Avg & Avg$_{11}$ & Avg$_{\mathrm{bal}}$ & \Norm & \NormD \\
\midrule
Student & 33.73 & 26.17 & 52.80 & 38.89 & 22.69 & 35.72 & 34.86 & 38.43 & 0.000 & 0.000 \\
Nemotron (logic expert) & 51.47 & 40.50 & 59.70 & 47.37 & 43.68 & 48.90 & 49.31 & 51.47 & 0.558 & 0.913 \\
JustRL (math expert) & 39.54 & 40.77 & 61.72 & 43.74 & 27.07 & 43.68 & 43.66 & 46.58 & 0.337 & 0.506 \\
Archer2.0 (code expert) & 36.36 & 33.23 & 66.79 & 42.36 & 26.05 & 42.02 & 41.93 & 44.80 & 0.267 & 0.411 \\
STAR1 (safety expert) & 29.65 & 25.90 & 50.70 & 62.66 & 16.19 & 38.51 & 31.92 & 36.50 & 0.118 & 0.013 \\
UltraData-IF (IF expert) & 34.68 & 27.61 & 55.39 & 41.07 & 34.79 & 38.99 & 38.42 & 41.79 & 0.138 & 0.286 \\
\midrule
\multicolumn{11}{l}{\textit{No replay}} \\
Seq & 42.70 & 32.66 & 61.16 & 39.30 & 36.34 & 42.87 & 43.84 & 45.84 & 0.303 & 0.539 \\
Seq + Recovery & 45.96 & 37.30 & 66.54 & 53.58 & 35.39 & \textbf{48.64} & \textbf{47.29} & \textbf{46.89} & \textbf{0.547} & \textbf{0.820} \\
Seq + Recovery-70 & 46.12 & 37.22 & 66.23 & 52.91 & 34.71 & 48.35 & 47.10 & 46.23 & 0.535 & 0.800 \\
\midrule
\multicolumn{11}{l}{\textit{Shared budget}} \\
Replay-KL & 45.89 & 35.32 & 63.71 & 48.66 & 36.29 & 46.66 & 46.12 & 48.51 & 0.464 & 0.725 \\
Replay-KL + SATS & 45.08 & 35.49 & 64.43 & 49.98 & 35.14 & 46.80 & 45.94 & 48.46 & 0.470 & 0.721 \\
Replay-KL + OPRD & 46.07 & 36.13 & 65.33 & 50.74 & 36.56 & \textbf{47.71} & \textbf{46.88} & 48.79 & \textbf{0.508} & \textbf{0.784} \\
\method (shared budget) & 45.03 & 36.09 & 66.27 & 50.66 & 36.29 & 47.62 & 46.80 & \textbf{49.00} & 0.504 & 0.780 \\
\midrule
\multicolumn{11}{l}{\textit{Shared budget, with a final recovery stage (keeps $M_1, \dots, M_5$)}} \\
Replay-KL + Recovery & 46.61 & 38.81 & 66.35 & 54.00 & 35.28 & 49.14 & 47.81 & 47.20 & 0.569 & 0.847 \\
Replay-KL + SATS + Recovery & 47.83 & 38.09 & 65.84 & 54.08 & 35.28 & 49.15 & 47.81 & 47.14 & 0.569 & 0.845 \\
Replay-KL + OPRD + Recovery & 46.95 & 38.07 & 67.62 & 53.92 & 34.86 & \textbf{49.24} & \textbf{47.97} & \textbf{47.24} & \textbf{0.573} & \textbf{0.851} \\
\method (shared budget) + Recovery & 45.62 & 37.03 & 68.00 & 54.98 & 35.03 & 49.07 & 47.45 & 46.98 & 0.566 & 0.839 \\
\midrule
\multicolumn{11}{l}{\textit{Fixed share, $b = 3$}} \\
Replay-KL, $b = 3$ & 45.59 & 38.04 & 65.32 & 51.75 & 35.76 & 48.12 & 47.13 & \textbf{47.64} & 0.525 & 0.800 \\
\method, $b = 3$ & 44.94 & 38.16 & 67.01 & 52.01 & 35.36 & \textbf{48.36} & \textbf{47.37} & 47.27 & \textbf{0.536} & \textbf{0.814} \\
\midrule
\multicolumn{11}{l}{\textit{Fixed share, $b = 7$}} \\
Replay-KL (fixed share) & 45.82 & 38.37 & 67.52 & 53.34 & 35.10 & 48.95 & 47.76 & 47.53 & 0.561 & 0.840 \\
\method & 47.55 & 38.17 & 66.54 & 52.27 & 35.54 & 48.91 & 47.99 & 47.52 & 0.559 & 0.842 \\
\method + Recovery & 47.01 & 37.63 & 68.49 & 53.52 & 34.35 & \textbf{49.19} & \textbf{48.01} & \textbf{47.57} & \textbf{0.571} & \textbf{0.847} \\
\midrule
\multicolumn{11}{l}{\textit{No training, snapshot-free: running merges of the experts}} \\
Uniform average of the five experts & 41.63 & 35.25 & 61.94 & 50.19 & 32.01 & 45.07 & 43.68 & 47.20 & 0.396 & 0.593 \\
TIES, stage by stage & 52.46 & 42.80 & 68.67 & 44.69 & 40.01 & 50.42 & 51.98 & 53.24 & 0.623 & 1.001 \\
\midrule
\multicolumn{11}{l}{\textit{No training, snapshots or experts kept to the end (S2)}} \\
TIES of the Seq snapshots & 50.05 & 37.10 & 66.63 & 48.59 & 30.61 & 47.74 & 47.51 & 49.99 & 0.509 & 0.744 \\
TIES merge of the five experts & 52.57 & 42.65 & 68.05 & 54.62 & 39.17 & 52.29 & 51.65 & 53.97 & 0.702 & 1.061 \\
\midrule
\multicolumn{11}{l}{\textit{Not continual: every expert at the same time}} \\
Joint MOPD, 208 steps (joint2) & 46.28 & 38.24 & 65.79 & 55.32 & 34.70 & 49.02 & 47.30 & 47.60 & 0.564 & 0.829 \\
\bottomrule
\end{tabular}
\caption{Scores after the fifth stage on the 14 benchmarks of the five domains. Domain columns are domain averages, Avg is the mean of the 14 benchmark scores (\cref{eq:avg}), Avg$_{11}$ the mean of the 11 benchmarks outside safety, Avg$_{\mathrm{bal}}$ the 14-benchmark mean with the three safety scores replaced by three copies of a balanced safety score that weights refusal of harmful prompts and compliance with safe prompts equally ($S_{\mathrm{bal1}}$, \cref{app:balanced}), and \Norm and \NormD are the normalized scores of \cref{sec:metrics}. Student and expert rows are single evaluations; the other rows are means over three evaluation seeds. The safety average includes the compliance rate on the safe prompts of XSTest, so over-refusal lowers it, while its other two benchmarks reward refusal. Joint2 trains on the 104 blocks of Seq twice. Seq + Recovery-70 has a 70-step recovery stage (\cref{sec:q-budget}). The merges are not trained. The uniform average of the experts can be kept as a running average. TIES \citep{yadav2023ties} applied stage by stage needs only the previous merge, the expert of the stage and the base model. Both therefore fit snapshot-free continual learning. TIES of the Seq snapshots needs every snapshot, and the TIES merge of the five experts needs the experts at the end. TIES keeps the largest 20\% of each task vector and scales the merged task vector by $\lambda = 1$; both values were set before any evaluation (\cref{app:baselines}). Bold marks the highest Avg, Avg$_{11}$, Avg$_{\mathrm{bal}}$, \Norm and \NormD in each group of continual rows; the rows with + Recovery keep every stage snapshot (S2). Configurations with one evaluation seed and the per-benchmark scores are in \cref{app:full}.}\label{tab:five}
\end{table*}

\begin{figure}[t]
\centering
\IfFileExists{figures/oldnine.pdf}{\includegraphics{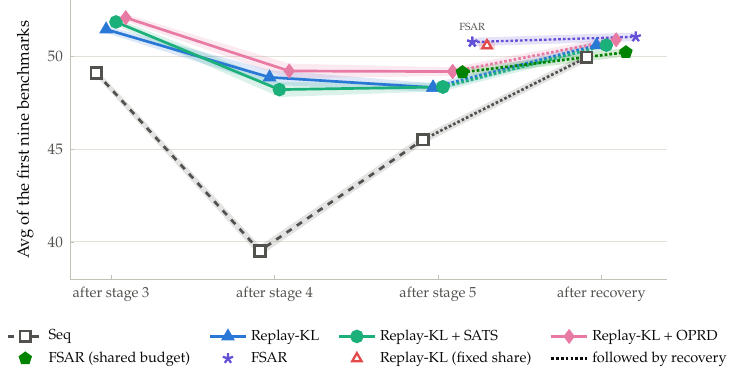}}{\begin{tikzpicture}
\begin{axis}[vizaxis, width=\columnwidth, height=6.3cm, xmin=0.75, xmax=4.4, ymin=38, ymax=53, xtick={1,2,3,4}, xticklabels={after stage 3, after stage 4, after stage 5, after recovery}, ylabel={Avg of the first nine benchmarks}, legend style={at={(0.5,-0.15)}, anchor=north, legend columns=4, /tikz/every even column/.append style={column sep=7pt}}]
\addplot[name path=u23, draw=none, forget plot] coordinates {(0.91,49.29) (1.91,39.80) (2.91,45.74)};
\addplot[name path=l23, draw=none, forget plot] coordinates {(0.91,48.91) (1.91,39.24) (2.91,45.28)};
\addplot[mSeq, fill opacity=0.16, draw=none, forget plot] fill between[of=u23 and l23];
\addplot[mSeq, dashed, mark=square*, mark options={solid, fill=white}, mark size=2.6pt] coordinates {(0.91,49.1) (1.91,39.52) (2.91,45.51)};
\addlegendentry{Seq}
\addplot[name path=u24, draw=none, forget plot] coordinates {(2.91,45.74) (3.91,50.17)};
\addplot[name path=l24, draw=none, forget plot] coordinates {(2.91,45.28) (3.91,49.71)};
\addplot[mSeq, fill opacity=0.16, draw=none, forget plot] fill between[of=u24 and l24];
\addplot[mSeq, densely dotted, mark=square*, mark options={solid, fill=white}, mark size=2.6pt, forget plot] coordinates {(2.91,45.51) (3.91,49.94)};
\addplot[name path=u25, draw=none, forget plot] coordinates {(0.97,51.63) (1.97,49.26) (2.97,48.54)};
\addplot[name path=l25, draw=none, forget plot] coordinates {(0.97,51.25) (1.97,48.46) (2.97,48.08)};
\addplot[mKL, fill opacity=0.16, draw=none, forget plot] fill between[of=u25 and l25];
\addplot[mKL, solid, mark=triangle*, mark options={solid}, mark size=2.6pt] coordinates {(0.97,51.44) (1.97,48.86) (2.97,48.31)};
\addlegendentry{Replay-KL}
\addplot[name path=u26, draw=none, forget plot] coordinates {(2.97,48.54) (3.97,50.82)};
\addplot[name path=l26, draw=none, forget plot] coordinates {(2.97,48.08) (3.97,50.36)};
\addplot[mKL, fill opacity=0.16, draw=none, forget plot] fill between[of=u26 and l26];
\addplot[mKL, densely dotted, mark=triangle*, mark options={solid}, mark size=2.6pt, forget plot] coordinates {(2.97,48.31) (3.97,50.59)};
\addplot[name path=u27, draw=none, forget plot] coordinates {(1.03,52.03) (2.03,48.60) (3.03,48.57)};
\addplot[name path=l27, draw=none, forget plot] coordinates {(1.03,51.65) (2.03,47.80) (3.03,48.11)};
\addplot[mSATS, fill opacity=0.16, draw=none, forget plot] fill between[of=u27 and l27];
\addplot[mSATS, solid, mark=*, mark options={solid}, mark size=2.6pt] coordinates {(1.03,51.84) (2.03,48.2) (3.03,48.34)};
\addlegendentry{Replay-KL + SATS}
\addplot[name path=u28, draw=none, forget plot] coordinates {(3.03,48.57) (4.03,50.82)};
\addplot[name path=l28, draw=none, forget plot] coordinates {(3.03,48.11) (4.03,50.36)};
\addplot[mSATS, fill opacity=0.16, draw=none, forget plot] fill between[of=u28 and l28];
\addplot[mSATS, densely dotted, mark=*, mark options={solid}, mark size=2.6pt, forget plot] coordinates {(3.03,48.34) (4.03,50.59)};
\addplot[name path=u29, draw=none, forget plot] coordinates {(1.09,52.26) (2.09,49.60) (3.09,49.40)};
\addplot[name path=l29, draw=none, forget plot] coordinates {(1.09,51.88) (2.09,48.80) (3.09,48.94)};
\addplot[mOPRD, fill opacity=0.16, draw=none, forget plot] fill between[of=u29 and l29];
\addplot[mOPRD, solid, mark=diamond*, mark options={solid}, mark size=2.6pt] coordinates {(1.09,52.07) (2.09,49.2) (3.09,49.17)};
\addlegendentry{Replay-KL + OPRD}
\addplot[name path=u30, draw=none, forget plot] coordinates {(3.09,49.40) (4.09,51.11)};
\addplot[name path=l30, draw=none, forget plot] coordinates {(3.09,48.94) (4.09,50.65)};
\addplot[mOPRD, fill opacity=0.16, draw=none, forget plot] fill between[of=u30 and l30];
\addplot[mOPRD, densely dotted, mark=diamond*, mark options={solid}, mark size=2.6pt, forget plot] coordinates {(3.09,49.17) (4.09,50.88)};
\addplot[name path=u31, draw=none, forget plot] coordinates {(3.15,49.36)};
\addplot[name path=l31, draw=none, forget plot] coordinates {(3.15,48.90)};
\addplot[mFSARs, fill opacity=0.16, draw=none, forget plot] fill between[of=u31 and l31];
\addplot[mFSARs, only marks, mark=pentagon*, mark options={solid}, mark size=2.6pt] coordinates {(3.15,49.13)};
\addlegendentry{\method (shared budget)}
\addplot[name path=u32, draw=none, forget plot] coordinates {(3.15,49.36) (4.15,50.44)};
\addplot[name path=l32, draw=none, forget plot] coordinates {(3.15,48.90) (4.15,49.98)};
\addplot[mFSARs, fill opacity=0.16, draw=none, forget plot] fill between[of=u32 and l32];
\addplot[mFSARs, densely dotted, mark=pentagon*, mark options={solid}, mark size=2.6pt, forget plot] coordinates {(3.15,49.13) (4.15,50.21)};
\addplot[name path=u33, draw=none, forget plot] coordinates {(3.21,50.99)};
\addplot[name path=l33, draw=none, forget plot] coordinates {(3.21,50.53)};
\addplot[mFSAR, fill opacity=0.16, draw=none, forget plot] fill between[of=u33 and l33];
\addplot[mFSAR, only marks, mark=star, mark options={solid}, mark size=2.6pt, line width=1.1pt] coordinates {(3.21,50.76)};
\addlegendentry{\method}
\addplot[name path=u34, draw=none, forget plot] coordinates {(3.21,50.99) (4.21,51.27)};
\addplot[name path=l34, draw=none, forget plot] coordinates {(3.21,50.53) (4.21,50.81)};
\addplot[mFSAR, fill opacity=0.16, draw=none, forget plot] fill between[of=u34 and l34];
\addplot[mFSAR, densely dotted, mark=star, mark options={solid}, mark size=2.6pt, line width=1.1pt, forget plot] coordinates {(3.21,50.76) (4.21,51.04)};
\addplot[name path=u35, draw=none, forget plot] coordinates {(3.25,50.80) (3.35,50.80)};
\addplot[name path=l35, draw=none, forget plot] coordinates {(3.25,50.34) (3.35,50.34)};
\addplot[mKLf, fill opacity=0.16, draw=none, forget plot] fill between[of=u35 and l35];
\addplot[mKLf, only marks, mark=triangle*, mark options={solid, fill=white}, mark size=2.6pt, line width=1.1pt] coordinates {(3.3,50.57)};
\addlegendentry{Replay-KL (fixed share)}
\node[font=\tiny, text=vizInkTwo, anchor=south] at (axis cs:3.21,51.0) {\method};
\addlegendimage{vizInk, densely dotted, line width=1.3pt}
\addlegendentry{followed by recovery}
\end{axis}
\end{tikzpicture}}
\caption{Average over the first nine benchmarks (logic, math, code) after stages 3, 4 and 5 and after the recovery stage; means over evaluation seeds, with the pooled standard error of each point as shaded bands (\cref{app:eval}). After stage 4 every method except Seq (two seeds) has one evaluation seed; all other points have three. Dotted segments join each method that has a recovery run to the same method followed by recovery; for Seq this is Seq + Recovery. \method, \method (shared budget) and Replay-KL (fixed share) have no points before stage 5.}
\label{fig:oldnine}
\end{figure}

\Cref{tab:matrix} lays out the design space after five stages. Under the shared budget the 7 replay steps of a stage are divided among the earlier domains, so the share of each shrinks as domains accumulate: over the five stages logic receives 16 replay steps, math 7, code 4 and safety 1. After three stages Replay-KL ends 1.17 points above Seq + Recovery (\cref{tab:three}). After five stages, every shared-budget replay method ends below Seq + Recovery: by 1.97 with the token-level loss alone, 1.83 with SATS, 0.93 with OPRD and 1.01 with all three terms. \Cref{fig:oldnine} follows the average over the first nine benchmarks. From stage 3 to stage 5, the shared-budget replay methods lose 3.13 (Replay-KL), 3.50 (Replay-KL + SATS) and 2.89 (Replay-KL + OPRD) points. Seq + Recovery ends at nearly the same level after three and after five stages (50.28 in \cref{tab:three} and 49.94).

A fixed share of $b = 7$ replay steps per earlier domain and stage closes this gap. Token-level replay with this share ends 2.29 points above Replay-KL and level with Seq + Recovery (+0.31, within the evaluation noise), with \Norm 0.561 against 0.547, and it stores no snapshot. With the three alignment terms, \method ends 1.28 points above \method (shared budget) and level with Seq + Recovery (+0.27). Its \Norm is 0.559, and its \NormD is 0.842 against 0.820 for Seq + Recovery. Against Seq + Recovery it is higher on logic (+1.59) and lower on safety ($-1.31$). Its differences on math (+0.87), code (0.00) and instruction following (+0.15) are within the evaluation noise. On the first nine benchmarks \method ends at 50.76, 1.63 above \method (shared budget) and 0.82 above Seq + Recovery, and Replay-KL (fixed share) at 50.57, 2.26 above Replay-KL.

The extra steps sit in different places: against Seq, the fixed share adds 70 replay steps inside the stages, 42 more than the shared budget, and recovery adds 35 steps at the end (\cref{tab:cost}), so the fixed share trains 174 steps and Seq + Recovery 139. Giving recovery as many training steps as the fixed share does not raise its score in these runs. Recovery extended to 70 steps, 14 blocks per domain chosen by the same rule, brings Seq and recovery to 174 steps and scores 48.35, 0.29 below the 35-step recovery, within the evaluation noise; over the two training runs of \cref{tab:trainseeds} the fixed share is level with both recovery stages ($-0.13$ and $-0.12$). Replay-KL (fixed share) and Seq + Recovery-70 then also match in work per step: every step samples 128 responses and runs one teacher forward pass and one student backward pass.

At 138 steps, token-level replay with $b = 3$ ends 0.52 below Seq + Recovery in Avg, beyond the noise. It is level on the 11 benchmarks outside safety ($-0.17$) and 0.75 above with balanced safety.

Against the shared budget, $b = 7$ changes both the distribution of replay steps over the earlier domains and their total number. The fixed share with $b = 3$ keeps the total close to that of the shared budget, 34 replay steps against 28. It spreads them differently: 12 replay steps in the fifth stage against 7, and 31 distinct prompt blocks replayed against 13. Token-level replay scores 46.66 with the shared budget, 48.12 with $b = 3$ and 48.95 with $b = 7$. With nearly the same number of replay steps, a fixed share of three per earlier domain and stage thus adds 1.45 points to the shared budget. The share $b = 7$ adds another 0.84, with 63 instead of 27 replay steps in stages 3 to 5. Once every earlier domain has a fixed share, of three replay steps or seven, the alignment terms leave the average within the evaluation noise. \method with $b = 3$ scores 48.36, 0.25 points above token-level replay with the same share, against $-0.05$ at $b = 7$ and +0.96 under the shared budget. Their gain therefore appears only under the shared budget. At $b = 3$ the only domain difference beyond the noise is code (+1.69). \method with $b = 3$ stays 0.59 points below token-level replay with $b = 7$, is level with Seq + Recovery ($-0.28$), and is 0.74 above \method (shared budget).

\subsection{Q3: Replay and recovery}
\label{sec:q-recovery}

When the stage snapshots are stored, replay followed by recovery gives the highest averages of the sequential configurations, level with fixed-share replay at $b = 7$ without recovery. The five replay methods followed by recovery end 0.43 to 0.60 points above Seq + Recovery, 0.52 on average (95\% interval 0.16 to 0.88; \cref{tab:matrix}): 0.50 for Replay-KL + Recovery, 0.51 for Replay-KL + SATS + Recovery, 0.60 for Replay-KL + OPRD + Recovery, 0.43 for \method (shared budget) + Recovery and 0.55 for \method + Recovery. All but the 0.43 are beyond the evaluation noise. Against token-level fixed-share replay with $b = 7$ and no recovery they end 0.11 to 0.29 points higher, within the noise.

The gain of the recovery stage itself shrinks as the average before it rises: 5.77 points after Seq, 2.47 after Replay-KL, 2.35 after Replay-KL + SATS, 1.53 after Replay-KL + OPRD, 1.44 after \method (shared budget) and 0.28 after \method, the last within the evaluation noise; after \method it moves code by +1.95, safety by +1.25 and instruction following by $-1.20$. These six configurations start 6.04 points apart (42.87 to 48.91) and end within 0.60 points of each other after recovery (48.64 to 49.24), so the gain of recovery is mostly the distance from its starting point to a common level: when replay is scarce, recovery still adds much; once the share is sufficient, little is left for it to add. Against \method (shared budget) + Recovery, \method is level in Avg ($-0.16$), level on the first nine benchmarks (+0.54) and 2.71 lower on safety; its code difference ($-1.46$) is within the noise.

\paragraph{Recovery from the original experts.}
The recovery stage above is taught by the stage snapshots. With the five original experts as teachers instead, on the same blocks, Replay-KL + expert recovery scores 49.18 and Seq + expert recovery 48.41 on evaluation seed 0. These are 0.19 and 0.13 points below the snapshot versions on the same seed, both within the noise of a comparison between single seeds (half-width 0.80, \cref{app:eval}). With expert recovery, replay before it adds 0.78 points on seed 0, also within that noise, against 0.83 with snapshot recovery on the same seed and 0.50 over three seeds, the last two beyond the evaluation noise.

\paragraph{Joint MOPD.}
Joint MOPD over the five domains trains from the student with all five experts at once and scores 48.68 on evaluation seed 0. Against the three-seed means of the other configurations, it is 0.04 points above Seq + Recovery and 0.28 and 0.23 below the fixed-share replay methods with $b = 7$. It is 0.39 to 0.57 below the replay methods followed by recovery. All these differences are within the noise of a single seed against three (half-width 0.66). It trains 104 steps, one pass over the new-domain blocks that Seq also sees. The sequential configurations with replay or recovery train 132 to 209 steps, because their replay and recovery blocks revisit prompts that were already trained (\cref{tab:cost}). Joint MOPD is therefore the joint reference with as many training steps as Seq (\cref{sec:byscenario}). The joint reference for the sequential configurations is the run over the same blocks twice, joint2 (208 steps, one fewer than the longest sequential configuration), which scores 49.02 over three evaluation seeds, 0.34 above the 104-step run, within the noise of a single seed against three. It is level with the fixed-share replay methods with $b = 7$, with the replay methods followed by recovery and with the 35-step recovery: $-0.22$ to +0.12 points against the first two and +0.38 against Seq + Recovery (95\% interval $-0.08$ to +0.85), all within the evaluation noise; against the mean of the 70-step recovery over two training runs it is +0.30, also within the noise. It is higher than the shared-budget replay methods without recovery, by 1.31 to 2.36 points, and than Replay-KL and \method with $b = 3$, by 0.91 and 0.66. With nearly as many steps as the longest sequential configurations, joint2 thus ends level with them in Avg.

Of the four merges without training (\cref{tab:five}), the two that need no snapshot score 45.07 for the uniform running average of the experts, below every replay configuration, and 50.42 for TIES applied stage by stage, which is judged safe on 19.6\% of the StrongREJECT prompts. TIES of the Seq snapshots scores 47.74, and TIES of the five experts, which needs every expert at the end, scores 52.29.

\subsection{Q4: What to distill on replay steps}
\label{sec:q-signal}

\begin{table*}[t]
\centering
\small
\setlength{\tabcolsep}{4pt}
\begin{tabular}{l c c c c c c c}
\toprule
& & & & & \multicolumn{2}{c}{MuSR} & \\
\cmidrule(lr){6-7}
Method & Avg & \Norm & $\Delta$Avg & Logic & score & $\BWT$ & cos \\
\midrule
Seq & 49.10 & 0.720 & $-$2.34 & 43.40 & 39.33 & $-$7.41 & n/a \\
Seq + Recovery & 50.28 & 0.793 & $-$1.17 & 47.74 & 39.86 & $-$6.88 & n/a \\
Replay-KL & 51.44 & 0.866 & & 48.52 & 42.64 & $-$4.10 & n/a \\
\quad + SATS & 51.84 & 0.890 & +0.40 & 50.19 & 45.94 & $-$0.79 & 0.04 / 0.04 \\
\quad + OPRD & \textbf{52.07} & \textbf{0.905} & \textbf{+0.63} & 49.88 & 44.14 & $-$2.60 & 0.23 / 0.28 \\
\quad + SEEKR & 51.38 & 0.862 & $-$0.06 & 50.20 & 45.55 & $-$1.19 & 0.13 / 0.13 \\
\quad + SATS + OPRD & 51.96 & 0.898 & +0.52 & 50.31 & 44.89 & $-$1.85 & 0.33 / 0.44 \\
\quad + SATS + value rel. & 51.89 & 0.894 & +0.45 & \textbf{50.83} & \textbf{46.16} & \textbf{$-$0.57} & 0.10 / 0.19 \\
\quad + SATS, WAAD regions & 51.74 & 0.884 & +0.30 & 49.92 & 44.66 & $-$2.07 & 0.06 / 0.05 \\
\midrule
Joint MOPD & 52.01 & 0.901 & & 49.73 & 44.20 & n/a & n/a \\
Joint Latent-MOPD & 52.31 & 0.919 & & 49.53 & 44.15 & n/a & n/a \\
\bottomrule
\end{tabular}
\caption{Three-stage prefix (logic, math, code) after the code stage. Avg is the nine-benchmark average (\cref{eq:avg}) and \Norm the normalized score (\cref{eq:norm}), anchored by the student and the best of the three experts on each benchmark; $\Delta$Avg is the difference to Replay-KL. MuSR, scored with the benchmark's parser (\cref{app:eval}), is given with its backward transfer from the shared stage-1 model (\cref{eq:bwt}). cos is the mean cosine between the auxiliary gradient and the token-level gradient over the replay steps of stage 2 / stage 3. The + SATS + OPRD and + SATS + value-relation rows are balanced 1:1:1. In these two rows the cosine of the added term is taken against the sum of the token gradient and the balanced SATS gradient. Their SATS gradient has cosine 0.04 / 0.08 and 0.03 / 0.04 with the token gradient. Continual rows are means over three evaluation seeds; joint rows are means of four evaluations of the same checkpoint (eight on MuSR), with Avg and \Norm computed from the printed per-benchmark scores (\cref{app:full}). Bold marks the best continual configuration in each column except cos.}
\label{tab:three}
\end{table*}

\Cref{tab:matrix} compares the alignment terms across replay budgets and with recovery. Under the shared budget, where replay is scarce, the hidden-state term of OPRD adds 1.04 points over Replay-KL (\Norm +0.044) and the three terms of \method add 0.96 (\Norm +0.041), while SATS alone adds 0.14, within the evaluation noise; adding SATS and value relations on top of the hidden-state term changes Avg by $-0.08$. With the fixed share $b = 7$ the three terms change Avg by $-0.05$ against Replay-KL (fixed share), with logic higher (+1.73) and safety lower ($-1.08$) beyond the noise and instruction following (+0.45) and code ($-0.98$) within it. After recovery the variants are within the evaluation noise of each other in Avg: against Replay-KL + Recovery, Replay-KL + SATS + Recovery changes Avg by +0.02, Replay-KL + OPRD + Recovery by +0.11, \method (shared budget) + Recovery by $-0.07$ and \method + Recovery by +0.05. The recovery stage itself uses the token-level loss only.

The gain of the three terms under the shared budget and its absence with a fixed share hold in all three averages (\cref{tab:five}): under the shared budget they add 0.67 points on the 11 benchmarks outside safety and 0.49 in Avg$_{\mathrm{bal}}$, both beyond the noise, while with $b = 3$ and $b = 7$ they stay within the noise in all three. Under the shared budget, the gain of the three terms exceeds their gain with $b = 7$ by 1.01 points in Avg (95\% interval 0.35 to 1.66). It exceeds their gain with $b = 3$ by 0.71 (95\% interval 0.06 to 1.37). On the 11 benchmarks outside safety these differences are 0.44 and 0.43 points, within the noise. In these runs alignment therefore adds to token-level replay where each earlier domain is replayed little.

On the three-stage prefix, \cref{tab:three} varies the alignment term under the shared budget, with the token-level channel and the balancing fixed. Replay-KL + OPRD replaces SATS with the hidden-state term on all 28 layers. Two variants add a term on top of Replay-KL + SATS: the hidden-state term on all 28 layers, or the value-relation term pooled over the same reasoning steps, layers and positions as SATS. Each added term is balanced to the token gradient separately (token : SATS : other $=$ 1:1:1). A third variant replaces the reasoning-step regions with WAAD chunks \citep{li2026attention}, and Replay-KL + SEEKR uses the per-head attention KL of SEEKR. Against Replay-KL, only + OPRD changes Avg beyond the evaluation noise (+0.63); + SATS adds 0.40, + SATS + value relation 0.45 and + SATS + OPRD 0.52, within it. With rollout seed 1, + SATS and + OPRD score 51.37 and 51.51 instead of 51.84 and 52.07; Replay-KL was trained once on the prefix. Under the benchmark's MuSR parser (\cref{app:eval}), every alignment term raises the logic average beyond the noise, from 1.36 (+ OPRD) to 2.32 (+ SATS + value relation), and + SATS raises MuSR by 3.31, beyond the noise; the 1.50 of + OPRD on MuSR and its 1.24 on the code average are within the noise. Stacked on Replay-KL + SATS, the hidden-state and value-relation terms give no gain on math ($-0.87$ and $-1.12$ against Replay-KL + SATS). With WAAD regions MuSR is 1.28 points lower than with reasoning steps, within the evaluation noise (\cref{app:pooling}).

After five stages, against Replay-KL, + SATS is lower on instruction following ($-1.15$) and higher on safety (+1.33), beyond the evaluation noise, and its logic difference ($-0.81$) is within it; + OPRD is higher on code (+1.61) and safety (+2.09), and its differences on math (+0.81), instruction following (+0.27) and MuSR (+0.97) are within the noise, as is the $-0.62$ of + SATS on MuSR. After recovery, Replay-KL + SATS + Recovery is higher than Replay-KL + Recovery on logic (+1.22) and MuSR (+3.75).

\paragraph{Gradient direction.}
\label{sec:direction}
On every replay step of stages 2 and 3, the cosine between the SATS gradient and the token-level gradient of Replay-KL + SATS lies between $-0.046$ and $0.079$: the two are nearly orthogonal. The same holds with WAAD regions ($-0.04$ to $0.10$). SEEKR's per-head attention gradient is weakly aligned with the token gradient (0.01 to 0.28, mean 0.13). The value-relation gradient is also weakly aligned with the token plus SATS gradient (0.04 to 0.34). The hidden-state gradient is more aligned: 0.09 to 0.45 against the token gradient in Replay-KL + OPRD, and 0.25 to 0.61 against the token plus SATS gradient when stacked. With a mean cosine of 0.23 to 0.28, the projection of the hidden-state gradient on the token-level gradient is about a quarter of its norm (5\% to 8\% of its squared norm), against about 4\% of the norm for the attention gradient. With $r = 1$, \cref{eq:balance} adds to the token gradient a second gradient of the same norm, so the gradient of a replay step has $\sqrt{2 + 2\cos\varphi}$ times the norm of the token gradient, where $\varphi$ is the angle between the two: 1.4 to 1.8 times over the replay steps of the five-stage runs with alignment terms. The balancing ratio and the pooling regions are examined in \cref{app:ratio,app:pooling}.

\subsection{Q5: Costs}
\label{sec:q-cost}

\paragraph{The newest domain.}
The fixed share of $b = 7$ lowers the score on the newest domain, instruction following, for token-level replay: by 1.19 points against the same signal under the shared budget, beyond the evaluation noise, with 28 replay steps against 21 new-domain steps in the fifth stage; for \method the decrease, 0.75, is within the noise. Against Seq + Recovery the differences are $-0.30$ and +0.15, both within the noise. With $b = 3$ the decreases against the shared budget are 0.52 for token-level replay, within the noise, and 0.93 for \method, beyond it. Averaged over these four fixed-share runs, the decrease is 0.85 points (95\% interval 0.36 to 1.34). The recovery stage also lowers the score on the last domain after five of the six configurations it follows. The decrease is 0.95 points after Seq, 1.00 after Replay-KL, 1.71 after Replay-KL + OPRD, 1.27 after \method (shared budget) and 1.20 after \method. After Replay-KL + SATS, recovery moves instruction following by +0.14, within the noise. In the recovery stage the prompts of the last domain are taught by $M_K$, the model the stage starts from. The steps on that domain therefore pull the domain back toward $M_K$ rather than above it, while the other four fifths of the steps follow the earlier teachers.

\paragraph{Over-refusal.}
Seq + Recovery, every replay method followed by recovery and both fixed-share methods with $b = 7$ raise StrongREJECT to 90 to 96 and the safe rate on the unsafe prompts of XSTest to 96.9 to 98.8. They also lower the compliance rate on the safe prompts of XSTest to 16.0 to 26.8, and joint2 lowers it to 28.8. The compliance rate is 89.4 for the student, 94.7 for Seq and 70.9 to 83.1 for the shared-budget replay methods, whose safe rates on the unsafe prompts are 71.5 to 84.9. All except the student's are means over three evaluation seeds. The safety expert, evaluated once, scores 25.2 on the safe prompts and 98.0 on the unsafe prompts. Both the recovery stage and the fixed share end closer to it than shared-budget replay. Safety accounts for 42\% to 65\% of the lead of Seq + Recovery over the shared-budget replay methods. Against Replay-KL + OPRD, for example, the safety average of Seq + Recovery is 2.83 points higher. Since safety holds 3 of the 14 benchmarks, this contributes 0.61 of the 0.93-point difference in Avg. The 70-step recovery lowers the compliance rate further, to 13.4.

The safety average counts refusal on two benchmarks and compliance on one. Balanced safety scores weight refusal of harmful prompts and compliance with safe ones equally (\cref{app:balanced}). Under them the shared-budget replay methods rank above Seq + Recovery, the fixed-share configurations and joint MOPD, the reverse of their order in Avg. Recovery after shared-budget replay lowers the balanced average by 1.3 to 2.0 points. In Avg$_{\mathrm{bal}}$ over two training runs, Replay-KL (fixed share) is level with Seq + Recovery (+0.24) and with Seq + Recovery-70 (+0.73, \cref{tab:trainseeds}).

\begin{table}[t]
\centering
\footnotesize
\setlength{\tabcolsep}{2.5pt}
\begin{tabular}{l c c l}
\toprule
& \multicolumn{2}{c}{Steps} & \\
\cmidrule(lr){2-3}
Method & $K = 3$ & $K = 5$ & Teachers kept \\
\midrule
Seq & 62 & 104 & $E_k$ \\
Seq + Recovery & 83 & 139 & $E_k$; then $M_1, \dots, M_K$ \\
Seq + Recovery-70 & n/a & 174 & $E_k$; then $M_1, \dots, M_K$ \\
Shared budget & 76 & 132 & $E_k$, $M_{k-1}$ \\
\quad + Recovery & n/a & 167 & $E_k$, $M_{k-1}$; then $M_1, \dots, M_K$ \\
Fixed share, $b = 3$ & n/a & 138 & $E_k$, $M_{k-1}$ \\
Fixed share, $b = 7$ & n/a & 174 & $E_k$, $M_{k-1}$ \\
\quad + Recovery & n/a & 209 & $E_k$, $M_{k-1}$; then $M_1, \dots, M_K$ \\
Joint MOPD & 62 & 104 & $E_1, \dots, E_K$ at once \\
Joint MOPD, two passes & n/a & 208 & $E_1, \dots, E_K$ at once \\
\bottomrule
\end{tabular}
\caption{Training steps and the teachers that must be kept besides the student after three and after five stages. The logic stage has 20 steps and every later stage 21 new-domain steps. The shared budget covers Replay-KL, its alignment variants and \method (shared budget); the fixed share with $b = 7$ covers Replay-KL (fixed share) and \method; the stages of $b = 3$ have 20, 28, 27, 30 and 33 steps; an indented + Recovery row adds the 35-step recovery stage to the replay configuration above it, and Seq + Recovery-70 has a 70-step recovery stage; the two-pass joint run is joint2. Recovery from the original experts has the steps of recovery from the snapshots and holds $E_1, \dots, E_K$ at the end instead of $M_1, \dots, M_K$. Replay steps add one forward and backward pass of the student for each auxiliary term. n/a: not evaluated after three stages.}
\label{tab:cost}
\end{table}

\paragraph{Steps and stored models.}
\Cref{tab:cost} summarizes what each choice needs. After five stages the fixed share with $b = 7$ adds 42 steps to shared-budget replay and 70 to sequential training, all of them replay steps inside the stages, and every replay configuration holds two teachers at a time. Seq + Recovery adds 35 steps and must keep every stage snapshot until the end, a requirement that grows with the number of stages. The replay steps of the fixed share grow quadratically with the number of stages (\cref{sec:cost,sec:discussion}). Measured as the sum of the logged step times on the four A100 GPUs, the five stages take 3.4 hours for Seq, 4.1 for Replay-KL, 5.5 for \method (shared budget), 5.8 for Replay-KL (fixed share) and 9.7 for \method; Seq + Recovery takes 4.5 hours, Seq + Recovery-70 5.6 and joint2 6.8. In stages 3 to 5 the median replay step takes 122 to 131 seconds with the token-level loss alone and 289 to 326 seconds with the three alignment terms. Their separate backward passes raise the median time of the student update on a replay step from between 17 and 20 seconds to between 137 and 157 seconds.

\subsection{By setting}
\label{sec:byscenario}

\begin{table}[t]
\centering
\footnotesize
\setlength{\tabcolsep}{3pt}
\begin{tabular}{l c c c c}
\toprule
& & \multicolumn{3}{c}{Gain over Seq} \\
\cmidrule(lr){3-5}
Configuration & Steps & Avg & Avg$_{11}$ & Avg$_{\mathrm{bal}}$ \\
\midrule
\multicolumn{5}{l}{\textit{S1: only the previous model is kept}} \\
Replay-KL, shared budget & 132 & +3.80 & +2.28 & +2.67 \\
Replay-KL + SATS & 132 & +3.94 & +2.10 & +2.63 \\
Replay-KL + OPRD & 132 & +4.84 & +3.04 & +2.95 \\
\method (shared budget) & 132 & +4.76 & +2.96 & \textbf{+3.16} \\
Replay-KL, $b = 3$ & 138 & +5.25 & +3.29 & +1.80 \\
\method, $b = 3$ & 138 & +5.50 & +3.53 & +1.43 \\
Replay-KL (fixed share) & 174 & \textbf{+6.09} & +3.92 & +1.70 \\
\method & 174 & +6.04 & \textbf{+4.15} & +1.68 \\
\midrule
\multicolumn{5}{l}{\textit{S2: every stage snapshot kept to the end}} \\
Seq + Recovery & 139 & +5.77 & +3.45 & +1.06 \\
Seq + Recovery-70 & 174 & +5.48 & +3.26 & +0.39 \\
Replay-KL + Recovery & 167 & +6.27 & +3.97 & +1.36 \\
Replay-KL + SATS + Recovery & 167 & +6.29 & +3.97 & +1.31 \\
Replay-KL + OPRD + Recovery & 167 & \textbf{+6.38} & +4.13 & +1.40 \\
\method (shared budget) + Recovery & 167 & +6.20 & +3.61 & +1.14 \\
\method + Recovery & 209 & +6.32 & \textbf{+4.17} & \textbf{+1.73} \\
\midrule
\multicolumn{5}{l}{\textit{Every expert at the same time}} \\
Joint MOPD (joint2) & 208 & +6.16 & +3.46 & +1.76 \\
\midrule
\multicolumn{5}{l}{\textit{No training}} \\
Uniform average (fits S1) & 0 & +2.21 & $-$0.16 & +1.36 \\
TIES stage by stage (fits S1) & 0 & +7.55 & +8.14 & +7.40 \\
TIES of the Seq snapshots (S2) & 0 & +4.87 & +3.67 & +4.15 \\
TIES (experts kept, S2) & 0 & +9.42 & +7.81 & +8.14 \\
\bottomrule
\end{tabular}
\caption{Gains over Seq after five stages by setting, in points of Avg, Avg$_{11}$ and Avg$_{\mathrm{bal}}$ (\cref{tab:five}), computed from the means over three evaluation seeds before rounding; Seq scores 42.87, 43.84 and 45.84. Steps are training steps (\cref{tab:cost}); with $b = 3$ the fixed share starts at stage 3, after the stage 2 of the shared budget. The uniform average of the experts and TIES applied stage by stage, which also keeps the base model, fit S1; TIES of the Seq snapshots needs the snapshots, and TIES of the experts needs the experts at the end, like recovery from them. Bold marks the largest gain in each of the S1 and S2 groups.}
\label{tab:byscenario}
\end{table}

After five stages, Seq ends 6.16 points below joint2, the joint MOPD with nearly as many steps as the longest sequential configuration. Joint MOPD with the 104 steps of Seq is 6.01 points above Seq on evaluation seed 0, so most of this gap comes from training the domains in sequence rather than from the extra steps of joint2. \Cref{tab:byscenario} gives the gains over Seq by setting in the three averages.

\paragraph{Only the previous model is kept (S1).}
Token-level replay closes 3.80 points with the shared budget, 5.25 with $b = 3$ and 6.09 with $b = 7$, which is level with recovery (+0.31) without storing any snapshot, and over two training runs it is level with both recovery stages ($-0.13$ and $-0.12$, \cref{tab:trainseeds}). The three alignment terms together add beyond the noise only under the shared budget, in all three averages; no single term does so in all three. With balanced safety the shared-budget configurations gain the most (2.63 to 3.16), because they keep more compliance on the safe prompts of XSTest, and over two training runs the fixed share of seven is level with both recovery stages (\cref{tab:trainseeds}). Of the merges that fit this setting, the uniform running average of the experts gains 2.21 in Avg and is level with Seq on the 11 benchmarks outside safety ($-0.16$), and TIES applied stage by stage, which keeps the base model besides the previous merge, gains 7.55 without training and is judged safe on 19.6\% of the StrongREJECT prompts against 91.1\% for the fixed share with $b = 7$.

\paragraph{Snapshots or experts kept to the end (S2).}
Recovery after Seq closes 5.77 points, and with the original experts as its teachers it is level with the snapshot version on evaluation seed 0. Replay before it adds 0.52 points in Avg on average (95\% interval 0.16 to 0.88). The smaller the gap a replay method leaves, the less recovery adds. After recovery, the replay signals are within the noise of each other in Avg. With balanced safety, recovery after shared-budget replay lowers the average by 1.3 to 2.0 points. TIES of the Seq snapshots needs the snapshots that recovery uses. It ends 0.90 below Seq + Recovery in Avg (95\% interval 0.44 to 1.37) and level with it on the 11 benchmarks outside safety (+0.21). With balanced safety it is 3.10 above, because it refuses less: it is judged safe on 39.4\% of the StrongREJECT prompts, against 95.2\% for Seq + Recovery.

\paragraph{Every expert at once.}
In Avg, joint2 ends level with token-level replay at $b = 7$, with replay followed by recovery ($-0.22$ to +0.12) and with the 35-step recovery (+0.38, 95\% interval $-0.08$ to +0.85), and with the 70-step recovery over two training runs (+0.30). On the 11 benchmarks outside safety, joint2 is level with Seq + Recovery (+0.01) and 0.51 below the mean of the five replay methods followed by recovery (95\% interval 0.07 to 0.94). With MBPP and MBPP+ counted once, it is 0.48 above Seq + Recovery. With balanced safety, it is 0.71 above Seq + Recovery and 0.38 above the mean of the five (95\% interval 0.02 to 0.73). These three differences are beyond the noise.

\section{Discussion}
\label{sec:discussion}

\paragraph{Replay and recovery.}
In these runs in-stage replay and a final recovery stage play different roles. Replay acts inside every stage and needs only the previous model and the prompts of earlier domains; recovery acts once at the end and needs every stage snapshot, or every expert. After scarce replay, recovery still adds much; after a sufficient fixed share, replay alone ends level with recovery alone, and recovery after it adds little (\cref{sec:q-recovery}). A pipeline that cannot store its snapshots thus reaches the level of recovery through the replay share, and one that stores them gains most from recovery when its replay was scarce.

\paragraph{Settings.}
When only the previous model is kept (S1), recovery is not available, and among the trained configurations compared here in-stage replay toward the previous student is the only one that keeps earlier skills at the level of recovery. Its main choice is the replay share of each earlier domain, and the replay signal matters where that share is small: in these runs a sufficient replay share closed more of the gap than any signal under the shared budget, and alignment added to token-level replay where replay was scarce (\cref{sec:q-budget,sec:q-signal}).

When the snapshots or the experts are kept to the end (S2), replay followed by recovery gives the highest averages of the trained sequential configurations; when all experts can be held at once, the 208-step joint MOPD ends level with the best sequential configurations in Avg (\cref{tab:byscenario}). S3, in which a long sequence is split into segments, is discussed under segment-wise consolidation below.

\paragraph{Merging and distillation.}
Distillation needs only the teachers' token distributions on the student's rollouts, not their weights, so it also applies to experts of another architecture over the same vocabulary, which cannot be merged in weight space. For experts trained from a shared base, supervised fine-tuning, soft-label distillation and MOPD have been found to reach nearly identical accuracy, with weight merging recovering much of the experts' capabilities \citep{xie2026rethinking}, and for experts in a shared basin the MOPD objective matches Fisher merging to second order \citep{liu2026fishermopd}. The first of these studies also finds that merging takes minutes of CPU time but degrades as model size decreases and task interference increases \citep{xie2026rethinking}. Other studies compare the two. MOPD scores above parameter merging in a study whose main results use Qwen3-30B-A3B \citep{ma2026mopd}. Merging has been reported to beat on-policy self-distillation for continual learning \citep{wu2026merging}. ACLArena compares merging with multi-teacher on-policy distillation for recovery \citep{wang2026aclarena}.

\paragraph{Frontier practice.}
DeepSeek-V4 replaces mixed reinforcement learning with one joint on-policy distillation stage \citep{deepseek2026v4}, and GLM-5 distills across stages after a sequence of stages \citep[\S3.5]{glm2026glm5}; neither report compares the two pipelines. Joint MOPD and Seq + Recovery here make that comparison with a 1.5B student. With 208 steps against 139, joint MOPD ends level with Seq + Recovery in Avg and on the 11 benchmarks outside safety, and above it with MBPP and MBPP+ counted once and with balanced safety (\cref{sec:byscenario}). Against replay followed by recovery it is level in Avg. Against the mean of the five replay methods followed by recovery, it is below on the 11 benchmarks outside safety and above with balanced safety.

\paragraph{Scaling with the number of stages.}
The fixed share grows the replay budget of a stage linearly with its index and the total quadratically (\cref{sec:cost}). With $b = 7$ and 21 new-domain steps per stage, stage 15 would have 98 replay steps against 21 new-domain steps, and 15 stages would have 735 replay steps against 315. A recovery stage grows linearly with the number of domains, to 105 steps for 15 domains at 7 blocks each, and it needs all 15 snapshots at the end. Each prompt of a recovery stage is routed to one teacher, so the target of each token does not become harder with more teachers, but the stage as a whole may: joint MOPD stages with more than ten teachers have been reported \citep{deepseek2026v4,nvidia2026nemotron3ultra,park2026solaropen2,kwai2026keyevl2} without experiments that vary the number of teachers, and IM-MOPD finds that joint distillation from five teachers under-recovers some of them \citep{kim2026immopd}.

These runs bracket how many replay steps per domain and stage suffice for token-level replay. In Avg the shared budget, which gives each earlier domain one or two replay steps in the fifth stage, ends below recovery, three steps per domain and stage end closer, and seven close the gap; on the 11 benchmarks outside safety three are already level with recovery, and with balanced safety the shared budget ends above it (\cref{sec:q-budget,sec:q-cost}).

\paragraph{Segment-wise consolidation.}
A middle ground would split a long sequence into segments, for example 15 stages into three segments of five. It would use fixed-share replay inside each segment and end each segment with a recovery stage. In that stage the model produced by the previous recovery teaches the domains of earlier segments. The replay budget and the number of teachers of a recovery stage are then both bounded by the segment size. At any time only the last consolidated model and the snapshots of the current segment are kept. The idea is close to Progress \& Compress, which alternates learning with consolidation into a knowledge base \citep{schwarz2018progress}, and to CPC, in which one stage can bring a group of teachers \citep{he2026cpc}. In LLMs, Nemotron-Cascade 2 places one multi-teacher stage inside its cascade of reinforcement learning stages \citep{yang2026nemotroncascade2} and Nemotron 3 Ultra runs MOPD twice \citep{nvidia2026nemotron3ultra}, but no LLM work that could be found ends each segment with a recovery stage in which the previous consolidated model teaches the earlier segments.

A release cycle is a case of this design (\cref{sec:scenarios}). A released model learns a new segment of domains while the experts of the earlier segment are gone. Only the released model can then teach the earlier domains, by replay inside the stages or in a recovery stage at the end of the segment. As releases accumulate, each earlier domain receives a smaller share of a shared replay budget, the condition under which the three alignment terms together and the hidden-state term alone added about one point here (\cref{sec:q-signal}), so the release cycle is where they would be tested next.

\paragraph{Limitations.}
The configurations are trained at one scale (a 1.5B student with experts trained from it) and in one domain order, and differences between configurations are observations under these conditions.

The rankings that involve shared-budget replay or recovery depend on how refusal and compliance are weighted (\cref{sec:q-cost,app:fivefull,app:balanced}). On the 11 benchmarks outside safety, Seq stays the lowest trained configuration. On these benchmarks, Replay-KL + OPRD and \method (shared budget) end level with Seq + Recovery ($-0.41$ and $-0.50$), and Replay-KL and Replay-KL + SATS end 1.17 and 1.35 points below it. On the same benchmarks, fixed-share replay and replay followed by recovery end $-0.17$ to $+0.72$ points against Seq + Recovery.

On BBH and GYM only the boxed answer is scored. At the final checkpoints, 13\% to 23\% of the BBH responses and 18\% to 43\% of the GYM responses give no boxed answer, \texttt{<answer>} tag or ``Answer:'' line (\cref{app:eval}); most of these responses state an answer in prose, mostly a wrong one. Counting the prose answers would raise every trained five-stage configuration by an estimated 0.523 to 0.825 points in Avg. It would change the difference between any two of them by at most 0.302 and the gain, loss or level label of 4 of the 136 pairs. Among them, Replay-KL + Recovery and Replay-KL + SATS + Recovery would become level with Seq + Recovery (+0.37 and +0.41). On MuSR the losses of the three-stage prefix measure the answer format. After the math stage the student ends most MuSR responses with a boxed answer instead of the ``ANSWER:'' line that the prompt asks for and the parser reads, so these responses are scored as random options. Read from the box, Seq gains 1.95 MuSR points over the three stages instead of losing 7.41, and + SATS leads Replay-KL on MuSR by 0.99 instead of 3.31; in the three-stage logic average, Replay-KL + SATS, Replay-KL + SEEKR and SATS with WAAD regions then lead Replay-KL by 0.87 to 0.90, against a half-width of 0.90 (\cref{app:eval}).

\paragraph{Future work.}
The next comparisons are rehearsal on stored responses of the experts, and joint MOPD and a recovery stage with the domains mixed within a batch. Longer stages and longer sequences would test the segment-wise consolidation above (S3), including the release cycle with and without alignment terms. Further directions are alignment toward the previous model on the prompts of the new domain when no earlier prompt can be kept, experts of other architectures, which cannot be merged in weight space and leave distillation as the way to combine them, larger students, other domain orders and alignment terms inside the recovery stage.

\section{Conclusion}
\label{sec:conclusion}

On a five-domain sequence with a 1.5B student and five experts trained from it, sequential on-policy distillation without replay or recovery ends 6.2 points below joint MOPD trained nearly as long as the longest sequential pipeline, and what the learner keeps decides which methods are available and how much of this gap they close. With only the previous model kept, in-stage replay toward the previous student closes 3.8 points under a shared replay budget and, with a fixed share of seven replay steps per earlier domain and stage, ends level with a final recovery stage without storing any snapshot; the hidden-state term on replay steps, alone or with the two other terms, adds about one point where replay is scarce. When every snapshot is stored, recovery closes 5.9 points over two training runs, its gain shrinks as the replay before it gets stronger, and replay followed by recovery ends level with the 208-step joint MOPD. Against the shared budget, a fixed share lowers the score on the newest domain. Its replay steps grow quadratically with the number of stages.

A release cycle combines these settings (\cref{sec:discussion}). There a released model is the only teacher of its earlier domains while it learns a new segment. It is where the alignment terms would be tested next.

\bibliographystyle{unsrtnat}
\bibliography{references}

\clearpage
\appendix
\onecolumn
\section{Notation}
\label{app:notation}

\begin{center}
\small
\begin{tabular}{l p{0.74\textwidth}}
\toprule
Symbol & Meaning \\
\midrule
$d_k$, $\mathcal{D}_k$ & domain of stage $k$ and its prompt set \\
$E_k$ & expert of domain $d_k$ \\
$\pi_0 = M_0$ & base student \\
$M_k$ & student at the end of stage $k$ \\
$\pi_\theta$ & student being trained \\
$\mathcal{B}_{<k}$ & replay buffer: prompts of $d_1, \dots, d_{k-1}$, no responses or labels \\
$T$ & teacher of a step: $E_k$ on new-domain steps, $M_{k-1}$ on replay steps \\
$m$ & new-domain steps between two replay steps under the shared budget \\
$b$ & replay steps that each earlier domain receives in each stage under the fixed share ($b = 3$ or $7$) \\
$V_t$, $\kappa$ & student's $\kappa$ most probable tokens at response position $t$ \\
$r_{t,v}$, $A_{t,v}$ & reward and advantage of candidate $v \in V_t$ (\cref{eq:tokreward}) \\
$\ltok$ & token-level loss (\cref{eq:tokloss}) \\
$\ell_{\mathrm{hid}}$, $\mathcal{L}_h$ & hidden-state loss (\cref{eq:hid}) and its layers \\
$\lsats$, $\ell_{\mathrm{VR}}$ & reasoning-step pooled attention loss (\cref{eq:sats}) and value-relation loss \\
$\mathcal{L}$, $H$, $\mathcal{R}_y$, $\mathcal{Y}$ & layers of the pooled terms, query heads per layer, reasoning steps of response $y$ in the window, responses of a micro-batch with at least one step \\
$N$, $N_W$ & number of response positions of a micro-batch, and of those in the window (\cref{eq:tokloss,eq:hid}) \\
$W$ & window length: the alignment terms use the last $W$ response tokens \\
$g_{\mathrm{tok}}$, $g_{\mathrm{aux}}$, $\mathbf{u}$ & token-level gradient, auxiliary gradient, unit direction of the two pooled terms (\cref{eq:joint}) \\
$r$, $\rho$ & target ratio of the token part to the auxiliary part of the update, in norm (\cref{eq:balance}); weight of the hidden-state direction (\cref{eq:joint}); $r_{t,v}$ with subscripts is a reward \\
$D$, $d_{\mathrm{head}}$ & hidden size, head dimension \\
$\tau$, $s_\tau$, $\mathcal{T}$ & benchmark, its score, benchmarks of the domains learned up to stage $K$ \\
$\mathrm{Avg}$, $\Norm$, $\NormD$ & average over the benchmarks of the learned domains (\cref{eq:avg}), normalized score (\cref{eq:norm}), per-domain normalized score (\cref{sec:metrics}) \\
Avg$_{11}$, Avg$_{\mathrm{bal}}$ & Avg over the 11 benchmarks outside safety; Avg with the three safety scores replaced by three copies of the balanced score $S_{\mathrm{bal1}}$ (\cref{app:balanced}) \\
$\BWT_\tau$ & backward transfer (\cref{eq:bwt}) \\
\bottomrule
\end{tabular}
\end{center}

\section{Training details}
\label{app:training}

\begin{center}
\small
\begin{tabular}{l p{0.67\textwidth}}
\toprule
Setting & Value \\
\midrule
Prompts per step & 32 (one block, one domain) \\
Responses per prompt & 4, temperature 1.0, top-$p$ 1.0 \\
Max prompt / response length & 2048 / 16384 tokens \\
Optimizer & AdamW, $\beta = (0.9, 0.999)$, weight decay 0.01 \\
Learning rate & $10^{-5}$, constant, no warmup \\
Policy update & one epoch, one mini-batch per step; the importance ratio is one, so the clip ratio 0.2 (dual clip 3.0) never acts \\
Token channel & one advantage per candidate of the student's top 16 (\cref{eq:tokreward,eq:tokloss}); the weights $w_{t,v}$ are the student's probabilities renormalized over the 16; teacher log-probabilities over the full vocabulary; the top 16, the weights and the student's log-probabilities of \cref{eq:tokreward} come from a forward pass of the training model before the update, not from the inference engine; rewards divided by their sample standard deviation over the candidates of all response positions of each domain in the batch, padding excluded, no mean subtracted \\
Loss aggregation & per micro-batch; each GPU packs its 32 responses of a step into micro-batches by token count: the token-level and hidden-state terms average over response positions, SATS and VR over responses; each micro-batch loss is weighted by its share of the GPU's 32 responses, and gradients are averaged over the four GPUs \\
Optimizer state & a new AdamW state in every stage, including recovery; a stage starts from the weights of the previous one only \\
KL regularizer, entropy bonus & none \\
Gradient clipping & norm 1000 (LastOPD default; never reached, the largest gradient norm over the continual runs reported here is 140) \\
Hardware & 4 $\times$ A100 80GB, FSDP, vLLM rollouts (tensor parallel 1) \\
\bottomrule
\end{tabular}
\end{center}

\paragraph{Precision.} As in the LastOPD launcher, teachers are loaded with fp32 weights and run their forward pass under bf16 autocast; the student keeps fp32 master weights and uses FSDP mixed precision with bf16 parameters and fp32 gradient reduction.

\paragraph{Prompt formats.} Math prompts keep the DAPO-Math-17k instruction to write the answer on its own line after ``Answer:''. Logic prompts place the Reasoning Gym default instruction (answer inside \texttt{<answer>} tags) in the user turn before the question. Code prompts are the problem statements. Prompts longer than 2048 tokens under the student's chat template were removed beforehand so that no block loses prompts during training. Stage-4 prompts come from STAR-1 and STAR-benign-915, and stage-5 prompts from the instruction-following subset of UltraData-SFT-2605 (\cref{sec:setup}).

\paragraph{Replay and recovery positions.} Blocks are numbered within each domain from 0, and \cref{tab:positions} lists the blocks of each domain that are replayed in each stage and that enter the recovery stage. Under the shared budget a replay block follows every three new-domain blocks. Under the fixed shares the replay blocks are spread evenly among the 21 new-domain blocks, and stage 2 is the same as under the shared budget. Within a stage the replay blocks cycle over the earlier domains in sequence order, each domain taking its blocks in the listed order, and the recovery steps cycle over all domains of the sequence in the same way.

\begin{center}\begin{minipage}{\linewidth}\centering\captionsetup{hypcap=false}
\small
\setlength{\tabcolsep}{4pt}
\begin{tabular}{l l l l l}
\toprule
Replay & Logic & Math & Code & Safety \\
\midrule
\multicolumn{5}{l}{\textit{Shared budget}} \\
Stage 2 & 0, 3, 6, 10, 13, 16, 19 & & & \\
Stage 3 & 0, 6, 13, 19 & 0, 10, 20 & & \\
Stage 4 & 0, 10, 19 & 0, 20 & 0, 20 & \\
Stage 5 & 0, 19 & 0, 20 & 0, 20 & 0 \\
\midrule
\multicolumn{5}{l}{\textit{Fixed share, $b = 7$}} \\
Stage 3 & 0, 3, 6, 9, 11, 14, 17 & 0, 3, 6, 9, 12, 15, 18 & & \\
Stage 4 & 1, 4, 7, 10, 12, 15, 18 & 1, 4, 7, 10, 13, 16, 19 & 0, 3, 6, 9, 12, 15, 18 & \\
Stage 5 & 2, 5, 8, 10, 13, 16, 19 & 2, 5, 8, 11, 14, 17, 20 & 2, 5, 8, 11, 14, 17, 20 & 0, 3, 7, 10, 13, 17, 20 \\
\midrule
\multicolumn{5}{l}{\textit{Fixed share, $b = 3$}} \\
Stage 3 & 0, 7, 14 & 0, 8, 15 & & \\
Stage 4 & 2, 10, 17 & 2, 10, 18 & 0, 8, 16 & \\
Stage 5 & 5, 12, 19 & 5, 12, 20 & 4, 12, 20 & 0, 10, 20 \\
\bottomrule
\end{tabular}

\vspace{8pt}
\begin{tabular}{l l l}
\toprule
Recovery & Logic & Each later domain \\
\midrule
After three stages, 21 steps & 0, 3, 6, 10, 13, 16, 19 & 0, 3, 7, 10, 13, 17, 20 \\
After five stages, 35 steps & 0, 3, 6, 10, 13, 16, 19 & 0, 3, 7, 10, 13, 17, 20 \\
Seq + Recovery-70, 70 steps & 0, 1, 3, 4, 6, 7, 9, 10, 12, 13, 15, 16, 18, 19 & 0, 2, 3, 5, 6, 8, 9, 11, 12, 14, 15, 17, 18, 20 \\
\bottomrule
\end{tabular}
\captionof{table}{Replay and recovery blocks of each domain, numbered within the domain from 0. Top: the blocks of each earlier domain that are replayed in a stage; stage 2 replays the logic blocks of the shared budget under all three budgets. Bottom: the blocks that each domain contributes to the recovery stage; every domain after logic (math and code after three stages, math, code, safety and instruction following after five) contributes the same blocks.}
\label{tab:positions}
\end{minipage}\end{center}

\section{Evaluation details}
\label{app:eval}

\begin{center}
\small
\begin{tabular}{l l c c c c}
\toprule
Benchmark & Domain & Samples per item & Temperature & Max new tokens & Scored samples \\
\midrule
BBH & logic & 1 & 0.7 & 8192 & 1080 \\
Reasoning Gym & logic & 1 & 0.7 & 16384 & 950 \\
MuSR & logic & 1 & 0.7 & 8192 & 756 \\
AIME24 & math & 16 & 0.7 & 31744 & 480 \\
AIME25 & math & 16 & 0.6 & 31744 & 480 \\
Minerva & math & 4 & 0.7 & 31744 & 1088 \\
LCB-e & code & 4 & 0.6 & 16384 & 268 \\
MBPP & code & 2 & 0.7 & 16384 & 756 \\
MBPP+ & code & 2 & 0.7 & 16384 & 756 \\
StrongREJECT & safety & 4 & 0.6 & 8192 & 1252 \\
WildJailbreak & safety & 4 & 0.6 & 8192 & 1000 \\
XSTest & safety & 4 & 0.6 & 8192 & 1800 \\
IFEval & instruction following & 4 & 0.6 & 16384 & 2164 \\
IFBench & instruction following & 4 & 0.6 & 16384 & 1200 \\
\bottomrule
\end{tabular}
\end{center}

All benchmarks use top-$p$ 0.95 and the model's chat template. Each checkpoint other than the student and the experts is evaluated with the vLLM seeds 0, 1 and 2 unless a table or its caption states otherwise. The nine benchmarks of the three-stage prefix follow the sample counts and temperatures stated by Latent-MOPD \citep{fang2026latentmopd}, including temperature 0.6 for AIME25 and LCB-e.
\begin{itemize}
\item \textbf{Math.} The problem is followed by ``Please reason step by step, and put your final answer within \textbackslash boxed\{\}.'' and graded with the OPRD answer checker.
\item \textbf{BBH.} 40 items from each of the 27 subtasks, the same instruction; the last boxed answer is compared after normalization.
\item \textbf{Reasoning Gym.} 25 items from each of 38 tasks, generated with seed 4242, the same instruction; the last boxed answer is passed to the task's own scorer.
\item \textbf{MuSR.} The official chain-of-thought prompt of the benchmark, which asks for a last line ``ANSWER:'' with the choice number. As in the benchmark's parser, the scorer reads the last line that contains ``answer:'', regardless of case. It searches the whole response, thinking included. A random option, drawn with a fixed seed, is scored when the response has no such line or that line names no option.
\item \textbf{MBPP and MBPP+.} The user turn of the lm-evaluation-harness \texttt{mbpp\_plus\_instruct} task, zero-shot; the final answer after the thinking block is sanitized and run with EvalPlus.
\item \textbf{LCB-e.} The generic LiveCodeBench prompt; the last code block is run on the official tests. Responses are generated for the 165 easy problems dated from 2024-01-06 to 2025-01-04 and scored on the 67 dated from 2024-08-01, as in Latent-MOPD.
\item \textbf{Safety.} Responses are judged by Qwen3Guard-Gen-8B in response moderation mode \citep{zhao2025qwen3guard} on the text after the thinking block, or on the whole response when the reasoning never closes. StrongREJECT and WildJailbreak score the safe rate, with the judge's Controversial label counted as unsafe; WildJailbreak uses 250 adversarial harmful prompts sampled with seed 42 from the walledai mirror. On XSTest the safe prompts score the compliance rate, the share of responses that close their reasoning and are judged not to refuse, and the unsafe prompts the safe rate; the safety average uses the compliance rate. The judge's refusal label is binary, so a partial refusal counts as a refusal or as compliance as the judge labels it.
\item \textbf{Instruction following.} IFEval and IFBench are scored with the verifiers of the IFBench repository (the 25 instruction types of IFEval and the 58 of IFBench) on the answer after the thinking block; the score is prompt-level strict accuracy.
\end{itemize}

\paragraph{Truncation and answer extraction.} For the 20 five-stage configurations that are reported, the 17 trained rows of \cref{tab:five} and the three of \cref{tab:fiveone}, with all their evaluation seeds, at most 3.5\% of the responses of each logic benchmark reach the max new tokens. BBH and GYM are scored on the last boxed answer, while the logic training prompts ask for \texttt{<answer>} tags. Across these configurations, 14\% to 23\% of the BBH responses and 21\% to 43\% of the GYM responses contain no boxed answer. Also taking the last \texttt{<answer>} tag or ``Answer:'' line raises the two scores by at most 1.8 points and still leaves 13\% to 23\% of the BBH and 18\% to 43\% of the GYM responses without an answer. The stage-1 model gives a boxed answer in 82\% of its BBH and 67\% of its GYM responses, against 98\% and 96\% for Seq after the math stage. Read also from its \texttt{<answer>} tags and ``Answer:'' lines, the stage-1 model scores 61.20 on BBH and 40.08 on GYM instead of 54.14 and 34.25. The same reading changes the BBH and GYM scores of Seq, Replay-KL and Replay-KL + SATS after three stages by at most 0.55 points.

A sample of 360 responses from which no answer was extracted was read by an LLM (Claude), one agent per checkpoint reading each response once. The sample holds 40 responses per logic benchmark from each of Seq after five stages, Seq after the safety stage and \method after five stages. The LLM labelled each response as stating an answer, stating none, truncated, looping until truncation, or other (refusals included). It also extracted any answer stated in prose, and the official comparators then scored these answers. The comparators and the LLM could both judge 301 responses. On these, the comparators agree with the LLM's own verdict on 101 of 104 BBH, 74 of 81 GYM and 116 of 116 MuSR responses. Of the 360, 327 state an answer. Of the sampled responses of the two five-stage checkpoints, Seq and \method, the prose answer is right in 21\% on BBH, 15\% on GYM and 39\% on MuSR. With the same shares in every configuration, counting these answers would raise the Avg of each trained five-stage configuration by 0.523 (Replay-KL + OPRD + Recovery) to 0.825 points (Seq) and change the difference between any two of them by at most 0.302.

On MuSR, 2\% to 31\% of the five-stage responses have no ``ANSWER:'' line; after the math stage 99.7\% of the responses of Seq have none, and all but one of its 2,268 responses end with a boxed answer. An alternative reading takes the last boxed answer where no line with ``answer:'' names an option, keeps the official reading elsewhere, and counts the remaining responses at their expected chance score instead of a random option; the last change alone moves the stage-1 score from 46.74 to 46.79. Under this reading the shared stage-1 model scores 46.79. Seq scores 47.82 after the math stage and 48.74 after three stages, and Replay-KL + SATS and Replay-KL score 48.09 and 47.10 after three stages. After five stages, reading the box changes MuSR by $-0.10$ to +0.81 points per trained configuration and the five-stage MuSR differences of \cref{sec:q-signal} by less than 0.5. On the safe prompts of XSTest at most 0.4\% of the responses leave the reasoning unclosed; nearly all of the non-compliance is closed reasoning that the judge labels as refusal.

\paragraph{Overlap between training and evaluation prompts.} After lower-casing and collapsing white space, with symbols kept, no prompt of GYM (950), StrongREJECT (313), WildJailbreak (250), XSTest (450) or IFEval (541) is identical to or contained in a training prompt. The prompts checked are the student's 640 logic prompts and 672 stage-4 safety prompts. They also include the 56,339 instruction-following prompts of the Llama-Nemotron post-training data, which the model card of the logic expert Nemotron lists. The safety benchmarks share no 8-gram with the safety prompts, and no IFEval prompt shares half of its 13-grams with one training prompt. GYM and the logic prompts are drawn from the same 38 Reasoning Gym tasks with different seeds, so they share the 13-grams of the task templates. Of the 950 GYM items, 242 share at least 80\% of their 13-grams with one training prompt, although the instances themselves differ. Of these, 71 have the same answer as their most similar training prompt. Where several training prompts tie, the first in training order is used. Without these 242 items, GYM changes by $-0.93$ to +1.62 points across the 20 trained five-stage configurations of \cref{tab:five,tab:fiveone}. The ranking of these configurations on the reduced GYM has a Spearman rank correlation of 0.83 with their ranking on the full benchmark.

\paragraph{Evaluation noise.} After five stages, the pooled standard deviation across evaluation seeds has 38 degrees of freedom. The pool holds the trained configurations with three evaluation seeds (rollout seed 0), the uniform average of the five experts and the TIES merge of the five experts. In parentheses is the 95\% half-width for the difference between two configurations with three evaluation seeds each. The values are Avg 0.28 (0.46), the 11 benchmarks outside safety 0.34 (0.56), Avg$_{\mathrm{bal}}$ 0.28 (0.46), Avg with MBPP and MBPP+ counted once 0.27 (0.44), the first nine benchmarks 0.40 (0.66), logic 0.52 (0.86), math 0.66 (1.09), code 0.92 (1.52), safety 0.62 (1.02), instruction following 0.49 (0.81) and MuSR 1.61 (2.66). The logic SD is below a third of the MuSR SD because BBH and MuSR vary in opposite directions across evaluation seeds in this pool. After three stages, the pool holds the continual configurations of the three-stage prefix, with 26 degrees of freedom. The values are Avg 0.32 (0.54), logic 0.54 (0.90), math 0.67 (1.12), code 0.84 (1.40) and MuSR 1.25 (2.09). Each half-width is $t_{0.975,\nu}\, s \sqrt{\sum_i c_i^2 / n_i}$ for a contrast $\sum_i c_i \bar{x}_i$ of means over $n_i$ evaluation seeds, where $s$ is the pooled SD and $\nu$ its degrees of freedom. Between two configurations with one evaluation seed each, the half-width for their difference in five-stage Avg is 0.80.

\paragraph{Prompts and the Latent-MOPD reference.} Latent-MOPD does not give its prompts for MBPP, MuSR and Reasoning Gym. For these three, two or three prompt variants from public evaluation frameworks were compared on the base student and the three experts only. The variant closest to the scores that Latent-MOPD reports for these four models was fixed before any trained model was evaluated. Averaged over the nine benchmarks of the three-stage prefix, the local joint MOPD scores 52.01 against 52.02, the mean of the nine scores that Latent-MOPD reports for joint MOPD. The per-benchmark differences range from $-7.53$ (GYM) to +8.66 (MBPP) and are largest in size on MBPP, GYM and MuSR. The local Latent-MOPD scores 52.31 against 52.60 to 54.40 for the variants of Latent-MOPD reported in that paper \citep{fang2026latentmopd}.

\paragraph{MBPP and MBPP+ counted once.} MBPP and MBPP+ score the same 378 problems with the original and the extended tests. Averaging the two into one score gives a 13-benchmark average whose pooled standard deviation across evaluation seeds is 0.27 (half-width 0.44 for two configurations with three seeds). The pool of the five-stage evaluation noise consists of 19 rows of \cref{tab:five}. They are the trained configurations with three evaluation seeds (rollout seed 0), the uniform average of the five experts and the TIES merge of the five experts. Of the 171 pairwise comparisons between them, 3 change label: \method (shared budget) against Replay-KL with $b = 3$ becomes level ($-0.42$), and \method (shared budget) + Recovery and joint2 become 0.45 and 0.48 points above Seq + Recovery, beyond the noise. Two contrasts compare budgets in the gain of the three alignment terms over token-level replay with the same budget. The contrast between the shared budget and $b = 7$ stays beyond the noise (0.93, 95\% interval 0.30 to 1.55). The contrast between the shared budget and $b = 3$ becomes level (0.53, 95\% interval $-0.09$ to 1.16). Avg$_{11}$ and Avg$_{\mathrm{bal}}$ instead change the weight of safety; the main text names the averages under which each conclusion holds.

\section{Alignment terms and balancing}
\label{app:sats}

Reasoning-step regions are assigned once per rollout on the driver and shared by the student and the teacher, so both pool over the same queries. Queries are restricted to the last 2000 response tokens; keys run from the first real token to the last query, padding never enters a softmax, and positions beyond a row's causal range are zero. The pooled rows of \cref{eq:pool} are computed in query chunks, each recomputed in the backward pass, so the full attention matrix is never stored; grouped-query keys are repeated to the query heads, while the value relation is computed per key-value head. The teacher sends its pooled vectors per sample to the actor. Unit tests check the pooled rows against a naive full-softmax reference, and on real rollouts the batched result matches per-sample runs to within $2 \times 10^{-6}$ in fp32.

\begin{figure}[t]
\centering
\IfFileExists{figures/replaystep.pdf}{\includegraphics{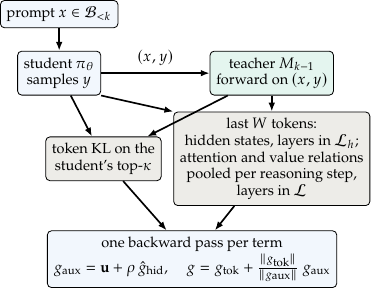}}{\begin{tikzpicture}[
  font=\scriptsize,
  box/.style={draw, rounded corners=2pt, align=center, inner sep=3pt, fill=vizBlue!6},
  tbox/.style={draw, rounded corners=2pt, align=center, inner sep=3pt, fill=vizAqua!12},
  lbox/.style={draw, rounded corners=2pt, align=center, inner sep=3pt, fill=vizGrid!60},
  arr/.style={-{Latex[length=1.5mm]}, thick},
]
\node[box] (x) at (0,0) {prompt $x \in \mathcal{B}_{<k}$};
\node[box] (s) at (0,-1.0) {student $\pi_\theta$\\samples $y$};
\node[tbox] (t) at (3.6,-1.0) {teacher $M_{k-1}$\\forward on $(x, y)$};
\node[lbox] (tok) at (0.75,-2.45) {token KL on the\\student's top-$\kappa$};
\node[lbox] (att) at (3.6,-2.45) {last $W$ tokens:\\hidden states, layers in $\mathcal{L}_{h}$;\\attention and value relations\\pooled per reasoning step,\\layers in $\mathcal{L}$};
\node[box] (bal) at (2.25,-4.15) {one backward pass per term\\$g_{\mathrm{aux}} = \mathbf{u} + \rho\, \hat{g}_{\mathrm{hid}}$, \quad $g = g_{\mathrm{tok}} + \frac{\lVert g_{\mathrm{tok}}\rVert}{\lVert g_{\mathrm{aux}}\rVert}\, g_{\mathrm{aux}}$};
\draw[arr] (x) -- (s);
\draw[arr] (s) -- node[above] {$(x, y)$} (t);
\draw[arr] (s) -- (tok);
\draw[arr] (t) -- (tok);
\draw[arr] (t) -- (att);
\draw[arr] (s.south east) -- (att.north west);
\draw[arr] (tok) -- (bal);
\draw[arr] (att) -- (bal);
\end{tikzpicture}}
\caption{One replay step of \method. The student answers a prompt of an earlier domain; the previous-stage student scores the answer at the token level and provides hidden-state targets and region-pooled attention and value-relation targets; each term has its own backward pass, and the alignment gradients are combined by \cref{eq:joint} and balanced against the token gradient by \cref{eq:balance} before the optimizer step.}
\label{fig:replaystep}
\end{figure}

On a replay step with one alignment term the actor runs the micro-batches twice. The first pass accumulates $g_{\mathrm{tok}}$, which is saved; the second accumulates the auxiliary gradient. Norms and the inner product are reduced over all FSDP shards and ranks, the combination of \cref{eq:balance} is written into the parameter gradients, and the optimizer step follows. For \method each of the three terms has its own pass after the token pass (\cref{fig:replaystep}): the SATS and value-relation passes give $\mathbf{u}$ of \cref{eq:joint}, which is kept in host memory while the hidden-state pass runs, and $g_{\mathrm{aux}} = \mathbf{u} + \rho\,\hat{g}_{\mathrm{hid}}$ is then scaled to the norm of the token gradient. The 1:1:1 variants of \cref{tab:three} instead balance each added term to the token gradient alone and add it into the saved buffer before the next term's pass. New-domain steps run a single pass.

\paragraph{A worked example.} On a replay step of a two-step test run of \method in the code stage, the token gradient has norm 26.35. The raw SATS and value-relation gradients have norms 0.923 and 0.268 with cosine 0.248. The sum of their unit vectors, $\hat{g}_{\mathrm{SATS}} + \hat{g}_{\mathrm{VR}}$, therefore has norm $1.580 = \sqrt{2 + 2 \times 0.248}$. The raw hidden-state gradient has norm $7.9 \times 10^{-5}$ and cosine 0.575 with $\mathbf{u}$. With $\rho = 2$, this gives $\lVert \mathbf{u} + 2\,\hat{g}_{\mathrm{hid}} \rVert = 2.702 = \sqrt{5 + 4 \times 0.575}$. The logged scale $c = \lVert g_{\mathrm{tok}} \rVert / \lVert g_{\mathrm{aux}} \rVert$ of \cref{eq:balance} is 9.755 (26.35 / 2.702 up to rounding). The pooled part thus enters with norm 9.755 and the hidden-state part with 19.51, and their sum has the norm of the token gradient. Each sample has about 95 reasoning-step regions in the window. For comparison, on the replay steps of stages 2 and 3 the raw hidden-state gradient of Replay-KL + OPRD has norms between $1.8 \times 10^{-5}$ and $9.5 \times 10^{-5}$, and the SATS gradient of Replay-KL + SATS between 0.03 and 0.43.

\section{Gradient ratio}
\label{app:ratio}

\begin{figure}[t]
\centering
\IfFileExists{figures/ratio.pdf}{\includegraphics{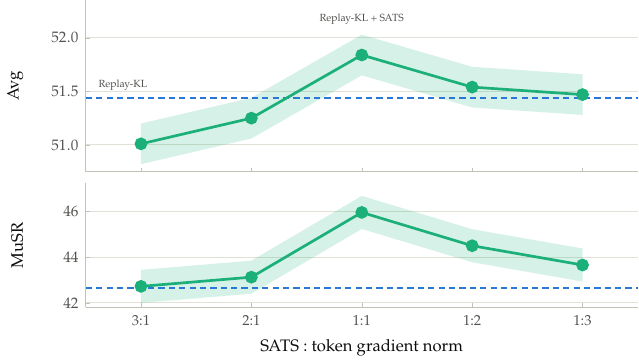}}{\begin{tikzpicture}
\pgfplotsset{
  ratiopanel/.style={
    vizaxis, scale only axis, width=0.8\columnwidth,
    xmin=0.5, xmax=5.5, xtick={1,2,3,4,5},
    ylabel style={font=\footnotesize, at={(axis description cs:-0.1,0.5)}, anchor=south},
  },
}
\begin{axis}[
  ratiopanel, name=avgpanel, height=2.9cm,
  ymin=50.75, ymax=52.35, ytick={51,51.5,52}, xticklabels={},
  yticklabel style={/pgf/number format/fixed, /pgf/number format/precision=1, /pgf/number format/fixed zerofill},
  ylabel={Avg},
]
\addplot[mKL, densely dashed, line width=1pt, domain=0.5:5.5, samples=2] {51.44};
\node[font=\tiny, text=vizInkTwo, anchor=south west] at (axis cs:0.55,51.44) {Replay-KL};
\addplot[name path=au, draw=none, forget plot] coordinates {(1,51.20) (2,51.44) (3,52.03) (4,51.73) (5,51.66)};
\addplot[name path=al, draw=none, forget plot] coordinates {(1,50.82) (2,51.06) (3,51.65) (4,51.35) (5,51.28)};
\addplot[mSATS, fill opacity=0.18, draw=none, forget plot] fill between[of=au and al];
\addplot[mSATS, mark=*, mark size=2.4pt] coordinates {(1,51.01) (2,51.25) (3,51.84) (4,51.54) (5,51.47)};
\node[font=\tiny, text=vizInkTwo, anchor=south] at (axis cs:3,52.06) {Replay-KL + SATS};
\end{axis}
\begin{axis}[
  ratiopanel, at={($(avgpanel.south west)+(0,-0.2cm)$)}, anchor=north west, height=2.1cm,
  ymin=41.8, ymax=47.2, ytick={42,44,46},
  xticklabels={3:1, 2:1, 1:1, 1:2, 1:3},
  xlabel={SATS : token gradient norm},
  ylabel={MuSR},
]
\addplot[mKL, densely dashed, line width=1pt, domain=0.5:5.5, samples=2] {42.64};
\addplot[name path=mu, draw=none, forget plot] coordinates {(1,43.44) (2,43.84) (3,46.66) (4,45.21) (5,44.37)};
\addplot[name path=ml, draw=none, forget plot] coordinates {(1,42.00) (2,42.40) (3,45.22) (4,43.77) (5,42.93)};
\addplot[mSATS, fill opacity=0.18, draw=none, forget plot] fill between[of=mu and ml];
\addplot[mSATS, mark=*, mark size=2.4pt] coordinates {(1,42.72) (2,43.12) (3,45.94) (4,44.49) (5,43.65)};
\end{axis}
\end{tikzpicture}}
\caption{Nine-benchmark average (top) and MuSR (bottom) of Replay-KL + SATS on the three-stage prefix. The horizontal axis is the norm ratio of the SATS part to the token part of the update on replay steps. This ratio is the inverse of the target ratio $r$ of \cref{eq:balance}. The ratios 3:1 and 2:1 are $r = 0.33$ and $0.5$, which scale the token gradient to a third and a half of its norm. The ratios 1:2 and 1:3 are $r = 2$ and $3$. Points are means over three evaluation seeds, with the pooled standard error of each point as shaded bands (\cref{app:eval}). Dashed lines: Replay-KL, which has no SATS term. MuSR is scored with the benchmark's parser (\cref{app:eval}).}
\label{fig:ratio}
\end{figure}

\Cref{fig:ratio} sweeps the balancing ratio of Replay-KL + SATS on the three-stage prefix. Equal norms give the highest nine-benchmark average. Under the benchmark's parser, the nine-benchmark average falls below its value at equal norms by 0.82 at $r = 0.33$ and by 0.59 at $r = 0.5$, beyond the evaluation noise. It falls by 0.29 at $r = 2$ and 0.37 at $r = 3$, within the noise. With the SATS part at two or three times the norm of the scaled token part ($r = 0.5$ and $0.33$), the nine-benchmark average ends at 51.25 and 51.01, level with Replay-KL (51.44). The logic average lies below its value at equal norms beyond the noise at $r = 0.33$ and $0.5$ (by 1.32 and 1.39 points) and within it at $r = 2$ and $3$. MuSR follows the same shape as the nine-benchmark average: 42.72, 43.12, 45.94, 44.49 and 43.65 at $r = 0.33$, $0.5$, $1$, $2$ and $3$. On both sides of equal norms it moves back toward the Replay-KL level (42.64). Under the benchmark's parser, only equal norms raise it above Replay-KL beyond the noise (+3.31; the +1.85 at $r = 2$ is within it). With MuSR read from the boxed answer (\cref{app:eval}), the other ratios end 0.21 to 0.34 below equal norms in the nine-benchmark average and within 0.43 of them in the logic average, all within the noise.

\section{Pooling regions}
\label{app:pooling}

WAAD chunks \citep{li2026attention} partition a response differently from reasoning steps. When the teacher recuts the regions into WAAD chunks, only 21\% (stage 2) and 17\% (stage 3) of the reasoning-step boundaries remain region boundaries. On the three-stage prefix, pooling over reasoning steps and over WAAD chunks gives level nine-benchmark averages (51.84 and 51.74, 0.10 apart). MuSR is 1.28 points higher with reasoning steps. Both differences are within the evaluation noise.

\section{Baseline details}
\label{app:baselines}

\paragraph{Seq + Recovery.} The recovery teachers are the stage-final students $M_1, \dots, M_K$, following the text of GLM-5 (\S3.5 of its report) and of ACLArena. The released ACLArena recovery script routes each domain to an independently trained expert, the variant reported here as expert recovery (\cref{sec:q-recovery}). The recovery stage starts from $M_K$, uses the token-level loss of \cref{sec:replay}, and routes each domain's prompts to its stage-final student. For the last domain this teacher is $M_K$ itself, so the last stage is not erased by recovery. It has 21 steps after three stages and 35 after five. X + Recovery applies the same stage after a replay method X, with X's own stage-final students as teachers.

\paragraph{Replay-KL + SEEKR.} The port keeps the loss of SEEKR's released code \citep{he2024seekr}. It is the forward KL divergence $\mathrm{KL}(M_{k-1} \,\|\, \pi_\theta)$ between the attention rows of the previous-stage student and of the student on the selected heads, averaged per layer and then over layers. The SEEKR paper writes it as a sum over the selected heads and queries. Three changes follow from the setting. Head sensitivity is the squared gradient of the token-level distillation loss with respect to the attention probabilities, because no gold answers exist; SEEKR uses the cross-entropy on gold answers. Query rows are drawn from the response only (100 per response, uniformly at random). Heads are reselected after every stage from the importance (task sensitivity times forgettability) accumulated so far. The selection keeps SEEKR's budget ratios: 42 of the 336 heads, chosen within 21 of the 28 layers. After the logic stage the 42 selected heads lie in layers 0, 14, 16, 17, 19 to 25 and 27. The term is balanced to the token gradient at 1:1, like Replay-KL + SATS.

\paragraph{Merges.} The TIES merges follow the reference implementation \citep{yadav2023ties}. A task vector is the difference between a model's weights and those of the base model, the student. Each task vector keeps the entries whose magnitude reaches the 80th percentile of its magnitudes over the whole model, ties at that threshold included. For each parameter, the elected sign is that of the sum of its kept entries over the task vectors, or the overall majority sign where that sum is zero. The kept entries that agree with this sign are averaged, and the average, scaled by $\lambda = 1$, is added to the base model. TIES applied stage by stage takes the logic expert as the first merge and at each later stage merges the task vectors of the previous merge and of the stage's expert, each trimmed anew; every merge is computed in fp32 and stored in bf16. The instruction-following expert changes fewer than 20\% of its entries, so trimming keeps all of its changes. The uniform average of the experts is the mean of their weights. Relative to the student's weights, the task vectors of the logic, math, code, safety and instruction-following experts have norms of 0.32\%, 0.50\%, 0.25\%, 0.16\% and 0.04\% and change 99.95\%, 66\%, 54\%, 44\% and 17\% of the entries.

\section{Full results}
\label{app:full}

\paragraph{Three-stage prefix.} The two tables below give the configurations of the three-stage prefix, each with three evaluation seeds; the first also lists the student, the three experts and the two joint references (\cref{tab:threebench}).

\begin{center}\begin{minipage}{\linewidth}\centering\captionsetup{hypcap=false}
\begin{minipage}{\linewidth}
\centering
\scriptsize
\setlength{\tabcolsep}{3pt}
\begin{tabular}{l ccc ccc ccc c}
\toprule
Method & BBH & GYM & MuSR & AIME24 & AIME25 & Minerva & LCB-e & MBPP & MBPP+ & Avg \\
\midrule
Student & 38.89 & 22.60 & 39.68 & 26.25 & 23.96 & 28.31 & 52.99 & 57.41 & 48.02 & 37.57 \\
JustRL (math expert) & 44.35 & 28.50 & 45.77 & 51.04 & 37.92 & 33.36 & 64.93 & 65.08 & 55.16 & 47.35 \\
Archer2.0 (code expert) & 39.72 & 24.12 & 45.24 & 41.04 & 28.12 & 30.51 & 73.51 & 69.44 & 57.41 & 45.46 \\
Nemotron (logic expert) & 64.17 & 43.83 & 46.43 & 52.29 & 35.21 & 34.01 & 76.87 & 56.35 & 45.90 & 50.56 \\
Joint MOPD & 60.72 & 44.27 & 44.20 & 46.61 & 34.43 & 33.85 & 71.64 & 72.06 & 60.32 & 52.01 \\
Joint Latent-MOPD & 60.93 & 43.52 & 44.15 & 48.54 & 35.94 & 34.51 & 71.55 & 71.66 & 59.95 & 52.31 \\
\midrule
Seq & 60.31 & 30.55 & 39.33 & 45.49 & 32.85 & 33.49 & 68.91 & 70.86 & 60.14 & 49.10 \\
Seq + Recovery & 62.07 & 41.30 & 39.86 & 43.82 & 35.21 & \textbf{34.53} & 67.04 & 70.19 & 58.47 & 50.28 \\
Replay-KL & 62.01 & 40.91 & 42.64 & \textbf{47.22} & \textbf{35.28} & 33.58 & 70.52 & 70.86 & 59.96 & 51.44 \\
Replay-KL + SEEKR & 63.33 & 41.73 & 45.55 & 43.61 & 33.89 & 33.92 & 71.14 & 70.15 & 59.13 & 51.38 \\
Replay-KL + OPRD & 63.61 & 41.89 & 44.14 & 46.74 & 33.68 & 33.49 & \textbf{73.76} & 71.08 & 60.23 & \textbf{52.07} \\
Replay-KL + SATS & 63.46 & 41.16 & 45.94 & 46.04 & 34.51 & 34.22 & 70.77 & 70.72 & 59.70 & 51.84 \\
Replay-KL + SATS + OPRD & 63.64 & \textbf{42.40} & 44.89 & 45.49 & 33.33 & 33.33 & 73.13 & 71.12 & 60.27 & 51.96 \\
Replay-KL + SATS + value relation & \textbf{63.98} & 42.35 & \textbf{46.16} & 45.00 & 32.43 & 33.98 & 71.64 & 70.68 & \textbf{60.80} & 51.89 \\
Replay-KL + SATS, WAAD regions & 63.12 & 41.99 & 44.66 & 46.46 & 34.17 & 34.47 & 70.77 & 70.63 & 59.39 & 51.74 \\
Replay-KL + SATS, $r=0.33$ & 62.78 & 41.10 & 42.72 & 44.93 & 34.31 & 34.28 & 69.78 & 70.11 & 59.13 & 51.01 \\
Replay-KL + SATS, $r=0.5$ & 62.50 & 40.77 & 43.12 & 46.25 & 34.51 & 33.98 & 69.78 & 70.90 & 59.44 & 51.25 \\
Replay-KL + SATS, $r=2$ & 63.83 & 41.85 & 44.49 & 46.39 & 34.38 & 33.92 & 66.92 & \textbf{71.96} & 60.19 & 51.54 \\
Replay-KL + SATS, $r=3$ & 62.87 & 41.69 & 43.65 & 46.60 & 34.93 & 33.73 & 69.28 & 70.99 & 59.48 & 51.47 \\
\bottomrule
\end{tabular}
\captionof{table}{Three-stage prefix, scores per benchmark after the code stage and the nine-benchmark Avg. Continual rows are means over three evaluation seeds, and student and expert rows are single evaluations. Joint rows are means of four evaluations of the same checkpoint (eight on MuSR). Bold marks the best continual configuration in each column.}
\label{tab:threebench}
\end{minipage}
\end{minipage}\end{center}

\begin{center}\begin{minipage}{\linewidth}\centering\captionsetup{hypcap=false}
\small
\setlength{\tabcolsep}{4pt}
\begin{tabular}{l ccc ccc cc}
\toprule
& \multicolumn{3}{c}{Domain average} & \multicolumn{3}{c}{Avg per evaluation seed} & \multicolumn{2}{c}{Normalized score} \\
\cmidrule(lr){2-4} \cmidrule(lr){5-7} \cmidrule(lr){8-9}
Method & Logic & Math & Code & seed 0 & seed 1 & seed 2 & \Norm & \NormD \\
\midrule
Seq & 43.40 & 37.27 & 66.63 & 49.26 & 48.85 & 49.19 & 0.720 & 0.765 \\
Seq + Recovery & 47.74 & 37.85 & 65.23 & 50.78 & 49.93 & 50.12 & 0.793 & 0.826 \\
Replay-KL & 48.52 & 38.69 & 67.11 & 51.81 & 51.63 & 50.89 & 0.866 & 0.905 \\
Replay-KL + SEEKR & 50.20 & 37.14 & 66.81 & 51.07 & 51.89 & 51.19 & 0.862 & 0.894 \\
Replay-KL + OPRD & 49.88 & 37.97 & 68.35 & 51.92 & 52.14 & 52.14 & 0.905 & 0.943 \\
Replay-KL + SATS & 50.19 & 38.26 & 67.06 & 52.15 & 51.53 & 51.83 & 0.890 & 0.925 \\
Replay-KL + SATS + OPRD & 50.31 & 37.38 & 68.18 & 51.90 & 51.79 & 52.18 & 0.898 & 0.934 \\
Replay-KL + SATS + value relation & 50.83 & 37.14 & 67.71 & 51.98 & 52.03 & 51.66 & 0.894 & 0.927 \\
Replay-KL + SATS, WAAD regions & 49.92 & 38.36 & 66.93 & 51.45 & 51.75 & 52.02 & 0.884 & 0.919 \\
Replay-KL + SATS, $r=0.33$ & 48.87 & 37.84 & 66.34 & 51.02 & 50.87 & 51.16 & 0.839 & 0.873 \\
Replay-KL + SATS, $r=0.5$ & 48.80 & 38.25 & 66.70 & 50.91 & 51.23 & 51.60 & 0.854 & 0.890 \\
Replay-KL + SATS, $r=2$ & 50.06 & 38.23 & 66.35 & 51.57 & 51.37 & 51.70 & 0.872 & 0.905 \\
Replay-KL + SATS, $r=3$ & 49.40 & 38.42 & 66.58 & 51.07 & 52.00 & 51.33 & 0.867 & 0.902 \\
\bottomrule
\end{tabular}
\captionof{table}{Three-stage prefix: domain averages, the Avg of each evaluation seed, and \Norm and \NormD, with domain averages and normalized scores as means over the three evaluation seeds.}
\label{tab:threeseeds}
\end{minipage}\end{center}
\noindent \Norm is defined in \cref{eq:norm}, and the column \NormD is the per-domain score \citep{ma2026mopd}. \NormD rescales each domain average so that the student scores 0 and that domain's expert 1 (logic: Nemotron, math: JustRL, code: Archer2.0), then averages over the three domains. Computed from their printed rows in \cref{tab:threebench}, it is 0.940 for Joint MOPD and 0.960 for Joint Latent-MOPD. The student and expert rows of \cref{tab:threebench} are single evaluations, and the joint rows are means of four evaluations of the same checkpoint (eight on MuSR).

\paragraph{Scores after each stage.}
\label{app:stages}
\Cref{tab:stages} gives Seq and the shared-budget runs Replay-KL, Replay-KL + SEEKR, Replay-KL + SATS and Replay-KL + OPRD after stages 2 to 5. It lists the logic and math benchmarks after the math stage, from which the backward transfer of math in \cref{tab:bwt} is measured. It also lists BBH after the safety stage and the average over the first nine benchmarks after stages 3 to 5, which \cref{fig:oldnine} draws. The stage-1 model, shared by all methods, scores 54.14 on BBH, 34.25 on GYM and 46.74 on MuSR.

\begin{center}\begin{minipage}{\linewidth}\centering\captionsetup{hypcap=false}
\small
\setlength{\tabcolsep}{3.5pt}
\begin{tabular}{l ccc ccc c cc c}
\toprule
& \multicolumn{6}{c}{Stage 2 (math)} & Stage 3 & \multicolumn{2}{c}{Stage 4} & Stage 5 \\
\cmidrule(lr){2-7} \cmidrule(lr){9-10}
Method & BBH & GYM & MuSR & AIME24 & AIME25 & Minerva & Avg$_9$ & Avg$_9$ & BBH & Avg$_9$ \\
\midrule
Seq & 63.15 & 43.24 & 35.58 & 42.08 & 33.40 & \textbf{34.41} & 49.10 & 39.52 & 31.85 & 45.51 \\
Replay-KL & \textbf{64.32} & 42.77 & 40.04 & 43.68 & 33.75 & 33.98 & 51.44 & 48.86 & 51.20 & 48.31 \\
Replay-KL + SEEKR & 63.77 & \textbf{43.69} & 38.58 & 43.82 & 33.40 & 33.52 & 51.38 & n/a & n/a & n/a \\
Replay-KL + SATS & 63.86 & 42.30 & 39.20 & 43.68 & \textbf{34.10} & 34.19 & 51.84 & 48.20 & 49.35 & 48.34 \\
Replay-KL + OPRD & 63.98 & 43.61 & \textbf{40.61} & \textbf{44.17} & 33.33 & 34.22 & \textbf{52.07} & \textbf{49.20} & \textbf{53.98} & \textbf{49.17} \\
\bottomrule
\end{tabular}
\captionof{table}{Seq and the shared-budget runs Replay-KL, Replay-KL + SEEKR, Replay-KL + SATS and Replay-KL + OPRD after stages 2 to 5. Columns are scores after the math stage, BBH after the safety stage, and Avg$_9$, the average over the first nine benchmarks, after stages 3, 4 and 5. Means over three evaluation seeds, except after stage 4: two seeds for Seq and evaluation seed 0 for the replay methods. Replay-KL + SEEKR was run on the three-stage prefix only. Bold marks the highest value in each column.}
\label{tab:stages}
\end{minipage}\end{center}

\paragraph{Five stages.}
\label{app:fivefull}
The next three tables give the scores after the fifth stage, including the 70-step recovery, joint MOPD with 208 steps and the four merges without training. Scores are means over three evaluation seeds, and the student and expert rows of the first two tables are single evaluations. XSTest (safe) is the compliance rate on the safe prompts, which enters the safety average; XSTest (unsafe) is the safe rate on the unsafe prompts. The fourth table gives the four configurations trained with a second rollout seed.

\begin{center}\begin{minipage}{\linewidth}\centering\captionsetup{hypcap=false}
\scriptsize
\setlength{\tabcolsep}{2.5pt}
\begin{tabular}{l ccc ccc ccc}
\toprule
Method & BBH & GYM & MuSR & AIME24 & AIME25 & Minerva & LCB-e & MBPP & MBPP+ \\
\midrule
Student & 38.89 & 22.60 & 39.68 & 26.25 & 23.96 & 28.31 & 52.99 & 57.41 & 48.02 \\
Nemotron (logic expert) & 64.17 & 43.83 & 46.43 & 52.29 & 35.21 & 34.01 & 76.87 & 56.35 & 45.90 \\
JustRL (math expert) & 44.35 & 28.50 & 45.77 & 51.04 & 37.92 & 33.36 & 64.93 & 65.08 & 55.16 \\
Archer2.0 (code expert) & 39.72 & 24.12 & 45.24 & 41.04 & 28.12 & 30.51 & 73.51 & 69.44 & 57.41 \\
STAR1 (safety expert) & 25.83 & 20.00 & 43.12 & 26.88 & 22.71 & 28.12 & 48.13 & 56.48 & 47.49 \\
UltraData-IF (IF expert) & 40.56 & 23.27 & 40.21 & 30.21 & 23.96 & 28.68 & 55.60 & 60.19 & 50.40 \\
\midrule
Seq & 53.43 & 32.47 & 42.20 & 38.82 & 28.61 & 30.55 & 61.44 & 66.27 & 55.78 \\
Replay-KL & 56.36 & 37.53 & 43.78 & 41.81 & 32.22 & 31.92 & 66.67 & 67.99 & 56.48 \\
Replay-KL + SATS & 55.28 & 36.79 & 43.17 & 41.25 & 33.40 & 31.83 & 65.92 & 69.27 & 58.11 \\
Replay-KL + OPRD & 56.82 & 36.63 & 44.75 & 44.93 & 30.83 & 32.63 & 68.16 & 69.36 & 58.47 \\
\method (shared budget) & 55.49 & 33.86 & 45.72 & 44.10 & 32.29 & 31.89 & 70.02 & 69.58 & 59.22 \\
Replay-KL, $b = 3$ & 57.07 & 35.42 & 44.27 & 46.60 & 33.96 & 33.58 & 64.30 & 71.56 & 60.10 \\
\method, $b = 3$ & 55.37 & 35.93 & 43.52 & 47.08 & 34.79 & 32.60 & 71.14 & 70.81 & 59.08 \\
Replay-KL (fixed share) & 57.72 & 36.80 & 42.95 & 46.46 & \textbf{34.86} & 33.79 & 73.01 & 70.41 & 59.13 \\
\method & \textbf{59.07} & 37.56 & \textbf{46.03} & 46.94 & 33.75 & 33.82 & 69.78 & 70.72 & 59.13 \\
Seq + Recovery & 57.07 & 37.79 & 43.03 & 44.65 & 33.89 & 33.36 & 70.40 & 70.59 & 58.64 \\
Seq + Recovery-70 & 58.27 & 37.45 & 42.64 & 44.38 & 33.61 & 33.67 & 70.02 & 69.89 & 58.77 \\
Replay-KL + Recovery & 58.70 & 39.16 & 41.98 & \textbf{47.71} & 34.72 & 34.01 & 71.02 & 69.75 & 58.29 \\
Replay-KL + SATS + Recovery & 58.24 & \textbf{39.53} & 45.72 & 46.11 & 33.89 & \textbf{34.28} & 69.40 & 69.66 & 58.47 \\
Replay-KL + OPRD + Recovery & 58.77 & 38.34 & 43.74 & 46.81 & 34.44 & 32.97 & 73.01 & 70.46 & 59.39 \\
\method (shared budget) + Recovery & 55.74 & 36.27 & 44.84 & 44.79 & 32.99 & 33.30 & \textbf{74.50} & 70.19 & 59.30 \\
\method + Recovery & 58.77 & 37.65 & 44.62 & 47.08 & 32.57 & 33.24 & 72.76 & \textbf{72.31} & \textbf{60.41} \\
Uniform average of the five experts & 52.81 & 27.90 & 44.18 & 42.08 & 30.62 & 33.03 & 63.43 & 66.67 & 55.73 \\
TIES, stage by stage & 64.91 & 44.00 & 48.46 & 54.93 & 37.85 & 35.63 & 78.11 & 70.11 & 57.80 \\
Joint MOPD, 208 steps & 56.85 & 36.80 & 45.19 & 47.15 & 34.65 & 32.90 & 69.78 & 69.44 & 58.16 \\
TIES of the Seq snapshots & 66.39 & 45.92 & 37.83 & 45.21 & 32.36 & 33.73 & 70.40 & 70.06 & 59.44 \\
TIES merge of the five experts & 65.40 & 43.99 & 48.32 & 54.58 & 37.64 & 35.72 & 76.24 & 69.58 & 58.33 \\
\bottomrule
\end{tabular}
\captionof{table}{Five stages: scores per benchmark on logic, math and code. Student and expert rows are single evaluations; the other rows are means over three evaluation seeds. Bold marks the best continual configuration in each column.}
\label{tab:fivebench1}
\end{minipage}\end{center}

\begin{center}\begin{minipage}{\linewidth}\centering\captionsetup{hypcap=false}
\scriptsize
\setlength{\tabcolsep}{3pt}
\begin{tabular}{l ccc c cc c}
\toprule
& & & \multicolumn{2}{c}{XSTest} & & & \\
\cmidrule(lr){4-5}
Method & StrongREJECT & WildJailbreak & safe & unsafe & IFEval & IFBench & Avg \\
\midrule
Student & 12.86 & 14.40 & 89.40 & 24.2 & 36.04 & 9.33 & 35.72 \\
Nemotron (logic expert) & 22.92 & 23.80 & 95.40 & 45.2 & 68.11 & 19.25 & 48.90 \\
JustRL (math expert) & 18.93 & 14.40 & 97.90 & 34.0 & 41.64 & 12.50 & 43.68 \\
Archer2.0 (code expert) & 16.37 & 16.50 & 94.20 & 30.2 & 41.77 & 10.33 & 42.02 \\
STAR1 (safety expert) & 94.89 & 67.90 & 25.20 & 98.0 & 23.29 & 9.08 & 38.51 \\
UltraData-IF (IF expert) & 17.81 & 12.00 & 93.40 & 33.5 & 55.50 & 14.08 & 38.99 \\
\midrule
Seq & 12.17 & 11.00 & \textbf{94.73} & 28.4 & 55.62 & 17.06 & 42.87 \\
Replay-KL & 48.27 & 14.60 & 83.10 & 71.5 & 55.30 & 17.28 & 46.66 \\
Replay-KL + SATS & 53.35 & 15.63 & 80.97 & 75.2 & 53.90 & 16.39 & 46.80 \\
Replay-KL + OPRD & 64.56 & 16.73 & 70.93 & 84.9 & \textbf{56.07} & 17.06 & 47.71 \\
\method (shared budget) & 59.74 & 15.97 & 76.27 & 80.9 & 55.11 & \textbf{17.47} & 47.62 \\
Replay-KL, $b = 3$ & 84.64 & 27.73 & 42.87 & 96.5 & 55.53 & 16.00 & 48.12 \\
\method, $b = 3$ & 90.52 & 33.97 & 31.53 & 97.3 & 53.94 & 16.78 & 48.36 \\
Replay-KL (fixed share) & 91.13 & 42.07 & 26.83 & 97.9 & 53.64 & 16.56 & 48.95 \\
\method & 90.10 & 40.37 & 26.33 & 96.9 & 54.28 & 16.81 & 48.91 \\
Seq + Recovery & 95.23 & 44.53 & 20.97 & 98.8 & 53.34 & 17.44 & 48.64 \\
Seq + Recovery-70 & 95.37 & 49.97 & 13.40 & \textbf{99.0} & 52.82 & 16.61 & 48.35 \\
Replay-KL + Recovery & 95.63 & 48.53 & 17.83 & 98.6 & 53.65 & 16.92 & 49.14 \\
Replay-KL + SATS + Recovery & \textbf{95.95} & 49.73 & 16.57 & 98.6 & 53.48 & 17.08 & 49.15 \\
Replay-KL + OPRD + Recovery & 95.93 & 49.37 & 16.47 & 98.7 & 53.16 & 16.56 & \textbf{49.24} \\
\method (shared budget) + Recovery & 95.79 & \textbf{53.17} & 15.97 & 98.7 & 53.11 & 16.94 & 49.07 \\
\method + Recovery & 92.78 & 44.57 & 23.20 & 97.3 & 52.66 & 16.03 & 49.19 \\
Uniform average of the five experts & 43.10 & 17.63 & 89.83 & 61.9 & 50.54 & 13.47 & 45.07 \\
TIES, stage by stage & 19.60 & 17.17 & 97.30 & 37.9 & 62.85 & 17.17 & 50.42 \\
Joint MOPD, 208 steps & 92.07 & 45.13 & 28.77 & 96.8 & 53.88 & 15.53 & 49.02 \\
TIES of the Seq snapshots & 39.43 & 15.70 & 90.63 & 61.7 & 46.89 & 14.33 & 47.74 \\
TIES merge of the five experts & 58.63 & 19.13 & 86.10 & 75.0 & 61.37 & 16.97 & 52.29 \\
\bottomrule
\end{tabular}
\captionof{table}{Five stages: scores per benchmark on safety and instruction following, and the 14-benchmark Avg; XSTest safe is the compliance rate and XSTest unsafe the safe rate. Student and expert rows are single evaluations; the other rows are means over three evaluation seeds. Bold marks the best continual configuration in each column.}
\label{tab:fivebench2}
\end{minipage}\end{center}

\begin{center}\begin{minipage}{\linewidth}\centering\captionsetup{hypcap=false}
\scriptsize
\setlength{\tabcolsep}{2.5pt}
\begin{tabular}{l ccccc ccc cc}
\toprule
& \multicolumn{5}{c}{Domain average} & \multicolumn{3}{c}{Avg per evaluation seed} & \multicolumn{2}{c}{Normalized score} \\
\cmidrule(lr){2-6} \cmidrule(lr){7-9} \cmidrule(lr){10-11}
Method & Logic & Math & Code & Safety & IF & seed 0 & seed 1 & seed 2 & \Norm & \NormD \\
\midrule
Seq & 42.70 & 32.66 & 61.16 & 39.30 & 36.34 & 42.66 & 42.99 & 42.95 & 0.303 & 0.539 \\
Replay-KL & 45.89 & 35.32 & 63.71 & 48.66 & 36.29 & 46.80 & 46.79 & 46.40 & 0.464 & 0.725 \\
Replay-KL + SATS & 45.08 & 35.49 & 64.43 & 49.98 & 35.14 & 46.40 & 47.22 & 46.80 & 0.470 & 0.721 \\
Replay-KL + OPRD & 46.07 & 36.13 & 65.33 & 50.74 & 36.56 & 47.38 & 47.67 & 48.08 & 0.508 & 0.784 \\
\method (shared budget) & 45.03 & 36.09 & 66.27 & 50.66 & 36.29 & 47.30 & 47.69 & 47.89 & 0.504 & 0.780 \\
Replay-KL, $b = 3$ & 45.59 & 38.04 & 65.32 & 51.75 & 35.76 & 48.16 & 48.05 & 48.13 & 0.525 & 0.800 \\
\method, $b = 3$ & 44.94 & 38.16 & 67.01 & 52.01 & 35.36 & 48.39 & 48.62 & 48.07 & 0.536 & 0.814 \\
Replay-KL (fixed share) & 45.82 & 38.37 & 67.52 & 53.34 & 35.10 & 48.70 & 49.22 & 48.93 & 0.561 & 0.840 \\
\method & 47.55 & 38.17 & 66.54 & 52.27 & 35.54 & 48.74 & 48.72 & 49.26 & 0.559 & 0.842 \\
Seq + Recovery & 45.96 & 37.30 & 66.54 & 53.58 & 35.39 & 48.54 & 48.92 & 48.46 & 0.547 & 0.820 \\
Seq + Recovery-70 & 46.12 & 37.22 & 66.23 & 52.91 & 34.71 & 48.57 & 48.32 & 48.15 & 0.535 & 0.800 \\
Replay-KL + Recovery & 46.61 & 38.81 & 66.35 & 54.00 & 35.28 & 49.37 & 48.81 & 49.23 & 0.569 & 0.847 \\
Replay-KL + SATS + Recovery & 47.83 & 38.09 & 65.84 & 54.08 & 35.28 & 49.19 & 49.22 & 49.05 & 0.569 & 0.845 \\
Replay-KL + OPRD + Recovery & 46.95 & 38.07 & 67.62 & 53.92 & 34.86 & 49.49 & 49.48 & 48.76 & 0.573 & 0.851 \\
\method (shared budget) + Recovery & 45.62 & 37.03 & 68.00 & 54.98 & 35.03 & 48.87 & 49.16 & 49.17 & 0.566 & 0.839 \\
\method + Recovery & 47.01 & 37.63 & 68.49 & 53.52 & 34.35 & 49.47 & 49.05 & 49.05 & 0.571 & 0.847 \\
Uniform average of the five experts & 41.63 & 35.25 & 61.94 & 50.19 & 32.01 & 45.08 & 44.99 & 45.15 & 0.396 & 0.593 \\
TIES, stage by stage & 52.46 & 42.80 & 68.67 & 44.69 & 40.01 & 50.77 & 50.40 & 50.09 & 0.623 & 1.001 \\
Joint MOPD, 208 steps & 46.28 & 38.24 & 65.79 & 55.32 & 34.70 & 49.66 & 48.76 & 48.65 & 0.564 & 0.829 \\
TIES of the Seq snapshots & 50.05 & 37.10 & 66.63 & 48.59 & 30.61 & 48.00 & 47.65 & 47.56 & 0.509 & 0.744 \\
TIES merge of the five experts & 52.57 & 42.65 & 68.05 & 54.62 & 39.17 & 52.21 & 52.40 & 52.25 & 0.702 & 1.061 \\
\bottomrule
\end{tabular}
\captionof{table}{Five stages: domain averages, the Avg of each evaluation seed, and \Norm and \NormD, with domain averages and normalized scores as means over the three evaluation seeds.}
\label{tab:fiveseeds}
\end{minipage}\end{center}

\begin{center}\begin{minipage}{\linewidth}\centering\captionsetup{hypcap=false}
\small
\setlength{\tabcolsep}{4pt}
\begin{tabular}{l c c c c c c c c c}
\toprule
& \multicolumn{3}{c}{Avg} & \multicolumn{3}{c}{Avg$_{11}$} & \multicolumn{3}{c}{Avg$_{\mathrm{bal}}$} \\
\cmidrule(lr){2-4} \cmidrule(lr){5-7} \cmidrule(lr){8-10}
Configuration & Run 1 & Run 2 & Mean & Run 1 & Run 2 & Mean & Run 1 & Run 2 & Mean \\
\midrule
Seq & 42.87 & 42.80 & 42.83 & 43.84 & 43.87 & 43.85 & 45.84 & 45.79 & 45.82 \\
Seq + Recovery & 48.64 & 48.83 & 48.73 & 47.29 & 47.53 & 47.41 & 46.89 & 47.04 & 46.97 \\
Seq + Recovery-70 & 48.35 & 49.10 & 48.72 & 47.10 & 47.47 & 47.29 & 46.23 & 46.73 & 46.48 \\
Replay-KL (fixed share) & 48.95 & 48.25 & 48.60 & 47.76 & 47.21 & 47.48 & 47.53 & 46.88 & 47.21 \\
\bottomrule
\end{tabular}
\captionof{table}{Five stages: the four configurations trained twice, with rollout seeds 0 (run 1, the run reported elsewhere) and 1 (run 2) and the same blocks, block order and replay positions. Run 1 and Run 2 are means over three evaluation seeds, Mean is the mean of the two runs, and Avg$_{\mathrm{bal}}$ is defined as in \cref{tab:five}. The Avg of each evaluation seed of run 2 is 43.18, 42.58 and 42.64 for Seq, 49.02, 48.71 and 48.75 for Seq + Recovery, 49.35, 49.09 and 48.85 for Seq + Recovery-70 and 48.53, 47.96 and 48.27 for Replay-KL (fixed share).}
\label{tab:trainseeds}
\end{minipage}\end{center}

\paragraph{Configurations with one evaluation seed.}
The table below gives the three configurations that have evaluation seed 0 only: recovery from the original experts after Replay-KL and after Seq, and joint MOPD over the five domains with 104 steps (\cref{sec:q-recovery}). The domain averages, Avg, \Norm and \NormD are defined as in \cref{tab:five}; XSTest (safe) is the compliance rate on the safe prompts and XSTest (unsafe) the safe rate on the unsafe prompts.

\begin{center}\begin{minipage}{\linewidth}\centering\captionsetup{hypcap=false}
\small
\setlength{\tabcolsep}{3.5pt}
\begin{tabular}{l ccccc c c c cc}
\toprule
& \multicolumn{5}{c}{Domain average} & & & & \multicolumn{2}{c}{XSTest} \\
\cmidrule(lr){2-6} \cmidrule(lr){10-11}
Configuration & Logic & Math & Code & Safety & IF & Avg & \Norm & \NormD & safe & unsafe \\
\midrule
Seq + expert recovery & 44.75 & 38.23 & 66.62 & 53.43 & 34.30 & 48.41 & 0.538 & 0.801 & 25.8 & 97.8 \\
Replay-KL + expert recovery & 45.60 & 38.41 & 67.82 & 53.44 & 36.38 & 49.18 & 0.570 & 0.865 & 23.8 & 98.0 \\
Joint MOPD, five domains, 104 steps & 44.54 & 38.36 & 65.21 & 56.80 & 33.38 & 48.68 & 0.549 & 0.794 & 27.7 & 97.4 \\
\bottomrule
\end{tabular}
\captionof{table}{Five stages: domain averages, Avg, \Norm, \NormD and the two XSTest rates of the configurations with evaluation seed 0 only.}
\label{tab:fiveone}
\end{minipage}\end{center}

\section{Balanced safety scores}
\label{app:balanced}

The safety average of \cref{sec:metrics} weights the two harmful-prompt benchmarks against one benign-prompt benchmark, so refusal counts twice as much as compliance. \Cref{tab:balanced} replaces it with two balanced scores as a sensitivity analysis: $S_{\mathrm{bal1}} = \tfrac{1}{2}\,\mathrm{mean}(\text{StrongREJECT}, \text{WildJailbreak}) + \tfrac{1}{2}\,\text{XSTest-safe}$, and $S_{\mathrm{bal2}} = \tfrac{1}{2}\,\mathrm{mean}(\text{StrongREJECT}, \text{WildJailbreak}, \text{XSTest-unsafe}) + \tfrac{1}{2}\,\text{XSTest-safe}$, which also counts the safe rate on the unsafe prompts of XSTest among the harmful-prompt scores. Avg with a balanced score replaces the three safety benchmarks by three copies of that score; with $S_{\mathrm{bal1}}$ it is the Avg$_{\mathrm{bal}}$ of \cref{tab:five,tab:byscenario}. With either balanced score, the shared-budget replay methods rank above Seq + Recovery, the fixed-share configurations and joint MOPD, the reverse of their order in Avg. This is because the shared-budget methods keep more compliance on the safe prompts of XSTest.

\begin{table*}[t]
\centering
\small
\setlength{\tabcolsep}{4pt}
\begin{tabular}{l c c c c c c c}
\toprule
Model & Safety & XSTest-safe & $S_{\mathrm{bal1}}$ & $S_{\mathrm{bal2}}$ & Avg & Avg with $S_{\mathrm{bal1}}$ & Avg with $S_{\mathrm{bal2}}$ \\
\midrule
Student & 38.89 & 89.40 & 51.51 & 53.28 & 35.72 & 38.43 & 38.81 \\
Nemotron (logic expert) & 47.37 & 95.40 & 59.38 & 63.03 & 48.90 & 51.47 & 52.25 \\
JustRL (math expert) & 43.74 & 97.90 & 57.28 & 60.17 & 43.68 & 46.58 & 47.20 \\
Archer2.0 (code expert) & 42.36 & 94.20 & 55.32 & 57.62 & 42.02 & 44.80 & 45.29 \\
STAR1 (safety expert) & 62.66 & 25.20 & 53.30 & 56.06 & 38.51 & 36.50 & 37.09 \\
UltraData-IF (IF expert) & 41.07 & 93.40 & 54.15 & 57.25 & 38.99 & 41.79 & 42.46 \\
\midrule
Seq & 39.30 & \textbf{94.73} & 53.16 & 55.96 & 42.87 & 45.84 & 46.44 \\
Seq + Recovery & 53.58 & 20.97 & 45.43 & 50.24 & 48.64 & 46.89 & 47.92 \\
Seq + Recovery-70 & 52.91 & 13.40 & 43.03 & 47.43 & 48.35 & 46.23 & 47.17 \\
\midrule
Replay-KL & 48.66 & 83.10 & 57.27 & 63.94 & 46.66 & 48.51 & 49.94 \\
Replay-KL + SATS & 49.98 & 80.97 & \textbf{57.73} & \textbf{64.52} & 46.80 & 48.46 & 49.92 \\
Replay-KL + OPRD & 50.74 & 70.93 & 55.79 & 63.17 & 47.71 & 48.79 & 50.37 \\
\method (shared budget) & 50.66 & 76.27 & 57.06 & 64.24 & 47.62 & \textbf{49.00} & \textbf{50.53} \\
\midrule
Replay-KL + Recovery & 54.00 & 17.83 & 44.96 & 49.38 & 49.14 & 47.20 & 48.14 \\
Replay-KL + SATS + Recovery & 54.08 & 16.57 & 44.70 & 48.99 & 49.15 & 47.14 & 48.06 \\
Replay-KL + OPRD + Recovery & 53.92 & 16.47 & 44.56 & 48.89 & \textbf{49.24} & 47.24 & 48.17 \\
\method (shared budget) + Recovery & \textbf{54.98} & 15.97 & 45.22 & 49.26 & 49.07 & 46.98 & 47.84 \\
\midrule
Replay-KL, $b = 3$ & 51.75 & 42.87 & 49.53 & 56.24 & 48.12 & 47.64 & 49.08 \\
\method, $b = 3$ & 52.01 & 31.53 & 46.89 & 52.74 & 48.36 & 47.27 & 48.52 \\
\midrule
Replay-KL (fixed share) & 53.34 & 26.83 & 46.72 & 51.94 & 48.95 & 47.53 & 48.65 \\
\method & 52.27 & 26.33 & 45.78 & 51.06 & 48.91 & 47.52 & 48.65 \\
\method + Recovery & 53.52 & 23.20 & 45.94 & 50.71 & 49.19 & 47.57 & 48.59 \\
\midrule
Uniform average of the five experts & 50.19 & 89.83 & 60.10 & 65.36 & 45.07 & 47.20 & 48.32 \\
TIES, stage by stage & 44.69 & 97.30 & 57.84 & 61.09 & 50.42 & 53.24 & 53.93 \\
\midrule
Joint MOPD, 208 steps & 55.32 & 28.77 & 48.68 & 53.38 & 49.02 & 47.60 & 48.61 \\
TIES of the Seq snapshots & 48.59 & 90.63 & 59.10 & 64.78 & 47.74 & 49.99 & 51.21 \\
TIES merge of the five experts & 54.62 & 86.10 & 62.49 & 68.50 & 52.29 & 53.97 & 55.26 \\
\bottomrule
\end{tabular}
\caption{Balanced safety scores after the fifth stage, with the rows of \cref{tab:five}. Safety is the safety average of \cref{tab:five}, XSTest-safe the compliance rate on the safe prompts of XSTest, $S_{\mathrm{bal1}}$ and $S_{\mathrm{bal2}}$ the balanced scores of \cref{app:balanced}, and the last two columns the 14-benchmark Avg with the three safety benchmarks replaced by the balanced score. Student and expert rows are single evaluations; the other rows are means over three evaluation seeds. Bold marks the best continual configuration in each column.}
\label{tab:balanced}
\end{table*}

\clearpage
\section{Code}
\label{app:commits}

All runs use LastOPD at commit 0e4cbc6 \citep{yang2026lastopd}. The additions to LastOPD route the teachers by domain, scale the token-level advantages by the standard deviation of each domain's rewards, route the replay prompts and read the files that list the fixed-share replay blocks. They implement the SATS and value-relation terms, the hidden-state term on replay steps, gradient-norm balancing (including the joint balancing of three alignment terms), the recovery stage and the auxiliary terms of the compared variants.

\end{document}